\documentclass[11pt]{article}

\title{Robust Subspace Clustering via Thresholding}

\renewcommand\footnotemark{}

\author{Reinhard Heckel and Helmut B\"olcskei 
\thanks{Parts of this paper were presented at the 2013 IEEE International Conference on Acoustics, Speech, and Signal Processing (ICASSP) \cite{heckel_subspace_2013} and  at the 2013 IEEE International Symposium on Information Theory (ISIT) \cite{heckel_noisy_2013}.}
 \\[0.5em]
  \multicolumn{1}{p{.7\textwidth}}{\centering Dept. of IT \& EE, ETH Zurich, Switzerland}}

\date{July 2013; last revised August 2015}

\usepackage{setspace}

\usepackage{amssymb}

\usepackage[figuresright]{rotating}

\usepackage{endnotes}

\usepackage{amsmath,amssymb}
\usepackage{graphicx}
\usepackage{dcolumn}
\usepackage{bm}
\usepackage{amscd,amsthm}
\usepackage{ifthen}
\usepackage{appendix}

\usepackage{mathtools} 

\usepackage[applemac]{inputenc}

\usepackage{supertabular}
\usepackage{booktabs}

\usepackage{tikz}

\usepackage{colortbl}
\definecolor{tblblue}{RGB}{101,124,191}
\definecolor{tblred}{rgb}{1,0.93,0.93}
\definecolor{DarkBlue}{rgb}{0,0,0.7}

\usepackage{pgfplots}
\usepackage{subfigure}
\usetikzlibrary{pgfplots.groupplots}
\pgfplotsset{colormap/cool}

\usepackage{enumitem}
\usepackage{bm}

\usepackage{hyperref}
\hypersetup{
    bookmarks=true,         
    unicode=false,          
    pdftoolbar=true,        
    pdfmenubar=true,        
    pdffitwindow=false,     
    pdfstartview={FitH},    
    pdfauthor={Reinhard Heckel},     
    pdfsubject={Subject},   
    pdfcreator={Reinhard Heckel},   
    pdfproducer={Producer}, 
    pdfnewwindow=true,      
    colorlinks=true,       
    linkcolor=red,          
    citecolor=green,        
    filecolor=magenta,      
    urlcolor=cyan           
}

 \newcommand{\ve}[1]{\mathbf{#1} }

 \newcommand{\mbf}[0]{\mathbf }

\newcommand\norm[2][\Tnorm]{\ensuremath{{\left\Vert #2 \right\Vert}_{#1}}}

\newcommand\Tinnerprod{}
\newcommand{\innerprod}[3][\Tinnerprod]{\ifthenelse{\equal{#1}{}}{\ensuremath{\left<#2,#3\right>}}{\ensuremath{\left<#2,#3\right>_{#1}}}}

\newcommand\vect[1]{\mathbf #1}

\newcommand{\US}[1]{\mathbb S^{#1-1}} 
\newcommand{\USL}[0]{\mathbb S}

\newcommand\PR[1]{\ensuremath{ {\mathrm{P}}\!\left[#1\right]}}

\newcommand\Tex{}
\newcommand\EX[2][\Tex]{
\ifthenelse{\equal{#1}{}}{{\mathbb E}\!\left[#2\right]}{\ensuremath{{\mathbb E}_{#1}\left[ #2\right]}}}

\newcommand\Var[2][\Tex]{
\ifthenelse{\equal{#1}{}}{{\mathrm{Var} }[#2]}{\ensuremath{\mathrm{Var}_{#1}\left[ #2\right]}}}

\newcommand\ignore[1]{}

\newcommand\defeq{\coloneqq}

\newcommand{\reals}{\mathbb R}

\newtheorem{lemma}{Lemma}
\newtheorem{definition}{Definition}
\newtheorem{corollary}{Corollary}
\newtheorem{theorem}{Theorem}

\newtheorem{proposition}{Proposition}

\newtheorem*{algorithm}{TSC algorithm}

\newtheorem*{nfc}{No false connections property}

\newcommand\comp[1]{ \overline{#1}}

\newcommand{\pinv}[1]{  {#1}^{ \dagger } } 
\newcommand{\inv}[1]{  {#1}^{ -1 } } 

\newcommand{\herm}[1]{{#1}^T} 
\newcommand{\transp}[1]{{#1}^T} 

\newcommand{\id}[1]{1_{\{#1\}}}

\newcommand{\mintwo}[2]{ #1 \wedge #2 } 

\renewcommand{\d}{d} 

\newcommand{\affp}{\mathrm{aff}_\infty}
\newcommand{\aff}{\mathrm{aff}}

\newcommand{\cS}{S} 
\renewcommand{\d}{d} 
\newcommand{\X}{\mathcal X} 

\usepackage{bm}

\newcommand{\va}{\vect{a}}  
\newcommand{\vb}{\vect{b}}
\newcommand{\vc}{\vect{c}}  
  
\renewcommand{\ve}{\vect{e}}

\newcommand{\vp}{\vect{p}}

\newcommand{\vu}{\vect{u}}  
\newcommand{\vv}{\vect{v}}  

\newcommand{\vx}{\vect{x}}  
\newcommand{\vy}{\vect{y}}  
\newcommand{\vz}{\vect{z}}

\newcommand{\mA}{\vect{A}}  
\newcommand{\mB}{\vect{B}}

\newcommand{\mI}{\vect{I}}

\newcommand{\mU}{\vect{U}}
\newcommand{\mV}{\vect{V}}

\newcommand{\mX}{\vect{X}}

\newcommand{\mZ}{\vect{Z}}

\renewcommand{\S}{\mathcal T}

\newcommand{\D}{\mathcal D}
\newcommand{\tD}{\mathcal E}

\renewcommand{\O}{\mathcal O} 
\newcommand{\q}{q} 
\newcommand{\s}{s}

\renewcommand{\l}{\ell} 

\newcommand\LM[1]{\mathcal L(#1)}

\newcommand{\C}{C}

\begin{document}

\maketitle

\begin{abstract}
The problem of clustering noisy and incompletely observed high-dimensional data points into a union of low-dimensional subspaces and a set of outliers is considered. The number of subspaces, their dimensions, and their orientations are assumed unknown. We propose a simple low-complexity subspace clustering algorithm, which applies spectral clustering to an adjacency matrix obtained by thresholding the correlations between data points. 
In other words, the adjacency matrix is constructed from the nearest neighbors of each data point in spherical distance. A statistical performance analysis shows that the algorithm exhibits robustness to additive noise and succeeds even when the subspaces intersect.  Specifically, our results reveal an explicit tradeoff between the affinity of the subspaces and the tolerable noise level. We furthermore prove that the algorithm succeeds even when the data points are incompletely observed with the number of missing entries allowed to be (up to a log-factor) linear in the ambient dimension. We also propose a simple scheme that provably detects outliers, and we present numerical results on real and synthetic data. 
\end{abstract}

\section{Introduction}
\label{sec:intro}

One of the major challenges in modern data analysis is to extract relevant information from large high-dimensional data sets. The relevant features 
are often of limited complexity, or, more specifically, have low-dimensional structure. For example, images of faces are high-dimensional as the number of pixels is typically large, whereas the set of images of a given face under varying illumination conditions approximately lies in a 9-dimensional linear subspace \cite{basri_lambertian_2003}. 
This and similar insights for other types of data have motivated research on finding low-dimensional structure in high-dimensional data. 
A prevalent low-dimensional structure is that of data points lying in a union of  low-dimensional subspaces. The problem of finding the assignments of the data points to these (unknown) subspaces is referred to as subspace clustering \cite{vidal_subspace_2011} or hybrid linear modeling \cite{zhang_hybrid_2012}. 
An example application of subspace clustering is the following. Given a set of images of faces under varying illumination conditions, cluster the images such that each of the resulting clusters corresponds to a single person \cite{ho_clustering_2003}. Other application areas include unsupervised learning, image representation and segmentation \cite{hong_multiscale_2006}, computer vision, specifically motion segmentation \cite{vidal_motion_2004,rao_motion_2008}, and disease detection \cite{kriegel_clustering_2009}; we refer to \cite{vidal_subspace_2011} for a more complete list. 

Often the data available is corrupted by noise and contains outliers. The general subspace clustering problem we consider takes this into account and can be formalized as follows. 
Suppose we are given a set of $N$ data points in $\reals^m$, denoted by $\X$, and assume that 
\[
\X = \X_1 \cup ...  \cup  \X_L \cup  \O.
\] 
Here, $\O$ denotes a set of outliers and the $n_\l \defeq |\X_\l|$ points in $\X_\l$ are given by
\begin{align}
\vx_j^{(\l)} = \vy_j^{(\l)}  + \ve^{(\l)}_j, \quad j = 1,...,n_\l
\label{eq:pisrep}
\end{align}
where $\vy_j^{(\l)} \in \cS_\l$ with $\cS_\l$ a $d_\l$-dimensional linear subspace of $\reals^m$ and $\ve^{(\l)}_j \in \reals^m$ is noise. 
The association of the points in $\X$ with the $\X_\l$ and $\O$, the number of subspaces $L$, their dimensions $d_\l$, and their orientations are all unknown. We want to cluster the data points in $\X$, i.e., find their assignments to the sets $\X_1,...,\X_L, \O$.
Once these assignments have been identified, it is straightforward to extract approximations (recall that we have access to noisy data only) of the subspaces $\cS_\l$ through principal component analysis (PCA). 

Numerous  approaches to subspace clustering have been proposed in the literature, including algebraic, statistical, and spectral clustering methods; we refer to \cite{vidal_subspace_2011} for an excellent survey. 
Spectral clustering methods (see \cite{luxburg_tutorial_2007} for an introduction) have found particularly widespread use thanks to their excellent performance properties and efficient implementations. 
At the heart of spectral clustering lies the construction of an adjacency matrix $\mA \in \reals^{N\times N}$, where the $(i,j)$th entry of $\mA$ measures the similarity between the data points $\vx_i, \vx_j \in \X$. A typical measure of similarity is, e.g., $e^{-\mathrm{dist}(\vx_i,\vx_j)}$, where $\mathrm{dist}(\cdot,\cdot)$ is some distance measure \cite{vidal_subspace_2011}. 
The association of the points in $\X$ to the subspaces $\cS_\l$ is then estimated by applying spectral clustering  to $\mA$. 

As noted in \cite{soltanolkotabi_robust_2013} there are only a few subspace clustering algorithms that are computationally tractable \emph{and} known to succeed \emph{provably} under non-restrictive conditions such as, e.g., intersecting subspaces. 
A notable exception is the sparse subspace clustering (SSC) algorithm proposed by Elhamifar and Vidal \cite{elhamifar_sparse_2009,elhamifar_sparse_2013}, which applies  spectral clustering to an adjacency matrix 
$\mA$ obtained by sparsely representing each data point in terms of all the other data points through $\ell_1$-minimization. 
SSC provably succeeds (in a sense to be made precise later) in the noiseless case under very general conditions, as shown by Soltanolkotabi and Cand\`es in
\cite{soltanolkotabi_geometric_2011} via an elegant (geometric function) analysis. Most importantly, the results in \cite{soltanolkotabi_geometric_2011} reveal that SSC succeeds even when the subspaces $\cS_\l$ intersect (the linear subspaces $\cS_\l$ and $\cS_k$ are said to intersect if $\cS_\l \cap \cS_k \neq \{\vect{0}\}$).

Analytical performance results for subspace clustering of noisy data are even more scarce. 
Vidal noted in \cite{vidal_subspace_2011} that ``the development of theoretically sound algorithms~[...]~in the presence of noise and outliers is a very important open challenge.'' 
A significant step towards addressing this challenge was reported recently in \cite{soltanolkotabi_robust_2013}. 
Specifically, the robust SSC (RSSC) algorithm in \cite{soltanolkotabi_robust_2013} replaces the $\ell_1$-minimization step in SSC by an $\ell_1$-penalized least squares, i.e., Lasso, step and provably succeeds under Gaussian noise and 
 under very general conditions on the orientations of the subspaces $\cS_\l$. 
To construct the adjacency matrix $\mA$, SSC for the noiseless and RSSC for the noisy case require the solution of $N$ $\ell_1$-minimization and $N$ Lasso instances, respectively, each in $N$ variables; this poses significant computational challenges for large data sets.

\paragraph*{Contributions:}
We present a simple and computationally efficient subspace clustering algorithm, which applies spectral clustering to an adjacency matrix $\mA$ obtained by thresholding correlations between the data points in $\X$. In other words, $\mA$ is constructed from the nearest neighbors of each data point in spherical distance. 
The resulting algorithm is termed thresholding-based subspace clustering (TSC). 

For our analytical results, we consider a semi-random data model with deterministic subspaces and the data points sampled at random from these subspaces. Specifically, we sample uniformly at random from the intersection of the unit sphere and the corresponding subspace. 
The gist of the results we obtain is that TSC  succeeds \emph{provably}---even when the data is corrupted by additive Gaussian noise or incompletely observed---provided that the subspaces are sufficiently distinct and $\X$ contains sufficiently many points from each subspace. 
The measure of success we use in the noisy case and for incomplete observations is an intermediate performance measure as it does not address the clustering error, i.e., the fraction of misclassified points, directly. 
Rather, it guarantees that in the graph $G$ with adjacency matrix $\mA$, for each $\l$, the nodes corresponding to the points in $\X_\l$ are connected to other nodes corresponding to points in $\X_\l$ only. The same performance measure was used in \cite{soltanolkotabi_geometric_2011,soltanolkotabi_robust_2013,liu_robust_2010,dyer_greedy_2013} for SSC, RSSC, LRR (low-rank representation), and SSC-orthogonal matching pursuit (SSC-OMP). 
In the noiseless case, we obtain significantly stronger results which come in terms of conditions guaranteeing that the clustering error is zero. This is accomplished by analyzing the connectivity properties of the random nearest neighbor graph induced by the statistical data model we employ. 
The corresponding results (Theorems \ref{thm:TSCprob} and \ref{thm:aff2}) apply, however, to a smaller range of parameters $d_\l, n_\l$ when compared to ensuring no false connections only (Corollary \ref{cor:ofnoisycase}).

Our results for noisy data reflect the intuition that the more distinct the orientations of the subspaces, the more noise TSC tolerates. What is more, we find that TSC can succeed even under massive noise, provided that the subspaces are sufficiently low-dimensional. 
In practical applications the data points to be clustered are often incompletely observed, due to, e.g., scratches on images. 
Assuming that the orientation of the subspaces and the points on the subspaces are random,  
we prove that TSC can succeed even when the number of (arbitrary) missing entries in each data vector is (up to a log-factor) linear in the ambient dimension. 
Finally, we propose a simple scheme for outlier detection 
and we report corresponding analytical performance guarantees. 
Numerical results on synthetic data, on handwritten digits taken from the MNIST data base \cite{mnist_2013}, and on images of faces taken from the extended Yale Face Database B \cite{georghiades_illumination_2001,lee_acquiring_2005} complement our analytical results.

\paragraph*{Relation to previous work:} 
Lauer and Schnorr \cite{lauer_spectral_2009} apply spectral clustering to an adjacency matrix constructed from correlations between data points, albeit, without thresholding.  
More importantly, no analytical performance results are available for the algorithm in \cite{lauer_spectral_2009}. 
The local subspace affinity algorithm \cite{yan_general_2006} and the spectral local best-fit flats (SLBF) algorithm \cite{zhang_hybrid_2012} are based on spectral clustering applied to an adjacency matrix that is built from the nearest neighbors of each data point in Euclidean distance. 
Liu et al.~\cite{liu_robust_2010} consider spectral clustering applied to an adjacency matrix obtained from a low-rank representation (LRR) of the data points through nuclear norm minimization. 
The performance analysis conducted in \cite[specifically Theorem 3.1]{liu_robust_2010} 
shows that LRR succeeds provided that the subspaces $\cS_\l$ are independent (the linear subspaces $\cS_\l$ are said to be independent if the dimension of their (set) sum is equal to the sum of their dimensions), which implies that the subspaces must not intersect. 
 Moreover,  the nuclear norm minimization required by LRR results in significant computational complexity. 
The analytical conditions guaranteeing success of SSC, and RSSC for the noisy case, reported in \cite{soltanolkotabi_geometric_2011} and \cite{soltanolkotabi_robust_2013}, respectively, are very similar to those found for TSC in this paper. TSC is, however, computationally much less demanding than SSC/RSSC. 
These complexity savings may come at the cost of clustering performance. 
Experiments on real and synthetic data, many of which are reported in Section~\ref{sec:numres}, show that while there are situations where TSC outperforms SSC, SSC outperforming TSC is more common. 
Dyer et al.~\cite{dyer_greedy_2013} propose to substitute the $\ell_1$-minimization step in SSC by an orthogonal matching pursuit (OMP) step, and derive  
 performance guarantees for the resulting SSC-OMP algorithm. 

Lerman and Zhang \cite{lerman_robust_2011} consider the problem of recovering multiple subspaces from data drawn from a distribution on the union of these subspaces and pose recovery as a non-convex optimization problem. No computationally tractable algorithm to solve this recovery problem \cite{lerman_robust_2012} 
seems to be available, though.

The problem of fitting a single low-dimensional subspace to a data set consisting of a modest number of noisy inliers and a large number of outliers was considered in \cite{lerman_robust_2012}, along with a convex programming algorithm with analytical performance guarantees. 
Chen and Lerman \cite{chen_spectral_2009,chen_foundations_2009} propose subspace clustering algorithms based on spectral clustering, termed Spectral Curvature Clustering (SCC) and Theoretical SCC (TSCC), along with a strategy for outlier detection, and provide corresponding probabilistic performance analyses.

\paragraph*{Outline of the paper:}
The remainder of this paper is organized as follows. 
In Section~\ref{sec:tsc}, we introduce the TSC algorithm. 
Sections~\ref{sec:detss} and \ref{sec:noise} contain analytical performance results for the noiseless and the noisy case, respectively. In Section~\ref{sec:erasure}, we analyze the impact of incompletely observed data points on the performance of TSC. 
 Section~\ref{sec:detoutl} describes an outlier detection scheme and contains corresponding  performance results. In
Section~\ref{sec:comp}, we compare our analytical performance results for TSC to analytical performance results for SSC/RSSC and further subspace clustering algorithms. Section~\ref{sec:numres} contains numerical results on synthetic and on real data, 
including a comparison of TSC to SSC/RSSC. 

We discuss the various settings (noiseless, noisy, incomplete observations, and outliers) in an isolated fashion to keep the exposition accessible. All proofs are relegated to appendices. 

\paragraph*{Notation:} We use lowercase boldface letters to denote (column) vectors, e.g., $\vx$, and uppercase boldface letters to designate matrices, e.g., $\mA$. The superscript $\herm{}$ stands for transposition.  
For the vector $\vx$, 
$x_q$ denotes its $q$th entry. For the matrix $\mA$, $\mA_{ij}$ designates the entry in its $i$th row and $j$th column, $\pinv{\mA} \defeq \inv{(\herm{\mA} \mA )} \herm{\mA}$ is its pseudo-inverse, $\norm[2\to 2]{\mA} \defeq\;$ $\max_{\norm[2]{\vv} = 1  } \norm[2]{\mA \vv}$ its spectral norm, and $\norm[F]{\mA} \defeq (\sum_{i,j} |\mA_{ij}|^2 )^{1/2}$ its Frobenius norm.  $\mI_m$ denotes the $m\times m$ identity matrix. 
 $\log(\cdot)$ is the natural logarithm, 
 and $x \land y$ stands for the minimum of $x$ and $y$. 
$\LM{\cdot}$ denotes the Lebesgue measure. 
For the set $\S$, $|\S|$ designates its cardinality and $\comp{\S}$ is its complement. 
The set $\{1,...,N\}$ is denoted by $[N]$.
We write $\mathcal N( \boldsymbol{\mu},\boldsymbol{\Sigma})$ for a Gaussian random vector with mean $\boldsymbol{\mu}$ and covariance matrix $\boldsymbol{\Sigma}$. 
The unit sphere in $\reals^m$ is $\US{m} \defeq \{ \vx \in \reals^m \colon \norm[2]{\vx} = 1 \}$. 
$\id{A}(\cdot)$ denotes the indicator function of the set $A$. For notational convenience, we use the shorthand $\max_{k\neq \l}$ for $\max_{k \in [L] \colon k\neq \l}$ and $\max_{k,\l \colon k\neq \l}$ for $\max_{k,\l  \in [L ]\colon  k\neq \l}$. Similarly, $\max_{k\neq \l, j}$ is shorthand for $\max_{k \in [L] \colon k\neq \l, j \in [n_k]}$. 
We let $n_{\min} = \min_{\l \in [L]} n_\l$, $n_{\max} = \max_{\l \in [L]} n_\l$, and $\d_{\max} = \max_{\l \in [L]} \d_\l$. 
For random variables $X,Y$, we write $X\sim Y$ to indicate that $X$ and $Y$ have the same distribution. 
We say that a subgraph $H$ of a graph $G$ is connected if any two nodes in $H$ can be joined by a path that has all intermediate nodes lying in $H$.  
The subgraph $H$ of $G$ is called a connected component of $G$ if $H$ is connected and if there are no connections between nodes in $H$ and the remaining nodes in $G$ \cite{luxburg_tutorial_2007}. 
The $k$-nearest neighbor graph of a set of points $\{\va_1,...,\va_n\}$ with respect to the metric $s$ is the undirected graph with vertex set $\{\va_1,...,\va_n\}$ and edges between $\va_i$ and $\va_j$ if either $\va_i$ is among the $k$ nearest neighbors of $\va_j$ or $\va_j$ is among the $k$ nearest neighbors of $\va_i$, in both cases with respect to the metric $s$.

\section{\label{sec:tsc}The TSC algorithm}

The formulation of the thresholding-based subspace clustering (TSC) algorithm provided below assumes that outliers have already been removed from the data set $\X$, e.g., through the outlier detection scheme described in Section~\ref{sec:detoutl}, and that the data points in $\X$ are normalized. The latter assumption is relevant for Steps 1 and 2 below and is not restrictive as the data points can be normalized prior to clustering.

\begin{algorithm}
\label{alg:TSC}
Given a set of data points $\X$, an estimate of the number of subspaces $\hat L$ (estimation of $L$ from $\X$ is discussed in Section \ref{sec:estofL}), and the parameter $\q$ (the choice of $\q$ is discussed below), perform the following steps:

{\bf Step 1:} For every $\vx_j \in \X$, identify the set $\S_j \subset [N] \!\setminus\! \{j\}$ (recall that $N = |\X|$) of cardinality $q$ defined through 
\begin{equation*}
\left| \innerprod{\vx_j}{ \vx_i} \right| \geq \left| \innerprod{\vx_j  }{ \vx_p}\right|,  \text{ for all } i \in \S_j \text{ and all } p \notin \S_j.
\end{equation*}

{\bf Step 2:} Let $\vz_j \in \reals^N$ be the vector with $i$th entry 
$\exp(-2\arccos(\left| \innerprod{\vx_j }{ \vx_i }\right|))$ 
if $i\in \S_j$, and $0$ if $i\notin \S_j$.

{\bf Step 3:} Construct the adjacency matrix $\mA$ according to $\mA = \mZ + \herm{\mZ}$, where $\mZ = [\vz_1\,\cdots\,\vz_N]$. 

{\bf Step 4:} Apply normalized spectral clustering \cite{luxburg_tutorial_2007,ng_spectral_2001} to $(\mA, \hat L)$. 
\end{algorithm}

Since $\arccos(z)$ is decreasing in $z$ for $z \in [0,1]$, the set $\S_j$ is the set of $q$ nearest neighbors of $\vx_j$ with respect to the metric\footnote{$\tilde \s$ is not a distance metric in the strict sense as $\tilde s(\vx,-\vx) = 0$, but $-\vx \neq \vx$ for $\vx\neq \mathbf{0}$. It satisfies, however, the defining properties of a pseudo-distance metric \cite{kelley_general_1975}.}  
 $\tilde \s (\vx_i,\!\vx_j)\! \defeq \!  \arccos(\left|\innerprod{\vx_i}{\vx_j}\right|)$.  
TSC is therefore built on the premise, explained in Sections \ref{sec:motsimmeas} and \ref{sec:lstsc}, that the vectors close to $\vx_j$ in terms of the distance $\tilde \s$ also lie in the subspace $\vx_j$ lies in. 
This 
can be formalized in terms of the $q$-nearest neighbor graph with respect to the distance $\tilde s$, i.e., the graph $G$ with adjacency matrix $\mA$, simply referred to as ``the graph $G$'' in the remainder of the paper. 
  If each connected component in the graph $G$ corresponds to exactly one of the sets $\X_\l$, and if $\hat L = L$, then (normalized) spectral clustering yields correct segmentation of the data (i.e., it delivers the oracle segmentation $\X = \X_1 \cup ...  \cup  \X_L$ of $\X$) \cite[Prop.~4; Sec.~7]{luxburg_tutorial_2007} and the clustering error will be zero. 
Even when the connected components of $G$ do not correspond to the $\X_\l$ exactly, but the weights in the adjacency matrix $\mA$ corresponding to pairs of points 
that belong to different subspaces are small enough, TSC may still cluster the data correctly. The numerical results in Section \ref{sec:numres} demonstrate that the spectral clustering step can cope with such imperfections. 

  In the noiseless case we will be able to establish conditions that ensure zero clustering error.   In the noisy case we will work with an intermediate, albeit sensible, performance measure, also employed to assess the performance of the clustering algorithms considered in  \cite{dyer_greedy_2013,liu_robust_2010,soltanolkotabi_geometric_2011,soltanolkotabi_robust_2013}. 
  This performance measure is formalized through the following property: 
\begin{nfc}
\label{cond:nfc} G has no false connections if, for all $\l \in [L]$, the nodes in $G$ corresponding to $\X_\l$ are connected to other nodes corresponding to $\X_\l$ only. 
\end{nfc}

Ensuring the absence of false connections, does, however, not guarantee 
 that the connected components in $G$ correspond to the $\X_\l$, as the points in a given set $\X_\l$ may  split up into two or more distinct clusters. 
TSC (with input parameter $\q$) counters this problem by imposing that each node is connected to at least $q$ other nodes and choosing $q$ not too small relative to the $n_\l$. Taking $q$ too large, however, increases the chances of points from different sets $\X_\l$ being clustered together, thereby violating the no false connections property. 
Our analytical performance results for the noiseless case ensure correct segmentation of $\X$ by guaranteeing that $G$ has no false connections and the subgraphs corresponding to the $\X_\l$ are connected,  
 provided that $\q$ 
is sufficiently large relative to the values $\log n_\l$ and sufficiently small relative to the $n_\l$. 
The specific choice of $\q$ within this range will be seen to be irrelevant in terms of the \emph{analytical} performance guarantees we obtain, but it does have an impact on the actual performance of TSC in practice.

\subsection{Measuring similarity via $\tilde \s$
\label{sec:motsimmeas}}

To see that $\tilde \s (\vx_i, \vx_j) =  \arccos(\left|\innerprod{\vx_i}{\vx_j}\right|)$ leads to a sensible similarity measure for subspace clustering, 
consider the noiseless case, suppose that the subspaces $\cS_\l$ are orthogonal to each other, and take $q \leq \min_\l (|\X_\l|-d_\l)$. 
Then, $G$ has no false connections thanks to $\innerprod{\vx_p}{ \vx_j} = 0$ for all $\vx_p \in \X_\l, \vx_j \in \X_k$, $\l \neq k$, while for each $\l$, 
there are at least $|\X_\l|-d_\l$ inner products $\innerprod{\vx_p}{ \vx_j}$ with $\vx_p, \vx_j \in \X_\l$ that are non-zero, as no more than $d_\l$ points in a $d_\l$-dimensional subspace can be orthogonal to each other. 
The analytical results in the following sections show that $G$ can actually satisfy the no false connections property 
under much more general conditions, in particular even when the subspaces intersect. 
What lies beneath these results is the fact that for the statistical data model used throughout the paper, the magnitude of the inner product between the data points from the same subspace is typically larger than that between data points from different subspaces. 
This is also true in many practical problems, as e.g., the numerical results on clustering handwritten digits in Section \ref{sec:clusMNIST} show. 
A different rationale for $\tilde \s$ leading to a suitable similarity measure for subspace clustering is based on sparse signal representation theory, and is given next. 

\subsection{Least-squares TSC \label{sec:lstsc}}

A natural substitute for Step 2 in the TSC algorithm is to construct $\vz_j$ from the best linear approximation of $\vx_j$ in terms of the points indexed by $\S_j$. Specifically, let ${\mX}_{\S_j}$ be the matrix whose columns are the vectors in $\X$ indexed by $\S_j$, and  
substitute Step 2 by:

{\bf Step 2-LS:}  Set the entries of $\vz_j \in \reals^N$ indexed by ${\S_j}$ to the absolute values of  $\pinv{\mX}_{\S_j}  \vx_j$ and all other entries of $\vz_j$ to zero. 

The TSC algorithm with Step 2 replaced by Step 2-LS will henceforth be referred to as least squares (LS)-TSC. 

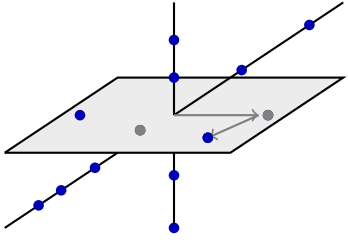
\begin{figure}
\begin{center}
	\begin{tikzpicture}
	\fill [gray!15] (0,0) -- (1.5,1) -- (4.5,1) -- (3,0) -- (0,0);
	\draw[thick] (0,0) -- (1.5,1) -- (4.5,1) -- (3,0) -- (0,0);
	\draw[thick] (0,-1) -- (1.5,0);
	\draw[thick] (2.25,0.5) -- (4.5,2);
	\draw[thick] (2.25,-1) -- (2.25,0);
	\draw[thick] (2.25,0.5) -- (2.25,2/1);
	
	\foreach \point in {(2.25,-1),(2.25,-0.3),(2.25,1),(2.25,1.5) } \fill [DarkBlue,opacity=1] \point circle (2pt);
	\foreach \n in {-1.2,-1,-0.7,0.6,1.2}\fill [DarkBlue,opacity=1] (1.5*\n+2.25,\n+0.5) circle (2pt);
	
	\foreach \point in {(1,0.5),
	(2.7,0.2), 
	(1.8,0.3), 
	(3.5,0.5) 
	}
	\fill [DarkBlue,opacity=1] \point circle (2pt);
		
	\draw[thick,->,gray] (2.25,0.5) -- (3.3755 , 0.5);
	\draw[thick,->,gray] (3.3755 , 0.5) -- (2.7225 , 0.21   );
	\foreach \point in {(1.8,0.3),(3.5,0.5)} \fill [gray,opacity=1] \point circle (2pt);
	
	\fill [DarkBlue,opacity=1] (2.7,0.2) circle (2pt);
	
	\end{tikzpicture}
\end{center}
\caption{\label{fig:sparsrep}
Each data point in a $d_\l$-dimensional subspace $\cS_\l$ can be represented as a linear combination of at most $d_\l$ other points from $\cS_\l$.
}
\end{figure}

The LS-variant of the TSC algorithm allows us to elicit a relationship between TSC, SSC, and SSC-OMP, with the common element being given by the insight that all three algorithms build their adjacency matrix based on sparse signal representation theory. 
To see this, we first note that
each data point in a $d_\l$-dimensional subspace $\cS_\l$ can be represented as a linear combination of at most $d_\l$ other data points in $\cS_\l$. 
A possible approach to measuring similarity,   
 put forward in \cite{elhamifar_sparse_2009,elhamifar_sparse_2013}, finds a sparse representation of each data point $\vx_j\in \X_\l$ in terms of all other data points $\X_1 \cup ... \cup \X_L \! \setminus\! \{\vx_j\}$, and uses the absolute values of the corresponding representation coefficients to 
 quantify the similarity between $\vx_j$ and all other data points. 
The hope is that the non-zeros in this  representation correspond to points in $\X_\l \setminus\{\vx_j\}$, see Figure \ref{fig:sparsrep} for an illustration.  
SSC, SSC-OMP, and LS-TSC implement this idea by finding a sparse linear representation of $\vx_j$ in terms of points in $\X \setminus \{\vx_j\}$ via $\ell_1$-minimization, OMP, and Steps 1 and 2-LS above, respectively. Note that Steps 1 and 2-LS above yield a sparse 
(if $q$ is small) 
linear representation of $\vx_j$ in terms of $\q$ points in $\X \setminus \{\vx_j\}$. 

The formal relationship 
between Steps 2 and 2-LS is brought out by noting that the non-zero entries of $\vz_j$ in Step 2 are given by element-wise application of $\exp(-2\arccos(|\cdot|) )$ to the vector $\transp{\mX}_{\S_j}  \vx_j$ whereas the nonzero entries of $\vz_j$ in Step 2-LS are obtained by element-wise application of $|\cdot|$ to the entries of a weighted version of $\transp{\mX}_{\S_j}  \vx_j$, namely $\pinv{\mX}_{\S_j}  \vx_j = \inv{ (\transp{\mX}_{\S_j} \mX_{\S_j}) }   \transp{\mX}_{\S_j}  \vx_j$. 

As our analytical performance results depend on connectivity properties of the graph $G$ only, and not on the weights assigned to the edges of $G$ (i.e., the values of the non-zero entries of $\mA$), it follows immediately that the corresponding statements hold true verbatim for LS-TSC. Owing to the spectral clustering Step 4, the values of the non-zero entries of $\mA$ do, however, make a difference in terms of practical performance. Corresponding numerical results will be provided in Section \ref{sec:numres}.

\subsection{\label{sec:estofL}Estimation of the number of subspaces}

The number of zero eigenvalues of the normalized Laplacian of the graph $G$ is equal to the number of connected components of $G$ \cite{spielman_spectral_2012}. 
It is therefore sensible to estimate the number of subspaces $L$ as the multiplicity of the eigenvalue $0$ of the normalized Laplacian of $G$. 
In practice, however, weights in the adjacency matrix $\mA$ corresponding to pairs $\vx_i,\vx_j$ that belong to different subspaces might be non-zero, but possibly small, in which case the number of connected components in $G$ may be smaller than $L$. 
This will result in eigenvalues that are not exactly equal to zero, but possibly small. A robust estimator for $L$ taking this into account is the so-called eigengap heuristic \cite{luxburg_tutorial_2007}: 
$
\hat L = \arg  \max_ {i \in [N-1]}  \allowbreak(\lambda_{i+1} - \lambda_{i}),
$
where $\lambda_1\leq \lambda_2 \leq ... \leq \lambda_N$ are the eigenvalues of the normalized Laplacian of $G$.

We note that while satisfying the no false connections property does not say anything about the quality of the estimate $\hat L$, establishing that the connected components in $G$ correspond to the $\X_\l$, 
as done below in the noiseless case, automatically guarantees that $\hat L = L$.

\section{\label{sec:detss}Performance results for the noiseless case}

We first consider noiseless data sets (i.e., $\vx_j^{(\l)} = \vy_j^{(\l)}$ in \eqref{eq:pisrep}) that have no outliers. 
In order to elicit the impact of the relative orientations of the subspaces $\cS_\l$ on the performance of TSC, we take the $\cS_\l$ to be deterministic and choose the points within the $\cS_\l$ randomly. Specifically, we represent the data points in $\cS_\l$ by  
$
\vx_j^{(\l)} = \mU^{(\l)} \va^{(\l)}_j
$
where $\mU^{(\l)} \in \reals^{m\times \d_\l}$ is an orthonormal basis 
for the $\d_\l$-dimensional subspace $\cS_\l$ and the $\va^{(\l)}_j \in \reals^{\d_\l}$ are i.i.d.~uniformly distributed on $\US{\d_\l}$ (throughout the paper, whenever we say that the $\va^{(\l)}_j$ or the $\ve^{(\l)}_j$ are i.i.d., we actually mean i.i.d.~across $j$ \emph{and} $\l$). Therefore,
the data points $\vx_j^{(\l)} = \mU^{(\l)} \va^{(\l)}_j$ are distributed uniformly on $\{\vx \in \cS_\l \colon \norm[2]{\vx} = 1 \}$,  
which ensures that the points are spread out on the subspaces, and avoids degenerate situations where  data points lie in preferred directions. 
For example, suppose that the points on say, a two-dimensional subspace $\cS_1$, are skewed towards two (distinct) directions. Then, there are two sensible segmentations. One is to assign the points corresponding to each direction to separate clusters, the other to assign all points to one cluster. 

Our results will be expressed in terms of two different notions of affinity between subspaces, namely 
\[
\affp(\cS_k,\cS_\l) \defeq \norm[2\to 2]{ \herm{\mU^{(k)}} \mU^{(\l)}  }
\]
and 
\[
\aff(\cS_k,\cS_\l) \defeq \frac{1}{\sqrt{ \mintwo{d_k}{d_\l} }} \norm[F]{ \herm{\mU^{(k)}} \mU^{(\l)}  }.
\]
The relation between the affinity notions $\affp(\cdot)$ and $\aff(\cdot)$ is brought out by  expressing them in terms of the principal angles between  $\cS_k$ and $\cS_\l$ according to 
\begin{align}
\affp(\cS_k,\cS_\l) = \cos(\theta_1)
\label{eq:affppa}
\end{align}
and
\begin{align}
\aff(\cS_k,\cS_\l) = \frac{\sqrt{ \cos^2( \theta_1) + ...+ \cos^2(\theta_{ \mintwo{d_k}{d_\l} })}}{\sqrt{\mintwo{d_k}{d_\l} }}
\label{eq:affpa}
\end{align}
where $\theta_1, ... ,\theta_{ \mintwo{d_k}{d_\l}}$ with $0 \leq \theta_1 \leq ... \leq  \theta_{ \mintwo{d_k}{d_\l} }\leq \pi/2$ 
denotes the principal angles between $\cS_k$ and $\cS_\l$, defined as follows. 
\begin{definition}
The principal angles $\theta_1,...,\theta_{\mintwo{d_k}{d_\l}}$ between the subspaces $\cS_k$ and $\cS_\l$ are defined recursively according to
\[
\cos(\theta_j) = \innerprod{\vv_j}{\vu_j}, \text{ where } (\vv_j,\vu_j) = \arg  \max \innerprod{\vv}{\vu}
\]
with the maximization carried out over all $\vv \in \cS_k\colon \norm[2]{\vv}=1$,  $\vu \in \cS_\l\colon \norm[2]{\vu}=1$, subject to $\innerprod{\vv}{\vv_i}=0$ and $\innerprod{\vu}{\vu_i}=0$ for all $i=1,...,j-1$ (for $j=1$, this constraint is void). 
\end{definition}

Note that $0 \leq \aff(\cS_k,\cS_\l)  \leq \affp(\cS_k,\cS_\l) \leq 1$.  
If $\cS_k$ and $\cS_\l$ intersect in $p$ dimensions, i.e., if $\cS_k \cap \cS_\l$ is $p$-dimensional, then $\cos(\theta_1)=...=\cos(\theta_p)=1$ \cite{golub_matrix_1996}. 
Hence, if $\cS_k$ and $\cS_\l$ intersect in $p\geq 1$ dimensions, we have $\affp(\cS_k,\cS_\l) = 1$ and $\aff(\cS_k,\cS_\l) \geq \sqrt{p/ (\mintwo{d_k}{d_\l}})$. 
We finally note that the affinity notion \cite[Definition~2.6]{soltanolkotabi_geometric_2011} and \cite[Definition~1.2]{soltanolkotabi_robust_2013}, relevant to the analysis of SSC and RSSC, is equal to $\aff(\cdot,\cdot)$. 

We are now ready to state our first main result.

\begin{theorem}
Suppose that $\X_\l, \l \in [L]$, is obtained by choosing $n_\l$ points i.i.d.~uniformly from $\{\vx \in \cS_\l \colon \norm[2]{\vx} = 1 \}$, independently across $\l$, and let $\X = \X_1 \cup ...  \cup  \X_L$. Pick $\rho \in [0,1)$ 
and suppose that $n_\l\geq n_0$, for all $\l \in [L]$, where $n_0$ is a constant that depends on $d_{\max}$ and $\rho$ only. Pick $\gamma>1$ and suppose that $q \in [c_2 \gamma \log n_{\max},   n_{\min}^\rho]$ with $c_2 = 6 (12 \pi)^{\d_{\max}-1}$. If  
\begin{align}
\max_{k,\l \colon k\neq \l} \affp(\cS_k,\cS_\l) < 1
\label{eq:condTSCaffpmthm}
\end{align}
then TSC delivers the correct segmentation of $\X$ with probability at least
$
1 - \sum_{\l =1}^L \big(  n_\l e^{-c_1(n_\l - 1)}  \allowbreak+ 2{n_\l}^{-\gamma+1} \big)
$, where $c_1$ is a numerical constant. 
\label{thm:TSCprob}
\end{theorem}

Theorem~\ref{thm:TSCprob} states that TSC delivers the correct segmentation of $\X$ 
with high probability if the subspaces do not intersect (recall that $\affp(\cS_k,\cS_\l) = 1$ if and 
only if $\cS_k$ and $\cS_\l$ intersect in at least one dimension) 
and if $\X$ contains sufficiently many points from each subspace ($n_\l \geq n_0$, for all $\l \in [L]$). 
Intuitively we expect that clustering 
becomes easier when the $n_\l$ increase. To see that Theorem~\ref{thm:TSCprob} confirms this intuition, set $n_\l = n$, for all $\l\in [L]$, and note that the probability of correct segmentation in Theorem~\ref{thm:TSCprob} increases in $n$.

Theorem~\ref{thm:TSCprob} furthermore shows that 
TSC delivers the correct segmentation of $\X$ 
asymptotically  in the number of points in $\X$ from each subspace, $n_\l$, even when the $n_\l$ scale differently (in a sense made precise below), and/or the number of subspaces, $L$, grows faster than one or more of the $n_\l$. 
To see this, fix the $d_\l$, and let $n_\l = n^{\kappa_\l}$, $L= n^\kappa$ for numerical constants $\kappa_\l$ and $\kappa$ (possibly $\kappa > \kappa_\l$, in which case $L$ grows faster than $n_\l$), and let $n\to \infty$. Choose $\gamma$ such that $(\gamma-1)\kappa_{\min} > \kappa$ where $\kappa_{\min} \defeq \min_\l \kappa_\l$.  
 With $\kappa_{\max} = \max_\l \kappa_\l$, for $q \in [c_2 \gamma \kappa_{\max} \log n, \, n^{\kappa_{\min} \rho}]$ with $c_2$ and $\gamma$ from Theorem~\ref{thm:TSCprob} 
 (the interval is guaranteed to be nonempty for $n$ sufficiently large as $c_2$ does not depend on $n$) 
 it then follows 
that TSC yields correct segmentation with probability at least 
\begin{align*}
&1 - \sum_{\l =1}^L \left(  n^{\kappa_\l} e^{-c_1(n^{\kappa_\l} - 1)}  + 2n^{-(\gamma-1) \kappa_\l} \right)  \\
 &\hspace{0.9cm}\geq 1 -  \left(  n^{\kappa_{\min} + \kappa} e^{-c_1(n^{\kappa_{\min}} - 1)}  + 2 n^{-(\gamma-1) \kappa_{\min} + \kappa} \right)
\end{align*}
which tends to $1$ as $n \to \infty$.

The proof of Theorem  \ref{thm:TSCprob} is based on the realization that the graph $G$ 
is a random graph owing to the random data model. Specifically, the proof is effected 
by showing that the connected components in $G$ correspond to the $\X_\l$ with probability satisfying the probability estimate in Theorem \ref{thm:TSCprob}. 
As for the choice of $q$ in Theorem~\ref{thm:TSCprob}, the upper bound on $q$ is used to establish that $G$ has no false connections, i.e., each $\vx_j \in \X_\l$ is connected to points in $\X_\l$ only, for all $\l$. 
An upper bound on $q$ is also necessary as obviously $q > n_{\min}$ results in $G$ necessarily having false connections. 
The lower bound on $q$ is needed to ensure that, in addition, the subgraphs $G(\X_\l)$ corresponding to the $\X_\l$ are connected, and hence the 
$G(\X_\l)$
form connected components. 
In fact, the lower bound on $\q$ (as a function of $n_{\max}$) is order-wise necessary for the
$G(\X_\l)$
to be connected. Specifically, there exists a constant $c$ that does not depend on $n_\l$, 
such that for $q = c \log n_\l$, $G(\X_\l)$ 
is not  connected with probability $1$ as $n_\l \to \infty$ (not shown here). 
The exponential dependency of the constant $c_2 = 6 (12 \pi)^{\d_{\max}-1}$ on $d_{\max}$ requires that the $n_\l$ be exponential in the $d_{\l}$ as this is necessary for the interval $[c_2 \gamma \log n_{\max},   n_{\min}^\rho]$ of admissible values for $q$ to be non-empty. 
While this restricts the range of parameters $d_\l, n_\l$,  Theorem \ref{thm:TSCprob} applies to, the statement in Theorem \ref{thm:TSCprob} is strongest possible as it guarantees that the clustering error is zero as opposed to ensuring no false connections only. Zero clustering error ensures that \emph{every} point in the data set is clustered correctly.
We finally note that the exponential dependency of $c_2$ on $d_{\max}$ appears to be an artifact of our proof technique, as indicated by numerical results (not shown here). In fact, these numerical results suggest that $c_2$ may even be a decreasing function of  $d_{\max}$.

Theorem~\ref{thm:TSCprob} does not apply to subspaces that intersect as $\affp(\cS_k,\cS_\l)\!=\!1$ in this case. We can, however, find a statement analogous to Theorem~\ref{thm:TSCprob}, but in terms of $\aff(\cS_k, \cS_\l)$, which applies to intersecting subspaces.

\begin{theorem}
Suppose that $\X_\l, \l \in [L]$, is obtained by choosing $n_\l$ points i.i.d.~uniformly from $\{\vx \in \cS_\l \colon \norm[2]{\vx} = 1 \}$, independently across $\l$, and let $\X = \X_1 \cup ...  \cup  \X_L$.  
Suppose furthermore that $q \in [c_1 \log n_{\max},  n_{\min}/6]$ with $c_1 = 18 (12 \pi)^{\d_{\max}-1}$. If
\begin{align}
\max_{k,\l \colon k\neq \l} \aff(\cS_k,\cS_\l) \leq   \frac{1}{15 \log N }  
\label{eq:condTSCaff2}
\end{align}
then TSC delivers the correct segmentation of $\X$ with probability at least
$
1 - 10/N - \sum_{\l=1}^L  \allowbreak ( n_\l e^{-c(n_\l-1)} + 2n_\l^{-2} )
$,
where $c>0$ is a numerical constant.
\label{thm:aff2}
\end{theorem}

The interpretation of Theorem~\ref{thm:aff2} is analogous to that of Theorem~\ref{thm:TSCprob} with the important difference that the right hand side (RHS) of \eqref{eq:condTSCaff2}, as opposed to the RHS of \eqref{eq:condTSCaffpmthm}, decreases, albeit very slowly, in the $n_\l$ as $N = \sum_\l n_\l$. 
At first sight this is counterintuitive as we expect that clustering becomes easier when the number of points in each subspace increases. However, our statement guarantees that \emph{every} point in the data set is clustered correctly, even though the subspaces are allowed to intersect (cf.~\eqref{eq:condTSCaff2}). As the total number of points, $N$, increases, we would expect that the probability that at least \emph{one} point is close to an intersection of two subspaces, and therefore  misclustered, increases. 
Ensuring that the success probability increases in the $n_\l$, therefore leads to the affinity condition \eqref{eq:condTSCaff2} becoming stricter as $N$ increases.

Again, the exponential dependency of the constant $c_1 = 18 (12 \pi)^{\d_{\max}-1}$ on $d_{\max}$ requires that the $n_\l$ be exponential in the $d_{\l}$. If one is content with satisfying the (weaker) no false connections property only, this dependency on $d_{\max}$ vanishes by virtue of a lower bound on $\q$ not being needed. 
Specifically, this leads to the following result. 
\begin{corollary}
Suppose that $\X_\l, \l \in [L]$, is obtained by choosing $n_\l$ points i.i.d.~uniformly from $\{\vx \in \cS_\l \colon \norm[2]{\vx} = 1 \}$, independently across $\l$, and let $\X = \X_1 \cup ...  \cup  \X_L$. Suppose furthermore that $q \leq n_{\min}/6$. If
\[
\max_{k,\l\colon k\neq \l}  \aff(\cS_k,\cS_\l)  \leq \frac{1}{15  \log N}
\]
then $G$ has no false connections with probability at least
$
1 - \frac{10}{N} - \sum_{\l\in [L]} n_\l e^{-c(n_\l-1)},
$
where $c>0$ is a numerical constant. 
\label{cor:ofnoisycase}
\end{corollary}

Note that Corollary \ref{cor:ofnoisycase} does not require any relation between the $n_\ell$ and the $d_\ell$, 
in particular the $n_\l$ can be linear in the $d_\l$. 
At first sight this might seem surprising as nearest neighbor algorithms often suffer from the  \emph{curse of dimensionality} \cite{hastie_elements_2009}, 
manifested by the neighborhood of a point in a high-dimensional space no longer being local \cite{hastie_elements_2009}, e.g., the vast majority of points chosen i.i.d.~on a high-dimensional unit sphere are essentially orthogonal to each other. 
Although TSC is a nearest neighbor algorithm, it relies  only on the premise that the vectors close to a given data point $\vx_j$ also lie in the subspace $\vx_j$ lies in. This premise only requires the affinities between the subspaces $\cS_\l$ to be sufficiently small, does not rely on a certain relation between the $n_\ell$ and the $d_\ell$, and does not break down when the subspaces are high-dimensional. 
To see all this, we next provide a back-of-the-envelope argument establishing the no false connections property under a (dimension-independent) condition on the affinity of the subspaces. 
The proofs of the no false connections property in Theorems \ref{thm:TSCprob}--\ref{thm:fullyrandomnew} and Corollary \ref{cor:ofnoisycase} are essentially formal versions of the argument below.

For ease of exposition, we set $\d_\l = \d$, $n_\l = |\X_\l| = n$, for all $\l$, and we take the $\va_i^{(\l)}$ in $\vx_i^{(\l)} = \mU^{(\l)} \va_i^{(\l)}$ to be i.i.d.~$\mathcal N(\mathbf{0},(1/\d) \mI_\d)$ (recall that $\mU^{(\l)}$ is an orthonormal basis for $\cS_\l$). 
As the corresponding direction vectors $\vx_i^{(\l)}/ \big\| \vx_i^{(\l)} \big\|_2$ are distributed uniformly on {$\{\vx \in \cS_\l \colon \norm[2]{\vx} = 1 \}$} and $ \big\| \vx_i^{(\l)}\big\|^2_2 =  \big\| \va_i^{(\l)} \big\|^2_2$ concentrates around its expectation $\EX{\big\| \va_i^{(\l)} \big\|^2_2} = 1$, this model is conceptually equivalent to the $\va_i^{(\l)}$ being i.i.d.~on the unit sphere, as 
assumed in the formal statements throughout the paper. 
The program of the back-of-the-envelope calculation below is as follows. 
We use the fact that the absolute value of the inner product between data points from within a given subspace concentrates around $c/\sqrt{d}$, whereas the absolute value of the inner product between data points from different subspaces, $\cS_k$ and $\cS_\l$, concentrates around a value $\leq \affp(\cS_k,\cS_\l)  c / \sqrt{d}$. 
Thus, even when the subspace dimension $\d$ is large, 
the maximum inner product between data points from a given subspace will still be larger than the largest inner product between data points from different subspaces, provided that the affinity is sufficiently small. 
More formally, the no false connections property holds if for $\vx_i \in \X_\l$, the corresponding set $\S_i$ from Step 2 of the TSC algorithm corresponds to points in $\X_\l$ only, for all $\vx_i$, and for all $\l$. 
The set $\S_i$ contains indices corresponding to
points in $\X_\l$ only if the $q$th largest value in the set
$
\Big\{  \big| \big< \vx_j^{(\l)} ,  \vx_i^{(\l)} \big> \big|, j\neq i \Big\}
$
exceeds the largest value in the set $\Big\{\big| \big< \vx_j^{(k)} ,  \vx_i^{(\l)} \big>\big|, j, k\neq \l \Big\}$. 
Conditioned on $\va_i^{(\l)}$, the random variable 
$
\big< \vx_j^{(\l)} ,  \vx_i^{(\l)} \big> 
= 
\big< \va_j^{(\l)} ,  \va_i^{(\l)} \big>
$
is zero-mean Gaussian with variance $\big \| \va_i^{(\l)} \big \|^2_2/\d$. 
A standard result from order statistics \cite{david_order_2004} shows  
that, with high probability, the $q$th largest value in the set $\Big\{\big|\big< \vx_j^{(\l)} ,  \vx_i^{(\l)} \big>\big|, j\neq i \Big\}$  
is no larger than 
\begin{align}
c_1 \sqrt{\log(n/q)} \frac{\big \| \va_i^{(\l)} \big \|_2}{\sqrt{\d}}. 
 \label{eq:1df}
\end{align}
Next, consider data points 
$\vx_j^{(k)} , \vx_i^{(\l)}$ 
 from different subspaces (i.e., $k \neq \l$), and note that
\begin{align*}
\big|\big< \vx_j^{(k)} ,  \vx_i^{(\l)} \big>\big|
&=
\big|\big< \mU^{(k)}\va_j^{(k)} ,  \mU^{(\l)} \va_i^{(\l)} \big>\big| \\
&\leq 
\big|\big< \va_j^{(k)} ,   \va_i^{(\l)} \big>\big|   \norm[2]{ \transp{\mU^{(k)}} \mU^{(\l)} } \\
&=
\big|\big< \va_j^{(k)} ,   \va_i^{(\l)} \big>\big| \affp(\cS_k,\cS_\l).
\end{align*}
As before, $\big< \va_j^{(k)} ,   \va_i^{(\l)} \big>$ is Gaussian with variance $\big\|\va_i^{(\l)} \big\|^2_2 /\d$. 
Again, 
it follows that the largest value in the set $\Big\{ \big|\big< \vx_j^{(k)} ,  \vx_i^{(\l)} \big>\big|, j, k\neq \l \Big\}$ is smaller than 
\begin{align}
c_2 \sqrt{\log((L-1)n)}  \frac{\big\|\va_i^{(\l)} \big\|_2}{\sqrt{\d}}  \affp(\cS_k,\cS_\l) 
 \label{eq:2df}
\end{align}
with high probability. 
We can hence expect TSC to succeed (with high probability) if \eqref{eq:2df} is smaller than \eqref{eq:1df} which leads to 
\[
\affp(\cS_k,\cS_\l) 
\leq 
\frac{c_1 \sqrt{\log(n/q)} }{c_2 \sqrt{\log((L-1)n)} }.  
\]
The RHS of this condition does not depend on the dimension of the subspaces, $\d$, 
which explains why TSC does not suffer from the curse of dimensionality. 
Note that while $\affp(\cS_k,\cS_\l)$ does depend on $\d$ through the subspaces $\cS_k$ and $\cS_\l$, it can easily be small for large $\d$ (e.g., for orthogonal subspaces $\cS_k,\cS_\l$ of dimension $\d$ in $\reals^m, m = 2\d$, $\affp(\cS_k,\cS_\l) = 0$, or consider Lemma \ref{thm:fullyrandom} in Appendix \ref{app:thm:fullyrandomnew} for a more interesting example, which shows that for $L$ subspaces with random orientations $\affp(\cS_k,\cS_\l)$, for all pairs $\cS_k, \cS_\l, k\neq \l$, is close to zero with high probability provided that $m \geq O(d+\log L)$).

\section{\label{sec:noise}Impact of noise}

In many practical applications the data points to be clustered are corrupted by measurement noise, typically modeled as additive Gaussian noise. 
It is therefore of interest to analyze the performance of TSC applied to noisy data. 

\begin{theorem}
Suppose that $\X_\l, \l \in [L]$, is obtained by choosing $n_\l$ points corresponding to $\cS_\l$ at random according to $\vx_j^{(\l)} = \vy^{(\l)}_j  + \ve^{(\l)}_j  , j \in [n_\l]$, where the $\vy^{(\l)}_j$ are chosen i.i.d.~uniformly from $\{\vy \in \cS_\l \colon \norm[2]{\vy} = 1 \}$, independently across $\l$, and the $\ve^{(\l)}_j$ are i.i.d.~$\mathcal N( \mathbf 0, (\sigma^2/m)  \mI_m)$, independent of the $\vy^{(\l)}_j$. 
Let $\X = \X_1 \cup ...  \cup  \X_L$ and suppose that $q \leq n_{\min}/6$. If
\begin{align}
\max_{k,\l\colon k\neq \l}  \aff(\cS_k,\cS_\l)  +  \frac{\sigma(1+\sigma)}{\sqrt{\log N}} \frac{\sqrt{d_{\max}}}{\sqrt{m}}  \leq \frac{1}{15  \log N}
\label{eq:condthmnoisycase}
\end{align}
with $m \geq 6 \log N$, then $G$ has no false connections with probability at least
$
1 - \frac{10}{N} \allowbreak- \sum_{\l\in [L]} \allowbreak n_\l e^{-c(n_\l-1)},
$
where $c>0$ is a numerical constant. 
\label{thm:noisycase}
\end{theorem}

First, note that, unlike in the noiseless case, the data points $\vx_j^{(\l)}$ in Theorem~\ref{thm:noisycase} do not have unit norm. However, since $\ve^{(\l)}_j$ concentrates around its mean, the norms $\|\vx_j^{(\l)}\|_2$ are close to each other with high probability. TSC 
also applies to points that are unnormalized, with the only difference that  $\exp(-2\arccos(\left| \innerprod{\vx_j }{ \vx_i }\right|))$ in Step 2 has to be replaced by $\exp(-2\arccos(\left| \innerprod{\vx_j }{ \vx_i } \right| \allowbreak / (\norm[2]{\vx_j} \norm[2]{\vx_i}) ))$. 
Second, note that Theorem~\ref{thm:noisycase}, unlike the results in the noiseless case in Theorems \ref{thm:TSCprob} and \ref{thm:aff2} only ensures the absence of false connections in $G$ and hence does not guarantee zero clustering error. 
Theorem~\ref{thm:noisycase} states that TSC succeeds 
 (in the sense of $G$ having no false connections) 
with high probability if $\X$ contains sufficiently many points from each subspace (see the probability estimate in Theorem \ref{thm:noisycase}) and if the additive noise variance and the affinities between the subspaces are sufficiently small.  

Condition~\eqref{eq:condthmnoisycase} nicely reflects the intuition that the more distinct the orientations of the subspaces the more noise TSC tolerates. 
What is more, Condition~\eqref{eq:condthmnoisycase} reveals that TSC can succeed even under massive noise, i.e., even if $\sigma^2 = \EX{\big\| \ve^{(\l)}_j \big\|_2^2} > \norm[2]{\vy^{(\l)}_j}^2=1$, provided that the dimensions of the subspaces are sufficiently small relative to the ambient dimension. 

The intuition behind the factor $\sigma (1+\sigma) \sqrt{\d_{\max}/m}$ in \eqref{eq:condthmnoisycase}, made rigorous in the proof of Theorem~\ref{thm:noisycase}, is as follows. 
Assume, for simplicity, that $\d_\l = \d$, for all $\l$, and consider the most favorable situation of subspaces that are orthogonal to each other, i.e., $\aff(\cS_k,\cS_\l)= 0$, for all pairs $(k,\l)$ with $k\neq \l$. 
Recall that TSC relies on the inner products between points within a given subspace to typically be larger than the inner products between points in distinct subspaces. 
First, note that $\innerprod{\vx_j}{\vx_i} = \innerprod{\vy_j}{\vy_i} + \innerprod{\ve_j}{\ve_i} + \innerprod{\vy_j}{\ve_i} + \innerprod{\ve_j}{\vy_i}$. Then, under the statistical data model of Theorem \ref{thm:noisycase}, we have 
$
\left(\EX{\left| \innerprod{\vy_j}{\vy_i} \right|^2}\right)^{1/2} = \frac{1}{\sqrt{d}}
$
if $\vy_j, \vy_i \in \cS_\l$ and 
$
\innerprod{\vy_j}{\vy_i}  = 0
$
if $\vy_j \in \cS_k$ and $\vy_i \in \cS_\l$, with $k\neq \l$. 
If the terms $\innerprod{\ve_j}{\ve_i}$, $\innerprod{\vy_j}{\ve_i}$, and  $\innerprod{\ve_j}{\vy_i}$ are all small relative to $\frac{1}{\sqrt{d}}$, we have a margin on the order of $\frac{1}{\sqrt{d}}$ to distinguish pairs of points from within a given subspace from pairs of points from different subspaces. 
Indeed, $\innerprod{\vy_j}{\ve_i}$ and $\innerprod{\ve_j}{\vy_i}$ are small relative to $\frac{1}{\sqrt{d}}$ if $\frac{\sigma}{\sqrt{m}}$ is small relative to $\frac{1}{\sqrt{d}}$ (cf.~\eqref{eq:noiseinss}),  while $\frac{\sigma^2}{\sqrt{m}}$ being small relative to  $\frac{1}{\sqrt{d}}$ ensures that $\innerprod{\ve_j}{\ve_i}$ is small relative to $\frac{1}{\sqrt{d}}$ (cf.~\eqref{eq:noisenoisinnerpr}). 
These two conditions are obviously satisfied when $\sigma (1+\sigma) \sqrt{\d/m}$ is small.

\newcommand{\p}[0]{p}

\section{\label{sec:erasure}Incomplete data}

In practical applications the data points to be clustered are often incompletely observed, think of, e.g., images that exhibit scratches or have missing parts. It is therefore of significant interest to understand the impact of incomplete observations on the performance of TSC. Corresponding results for deterministic subspaces will necessarily depend on the specific orientations of the subspaces and will hence take on a form which makes it difficult to draw insightful conclusions. 
To make the problem analytically more tractable, we assume both the orientations of the subspaces as well as the data points in the subspaces to be random. Specifically, we will take the basis matrices $\mU^{(\l)}$ of the subspaces $\cS_\l$ to be i.i.d.~Gaussian random matrices, 
which ensures that each $\mU^{(\l)}$ is approximately orthonormal with high probability (rather than the $\mU^{(\l)}$ being strictly orthonormal as in the previous sections). 
For simplicity of exposition, throughout this section, we take the subspaces $\cS_\l$ to have equal dimension $d$ 
and let the number of points in each of the subspaces be $n$. 
We furthermore set the values of the unobserved entries in each data vector to zero and keep working in the original $m$-dimensional ambient space. As the TSC algorithm depends on inner products between the data points only this ensures that the missing observations will result in zero contributions.

\begin{theorem}
Suppose that $\X_\l$ is obtained by choosing $n$ points corresponding to $\cS_\l$ according to $\vx_j^{(\l)} = \mU^{(\l)} \va^{(\l)}_j, j \in [n]$, where the $\va^{(\l)}_j$ are i.i.d.~uniform on $\US{\d}$,  and set $\X = \X_1 \cup ...  \cup  \X_L$. Let the entries of the $\mU^{(\l)} \in \reals^{m\times d}$ be i.i.d.~$\mathcal N(0,1/m)$.
Pick $\rho \in [0,1)$ and suppose that $n \geq n_0$, where $n_0$ is a constant that depends on $d$ and $\rho$ only. 
Suppose furthermore that $q \leq n^\rho$, and  
assume that in each $\vx_j \in \X$ up to $s$ arbitrary entries (possibly different for different $\vx_j$) are unobserved, i.e., set to $0$. 
 If 
\begin{align}
m \geq 
3 c_4 d  + s\left( c_4 \log\left( \frac{me}{2s}\right) + c_3  \right)  + c_4\log L 
\label{eq:finalcondfullyrand}
\end{align}
then $G$ has no false connections with probability at least
$
1 - L n   e^{-c_1 (n-1)}
$,
where $c_1,c_2,c_3,c_4>0$ are numerical constants. If $s=0$, \eqref{eq:finalcondfullyrand} reduces to $m \geq 
3c_4 d  + c_4\log L  $. 

\label{thm:fullyrandomnew}
\end{theorem}

Theorem~\ref{thm:fullyrandomnew} shows that the number of missing entries in the data vectors is allowed to be (up to a log-factor) linear in the ambient dimension. 
We can furthermore conclude 
that TSC succeeds (in the sense of $G$ having no false connections) 
with high probability even when the dimensions of the subspaces are 
linear in the ambient dimension. 
This should, however, be taken with a grain of salt as the fully random subspace model ensures that the subspaces are approximately pairwise orthogonal with high probability, and hence the affinities between the subspaces are close to zero.

\section{\label{sec:detoutl}Outlier detection}

We discuss the noiseless and the noisy case separately as the corresponding outlier models differ slightly. Moreover,  the proof for the noiseless case is very simple and insightful and thus warrants individual presentation. 

\subsection{\label{sec:outlnoisefree}Noise-free case}

Outliers are data points that do not lie in one of the low-dimensional subspaces $\cS_\l$ and do not exhibit low-dimensional structure. Here, this is conceptualized by assuming random outliers distributed uniformly on $\US{\d}$, the unit sphere of $\reals^m$. As before, the inliers are assumed to be distributed uniformly on $\cS_\l \cap \US{\d_\l}$. 
The outlier detection criterion we employ 
is based on the following observation. 
The maximum inner product between an outlier and any other point (be it outlier or inlier) in $\X$  is, with high probability, smaller than 
$
c  \sqrt{\log N}/\sqrt{m},
$ 
as made rigorous in the proof of Theorem \ref{thm:outldete} below. 
We therefore classify $\vx_j$ as an outlier if 
\begin{align}
\max_{i \in [N] \setminus \{j\}} \left| \innerprod{\vx_i}{\vx_j}\right| <  c \sqrt{ \log N} / \sqrt{m}.
\label{eq:outldetrule}
\end{align} 
The maximum inner product between any point $\vx_j \in \X_\l$ and the points in $\X_\l \setminus \{\vx_j\}$ is unlikely to be smaller than $1/\sqrt{\d_{\max}}$, as formalized in the proof of Theorem \ref{thm:outldete}. 
Hence, an inlier is unlikely to be misclassified as an outlier if 
$
c  \sqrt{\log N}/\sqrt{m} \leq 1/\sqrt{\d_{\max}},
$
i.e., if $d_{\max}/m$ is sufficiently small relative to $1/\sqrt{\log N}$. The following result formalizes this insight. 

\begin{theorem}
Suppose that the set of outliers, $\O$, is obtained by choosing $N_0$ outliers i.i.d.~uniformly on $\US{m}$, and that 
$\X_\l, \l \in [L]$, is obtained by choosing $n_\l$ points i.i.d.~uniformly from $\{\vx \in \cS_\l \colon \norm[2]{\vx} = 1 \}$, independently across $\l$. 
Set $\X = \X_1 \cup ...  \cup  \X_L \cup  \O$ and 
declare $\vx_j \in \X$ to be an outlier if \eqref{eq:outldetrule} holds with $c = \sqrt{6}$. 
Then, with $N= N_0 + \sum_{\l} n_\l$, all outliers are detected with probability at least $1-2N_0/N^2$. Furthermore, 
provided that
\begin{align}
\frac{\d_{\max}}{m} \leq \frac{1}{6 \log N}
\label{eq:condoutldet}
\end{align}
no inlier in $\cS_\l$ is misclassified as an outlier 
with probability at least 
\begin{equation}
1- n_\l e^{-\frac{1}{2}\log\left(\frac{\pi}{2}\right) (n_\l-1)}.
\label{eq:probestioutldete}
\end{equation} 
\label{thm:outldete}
\end{theorem}
\vspace{-0.5cm}

Theorem \ref{thm:outldete} states that under Condition \eqref{eq:condoutldet} and provided that the set $\X$ contains sufficiently many points from each subspace (cf.~\eqref{eq:probestioutldete}), 
outlier detection succeeds with high probability, i.e., every outlier is detected and no inlier is misclassified as an outlier. 
Note that this result does not make any assumptions on the orientations of the subspaces $\cS_\l$.

Since \eqref{eq:condoutldet} can be rewritten as
$
N_0 \leq e^{\frac{m}{6\d_{\max}}} - \sum_{\l} n_\l,
$
it follows that outlier detection succeeds even if the number of outliers is exponential in $m/\d_{\max}$. 

Finally, note that the outlier detection rule \eqref{eq:outldetrule} is very natural as it simply classifies those points as outliers whose (spherical) distance to \emph{all} other points, and hence also to their individual nearest neighbors, is large. The scheme provably works as the nearest neighbor of each inlier is typically much closer than the nearest neighbor of each outlier. 
The idea of performing outlier detection based on nearest neighbor distance properties appeared previously e.g.~in \cite{brito_connectivity_1997} (not in the context of subspace clustering though), where outliers are detected  based on the connectivity properties of mutual\footnote{In a mutual $k$-nearest neighbor graph, the points $\vx_i$ and $\vx_j$ are connected if $\vx_i$ is among the $k$-nearest neighbors of $\vx_j$ \emph{and} $\vx_j$ is among the $k$-nearest neighbors of $\vx_i$.} nearest neighbor graphs.

\subsection{Noisy case}

We next consider outlier detection under additive noise on the data points. To keep the analysis simple, 
 we change the outlier model slightly. Specifically, we assume the outliers to be $\mathcal N(\mbf 0, (1/m) \mI_m )$ distributed. Conceptually, this outlier model is equivalent to the one used in Section \ref{sec:outlnoisefree}, as the directions of the outliers in the present model, i.e., $\vx_i/\norm[2]{\vx_i}$, are uniformly distributed on $\US{m}$, and $\norm[2]{\vx_i}$ concentrates 
around $1$. 
We furthermore normalize the (noisy) data points such that the norm of the inliers also concentrates around $1$. This guarantees that outlier detection is not trivially accomplished by exploiting differences in the norms between inliers and outliers.

\begin{theorem}
Suppose that the set of outliers, $\O$, is obtained by choosing $N_0$ outliers i.i.d. \linebreak $\mathcal N(\mathbf 0,(1/m) \mI_m)$, and that $\X_\l, \l \in [L]$, is obtained by choosing $n_\l$ points corresponding to $\cS_\l$ according to $\vx_j^{(\l)} = \frac{1}{\sqrt{1+\sigma^2}} \allowbreak \left( \vy^{(\l)}_j  + \ve^{(\l)}_j \right),\allowbreak  j \in [n_\l]$, 
where the $\vy^{(\l)}_j$ are chosen i.i.d.~uniformly from $\{\vy \in \cS_\l \colon \norm[2]{\vy} = 1 \}$, independently across $\l$, and the $\ve^{(\l)}_j$ are i.i.d.~$\mathcal N(\mathbf 0,\allowbreak(\sigma^2/m) \mI_m )$.  
Let $\X = \X_1 \cup ...  \cup  \X_L \cup  \O$ and declare $\vx_j \in \X$ to be an outlier if \eqref{eq:outldetrule} holds with $c=2.3 \sqrt{6}$. 
Then, with $N=  N_0 + \sum_l n_\l$, assuming $m \geq 6 \log N$, all outliers are detected with probability at least $1 -3 \frac{N_0}{N^2}$. Furthermore, 
provided that
\begin{align}
\frac{\d_{\max}}{m} \leq  \frac{c_1}{(1+\sigma^2)^2 \log N}
\label{eq:condnoisyoutldet}
\end{align}
where $c_1$ is a numerical constant, no inlier belonging to $\cS_\l$ is misclassified as an outlier with probability at least
\begin{align}
1-  n_\l e^{-\frac{1}{2}\log\left(\frac{\pi}{2}\right) (n_\l-1) }  \allowbreak-n_\l^2 \frac{7}{N^3}.
\label{eq:outldeoutldetnoisyc}
\end{align}
\label{thm:outldetnoisyc}
\end{theorem}
\vspace{-0.5cm}

Theorem \ref{thm:outldetnoisyc} 
 shows that outlier detection can succeed even under massive noise   
provided that $d_{\max}/m$ is sufficiently small. 

\section{\label{sec:comp}Comparison with SSC/RSSC and other algorithms}

As mentioned in the introduction, there are only a few subspace clustering algorithms that are both computationally tractable and succeed \emph{provably} under non-restrictive conditions. Notable exceptions are the SSC algorithm  \cite{elhamifar_sparse_2009,elhamifar_sparse_2013}, and for the noisy case the RSSC algorithm \cite{soltanolkotabi_robust_2013} (an algorithm analogous to the RSSC algorithm was also studied in \cite{elhamifar_sparse_2013}). Since our analytical performance results are in the spirit of those for SSC and RSSC in \cite{soltanolkotabi_geometric_2011,soltanolkotabi_robust_2013}---in particular we use the same statistical data model---we next compare our findings to those in \cite{soltanolkotabi_geometric_2011,soltanolkotabi_robust_2013}. 
Analytical performance guarantees for SSC in the fully deterministic case can be found in \cite{elhamifar_sparse_2013}.

While SSC and RSSC employ a ``global'' criterion for building the adjacency matrix $\mA$ by sparsely representing each data point in terms of all the other data points through $\ell_1$-minimization or Lasso, TSC is based on a ``local'' criterion, namely the comparison of inner products of pairs of data points. This makes TSC computationally much less demanding than SSC and RSSC, while, perhaps surprisingly, essentially sharing the \emph{analytical} performance guarantees 
of SSC and RSSC. The complexity savings may, however, come at the cost of actual performance. Specifically, while there are situations where TSC outperforms SSC, SSC outperforming TSC is more common, as will be seen in the numerical results in Section \ref{sec:numres}.  

Concerning analytical performance guarantees, for SSC in the noiseless case, a result along the lines of Theorem~\ref{thm:aff2} was reported in \cite[Theorem~2.8]{soltanolkotabi_geometric_2011}, with the corresponding clustering condition in \cite[Theorem~2.8]{soltanolkotabi_geometric_2011} being identical (up to constants and log-factors) to our condition \eqref{eq:condTSCaff2}. However, the statement in \cite[Theorem~2.8]{soltanolkotabi_geometric_2011} is weaker than that in Theorem~\ref{thm:aff2} as it does not pertain to the clustering error directly, but rather ensures no false connections only. 
To prove that the clustering error is zero, we additionally establish that the subgraphs corresponding to the $\X_\l$ are connected. 
As already mentioned, this requires a lower bound on $\q$, which entails that the $n_\l$ be exponential in the $d_\l$. 
While this restricts the range of parameters $d_\l, n_\l$ Theorems \ref{thm:TSCprob} and \ref{thm:aff2} apply to, the corresponding statements are strongest possible as they guarantee that the clustering error is zero as opposed to ensuring no false connections only. 
Again, as mentioned before, this exponential dependency appears to be an artifact of the proof technique we employ. 

In the noisy case for RSSC a result analogous to our Theorem \ref{thm:noisycase} was reported in \cite[Theorem~3.1]{soltanolkotabi_robust_2013}, with the corresponding clustering condition in \cite[Theorem~3.1]{soltanolkotabi_robust_2013} being identical (again up to constants and log-factors) to our condition \eqref{eq:condthmnoisycase} with $\sigma(1+\sigma)$ in \eqref{eq:condthmnoisycase} replaced by $\sigma$. 
We note, however, that \cite{soltanolkotabi_robust_2013} requires $\sigma$ to be bounded in the sense of $\sigma \leq c$, for some constant $c$, an assumption not needed in our case. If we take $\sigma$ to satisfy $\sigma\leq c$, the factor $\sigma(1+\sigma)$ in Condition~\eqref{eq:condthmnoisycase} above can be replaced by $\sigma (1+c)$ and we would get a clustering condition that is equivalent (again up to constants and log-factors) to that in \cite{soltanolkotabi_robust_2013}. 
A result concerning clustering of incompletely observed data paralleling our Theorem \ref{thm:fullyrandomnew} does not seem to be available for SSC. 
The outlier detection scheme proposed in \cite{soltanolkotabi_geometric_2011} in the context of SSC is based on the premise that outliers can not be represented sparsely in terms of the other data points. This scheme 
 succeeds (i.e., every outlier is detected and no inlier is misclassified as an outlier) with probability at least $1 
- N_0 e^{- c \frac{n}{\log N}}  
- 
\sum_{\ell=1}^L n_\ell e^{-\sqrt{d_\ell} \sqrt{n_\ell - 1} }$ under the condition 
$
N_0\leq \min\{e^{c \sqrt{m}}/m , m \min_\l \left(n_\l/d_\l\right)^{cm/\d_\l} \} - \sum_{\l=1}^L n_\l 
$, while our outlier detection scheme succeeds under Condition (9), i.e., $N_0 \leq e^{\frac{m}{6 d_{\max}}  } - \sum_{\ell=1}^L n_\ell$, with probability at least $1 - \frac{2N_0}{N^2}- \sum_{\ell=1}^L n_\l e^{-\frac{1}{2}\log\left(\frac{\pi}{2}\right) (n_\l-1)}$. 
For both algorithms the number of outliers can be exponential in $m/\d_{\max}$, the success probability  increases in the $n_\ell$, and the $n_\l$ can be linear in the $d_\l$.

In terms of input parameters, RSSC in \cite{soltanolkotabi_robust_2013} chooses the Lasso regularization parameter $\lambda$ in a data-driven fashion, which makes the algorithm essentially parameterless. TSC in contrast has $q$ as an input parameter. A variation of TSC with a data-driven choice of $q$ was proposed in \cite{heckel_neighborhood_2014} and shown to lead to performance guarantees that are essentially equivalent to those reported in this paper. 

A comparison of the analytical performance results for RSSC (in particular \cite[Theorem~3.1]{soltanolkotabi_robust_2013}) 
 to those for a number of representative subspace clustering algorithms such as generalized PCA (GPCA) \cite{vidal_generalized_2005}, K-flats \cite{tseng_nearest_2000}, and LRR \cite{liu_robust_2010}, can be found in \cite[Section~5]{soltanolkotabi_robust_2013}. This comparison also features computational complexity and robustness aspects. 
As the main \emph{analytical} performance results for TSC are structurally equivalent to those for SSC and RSSC the conclusions drawn in the comparison in \cite[Section~5]{soltanolkotabi_robust_2013} essentially carry over to TSC.

\section{\label{sec:numres}Numerical results}

We use the following performance metrics. 
\begin{itemize}
\item The clustering error (CE) measures the fraction of misclassified points and is defined as follows. 
Denote the estimate of the number of subspaces by $\hat L$.  
Let $\vc \in [L]^N$ and $\hat \vc \in [\hat L]^N$ be the original and estimated assignments of the points in $\X$ to the individual subspaces. The CE is then defined as
\[
\mathrm{CE}(\hat \vc,\vc) = \min_\pi \left( 1 - \frac{1}{N}\sum_{i=1}^N 1_{\{\pi(\hat c_i)  =  c_i \}} \right)
\]
where the minimum is taken over all assignments $\pi \colon [L] \to [\hat L]$ (for $\hat L=L$, $\pi$ is simply a permutation). Note that $\pi$ appears naturally in the definition of the CE as the specific cluster indices are irrelevant to the CE. The problem of finding the optimal assignment $\pi$ can be cast as finding the maximal matching of a weighted bipartite graph, 
which can be solved efficiently via the Hungarian algorithm \cite{topchy_analysis_2004}. 

\item The error in estimating the number of subspaces $L$ is denoted as EL and takes the value $0$ if the estimate $\hat L$ is correct, $1$ if $L<\hat L$, and $-1$ if $L> \hat L$. We employ a signed error measure so as to be able to discriminate between under- and overestimation. 
In principle, EL averaged over problem instances, may therefore equal zero, while $\hat L \neq L$ for each individual problem instance. However, as it turns out (in the numerical results below), for a given choice of problem parameters, we get that either $L<\hat L$ or $L> \hat L$ almost consistently. 

\item The feature detection error (FDE) (for a given adjacency matrix $\mA$) is defined as
\[
\mathrm{FDE}(\mA) = 1 - \frac{1}{N} \sum_{i=1}^N \frac{ \norm[2]{\vb_{\vx_i}}}{\norm[2]{\vb_i} }  
\label{eq:fde}
\]
where $\vb_i$ is the $i$th column of the $N\times N$ adjacency matrix $\mA$ and $\vb_{\vx_i}$ is the vector containing the entries of $\vb_{i}$ corresponding to the set $\X_\l$ the data point $\vx_i$ lives in. 
The FDE measures to which extent points from different subspaces are connected in the graph $G$ (with adjacency matrix $\mA$), 
and equals zero if $G$ has no false connections. 
\end{itemize}

Throughout this section, we set $q = \max(3, \lceil  n/20\rceil)$ if the correct $L$ is provided
 to TSC, and $q = 2 \max(3, \lceil  n/20\rceil)$ if $L$ is estimated according to the eigengap heuristic. 
Matlab code to reproduce the results in this section is available at http://www.nari.ee.ethz.ch/commth/research/. 

\subsection{\label{sec:syntdata}Synthetic data}

Throughout Section \ref{sec:syntdata}, 
unless explicitly stated otherwise, we take 
$n_\l = n$ and $d_\l = d$, for all $\l$, and 
 generate the $d$-dimensional subspaces $\cS_\l$ by drawing i.i.d.~orthonormal basis matrices $\mU^{(\l)} \in \reals^{m\times d}$ uniformly at random from the set of all orthonormal matrices in $\reals^{m\times d}$.

\subsubsection{Intersection of subspaces}

\begin{figure}
\centering
\begin{tikzpicture}[scale=1] 
\begin{groupplot}[group style={group size=3 by 1,horizontal sep=0.8cm,vertical sep=1.2cm,xlabels at=edge bottom, ylabels at=edge left},
width=0.35\textwidth,/tikz/font=\small]
    \nextgroupplot[title={FDE}]
	\addplot +[mark=none,solid,DarkBlue] table[x index=0,y index=1]{./fig/data/FDE_SSintersect_TSC.dat}; 
	 \nextgroupplot[title={CE}]
	\addplot +[mark=none,solid,DarkBlue] table[x index=0,y index=1]{./fig/data/CE_SSintersect_TSC.dat}; 
	 \nextgroupplot[title={EL}]
	\addplot +[mark=none,solid,DarkBlue] table[x index=0,y index=1]{./fig/data/LS_SSintersect_TSC.dat}; 	
  \end{groupplot}
\end{tikzpicture}
\caption{\label{fig:intersect} 
Clustering error metrics as a function of the dimension of the intersection, $t$, for clustering points taken from two $10$-dimensional subspaces of $\reals^{200}$. 
}
\end{figure}
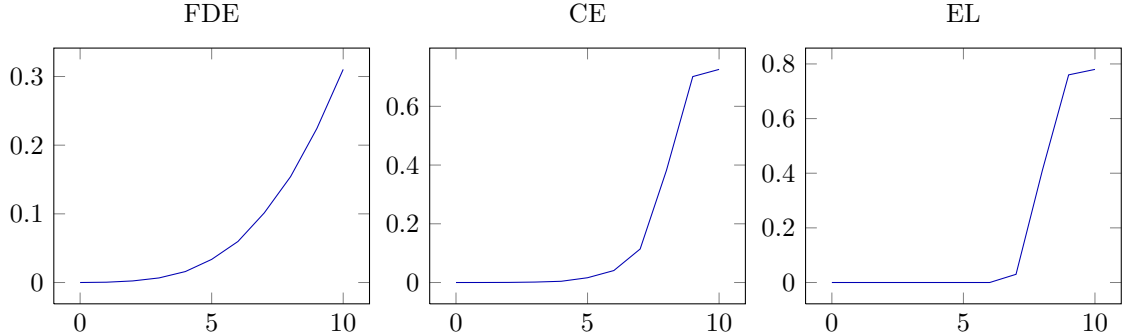

We next demonstrate that, as predicted by Theorem~\ref{thm:aff2}, TSC can succeed even when the subspaces $\cS_\l$ intersect. 
In order to facilitate comparison to SSC, we perform the same experiment as in \cite[Sec.~5.1.2]{soltanolkotabi_geometric_2011}. 
Specifically, we set $m=200$, $d=10$, and generate two subspaces, $\cS_1$ and $\cS_2$, at random through their defining bases $\mU^{(1)}$ and $\mU^{(2)}$ obtained as follows.
We choose, uniformly at random, from the set of all sets of $2d-t$ orthonormal vectors in $\reals^m$, a set of $2d-t$ orthonormal vectors, and identify the columns of $\mU^{(1)}$ and $\mU^{(2)}$ with the first and last $d$ of these vectors, respectively. This ensures that the intersection of $\cS_1$ and $\cS_2$ is at least of dimension $t$. Next, we generate $n=20d$ data points in each of the two subspaces according to $\vx_i^{(\l)} = \mU^{(\l)} \va_i^{(\l)}$, with the $\va_i^{(\l)}$ drawn i.i.d.~uniformly on $\US{d}$. For each $t = 0,...,d$ the CE, EL, and FDE are obtained by averaging over $100$ problem instances. 
From the results, shown in Figure~\ref{fig:intersect}, we can conclude that, as long as the dimension of the intersection of the subspaces is not too large, TSC does, indeed, yield a CE close to zero. The same experiment was performed for SSC in \cite[Sec.~5.1.2]{soltanolkotabi_geometric_2011} and delivered slightly better results.

\subsubsection{\label{sec:nriod} Influence of $d$, $n$, and incomplete data} 

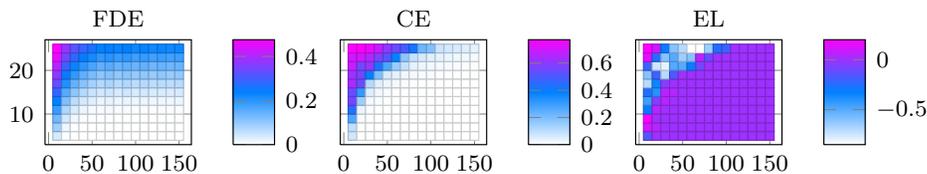
\begin{figure}
\centering
{
\pgfplotsset{every axis title/.append style={at={(0.5,0.85)}}}
\begin{tikzpicture}[scale=1.1,font=\scriptsize] 
\begin{groupplot}[group style={group size=3 by 1,horizontal sep=1.8cm,vertical sep=0.75cm,xlabels at=edge bottom, ylabels at=edge left,yticklabels at=edge left},
width=3.35cm,height=2.85cm,
colorbar]

    \nextgroupplot[title={FDE }]
	 \addplot[mark=square*,only marks, scatter, scatter src=explicit,
	 mark size=1.5,domain=0:1]
	  file {./fig/data_rev/FDE_varyd.dat};
	    
  \nextgroupplot[title={CE},colorbar]
	 \addplot[mark=square*,only marks, scatter, scatter src=explicit,
	 mark size=1.5]
 file {./fig/data_rev/CE_varyd.dat};
 	
\nextgroupplot[title={EL}, colorbar]
	  \addplot[mark=square*,only marks, scatter, scatter src=explicit,
  mark size=1.5]
   file {./fig/data_rev/LS_varyd.dat};
  \end{groupplot}
\end{tikzpicture}
}
\caption{\label{fig:varyd}
Clustering error metrics as a function of the dimension, $d$, of the subspaces on the vertical and the number of points taken from each subspace, $n$, on the horizontal axis, for $L=10$ subspaces of $\reals^{50}$. 
}

\end{figure}

The goal of the next experiment is to elicit the impact of $d$, $n$, and the number of missing entries in the data points on clustering performance, and to furthermore demonstrate that TSC can succeed even when $G$ has false connections.  
We generate $L=10$ subspaces of $\reals^{50}$, and vary their dimension $d$ and the number $n$ of points taken from each subspace. 
The individual data points are chosen according to the statistical model Theorem~\ref{thm:aff2} is based on.  
For each pair $(d,n)$, the FDE, CE, and EL are obtained by averaging over 20 problem instances. The results, depicted in Figure~\ref{fig:varyd}, show, as indicated in Section~\ref{sec:tsc}, that TSC can, indeed, succeed even when $G$ has false connections (i.e., when the FDE is non-zero). 

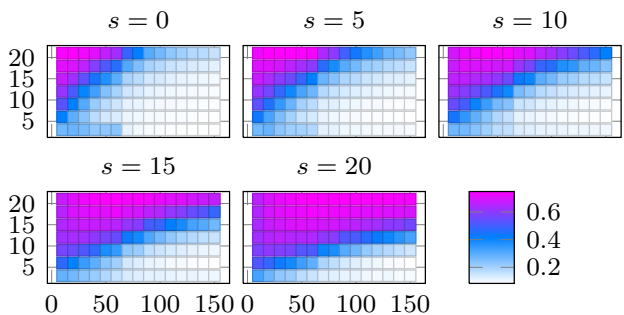
\begin{figure}
\centering
{
\pgfplotsset{every axis title/.append style={at={(0.5,0.85)}}}
\begin{tikzpicture}[scale=1.2,font=\scriptsize] 
\begin{groupplot}[group style={group size=3 by 2,horizontal sep=0.15cm,vertical sep=0.6cm,xlabels at=edge bottom, ylabels at=edge left,xticklabels at=edge bottom,yticklabels at=edge left},
width=3.6cm,height=2.6cm,
]
]
  \nextgroupplot[title={$s=0$}]
	 \addplot[mark=square*,only marks, scatter, scatter src=explicit,
	 mark size=1.8]
	 file {./fig/data_rev/erasure_0.dat};
	 	  
  \nextgroupplot[title={$s=5$}]
	 \addplot[mark=square*,only marks, scatter, scatter src=explicit,
	 mark size=1.8]
	 file {./fig/data_rev/erasure_5.dat};	
  
  \nextgroupplot[title={$s=10$}]
	 \addplot[mark=square*,only marks, scatter, scatter src=explicit,
	 mark size=1.8]
	 file {./fig/data_rev/erasure_10.dat};

  \nextgroupplot[title={$s=15$}]
	 \addplot[mark=square*,only marks, scatter, scatter src=explicit,
	 mark size=1.8]
	 file {./fig/data_rev/erasure_15.dat};
	 
  \nextgroupplot[title={$s=20$},colorbar]
	 \addplot[mark=square*,only marks, scatter, scatter src=explicit,
	 mark size=1.8]
	 file {./fig/data_rev/erasure_20.dat};

  \end{groupplot}
\end{tikzpicture}
}
\caption{\label{fig:varydsc}
Clustering error as a function of the dimension, $d$, of the subspaces on the vertical and 
the number of points taken from each subspace, $n$,  on the horizontal axis for $s$ missing entries in the data vectors. The results are for $L=6$ subspaces $\cS_\l$ of $\reals^{50}$, with $\aff(\cS_k,\cS_\l) \geq 1/\sqrt{3}$ for all $k,\l \in [L], k\neq \l$. 
}
\end{figure}

Next, we generate $L=6$ subspaces of $\reals^{50}$ by choosing their defining bases $\mU^{(\l)}$ as follows. We first draw $\mU\in \reals^{m\times d/3}$ (we restrict $d$ to integer multiples of $3$) uniformly from the set of all orthonormal matrices in $\reals^{m\times d/3}$. 
Then, we choose $\tilde \mU^{(\l)} \in \reals^{m\times 2d/3}, \l \in [L]$, independently across $\l$ and independently of $\mU$,  uniformly at random from the set of all orthonormal matrices in $\reals^{m\times 2d/3}$ that are orthogonal to $\mU$, and set $\mU^{(\l)} = [\tilde \mU^{(\l)}  \; \mU] \in \reals^{m\times d}$. 
 This ensures that the subspaces $\cS_\l$ with basis matrices $\mU^{(\l)}$ intersect in at least $d/3$ dimensions and hence $\aff(\cS_k,\cS_\l) \geq 1/\sqrt{3}$ for all $k,\l \in [L], k\neq \l$.  
The data points are chosen according to the statistical model Theorem~\ref{thm:aff2} is based on. For each data point $\vx_i$, we set the entries of $\vx_i$ with indices in $\D_i$ to zero, where the sets $\D_i$ are chosen independently and
uniformly at random from the set $\{\D\subseteq [m]\colon |\D|=s\}$. 
The results, 
summarized in Figure~\ref{fig:varydsc}, show that TSC can succeed even when a large fraction of the entries in each data vector is missing.

\subsubsection{Additive noise}

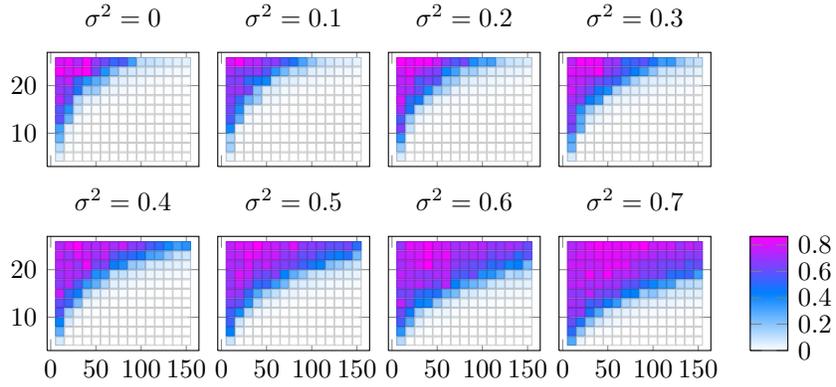
\begin{figure}
\centering
\begin{tikzpicture}[scale=1] 
\begin{groupplot}[group style={group size=4 by 2,horizontal sep=0.25cm,vertical sep=0.93cm,xlabels at=edge bottom, ylabels at=edge left,xticklabels at=edge bottom,yticklabels at=edge left},
width=3.6cm,height=3.1cm,/tikz/font=\small],point meta min = 0, point meta max=0.8]
  
  \nextgroupplot[title={$\sigma^2=0$}]
	 \addplot[mark=square*,only marks, scatter, scatter src=explicit,
	 mark size=1.6]
	 file {./fig/data_rev/CE_sig0.dat};
	 	  
  \nextgroupplot[title={$\sigma^2=0.1$}]
	 \addplot[mark=square*,only marks, scatter, scatter src=explicit,
	 mark size=1.6]
	 file {./fig/data_rev/CE_sig01.dat};	
  
  \nextgroupplot[title={$\sigma^2=0.2$}]
	 \addplot[mark=square*,only marks, scatter, scatter src=explicit,
	 mark size=1.6]
	 file {./fig/data_rev/CE_sig02.dat};

  \nextgroupplot[title={$\sigma^2=0.3$}]
	 \addplot[mark=square*,only marks, scatter, scatter src=explicit,
	 mark size=1.6]
	 file {./fig/data_rev/CE_sig03.dat};
	 
  \nextgroupplot[title={$\sigma^2=0.4$}]
	 \addplot[mark=square*,only marks, scatter, scatter src=explicit,
	 mark size=1.6]
	 file {./fig/data_rev/CE_sig04.dat};
	 
  \nextgroupplot[title={$\sigma^2=0.5$}]
	 \addplot[mark=square*,only marks, scatter, scatter src=explicit,
	 mark size=1.6]
	 file {./fig/data_rev/CE_sig05.dat};
	 
  \nextgroupplot[title={$\sigma^2=0.6$}]
	 \addplot[mark=square*,only marks, scatter, scatter src=explicit,
	 mark size=1.6]
	 file {./fig/data_rev/CE_sig06.dat};
	 
  \nextgroupplot[title={$\sigma^2=0.7$}, colorbar]
	 \addplot[mark=square*,only marks, scatter, scatter src=explicit,
	 mark size=1.6,colorbar=true]
	 file {./fig/data_rev/CE_sig07.dat};
	 	
  \end{groupplot}
\end{tikzpicture}
\vspace{-0.1cm}
\caption{\label{fig:noise}
Clustering error for data points taken from $L=10$  subspaces of $\reals^{50}$ corrupted by additive Gaussian noise, as a function of the dimension, $d$, of the subspaces on the vertical and 
the number of points taken from each subspace, $n$,  on the horizontal axis for different noise variances $\sigma^2$.
}
\end{figure}

We generate $L=10$ subspaces of $\reals^{50}$ and vary their dimension $d$ and the number of points $n$ taken from each subspace. The data points are subjected to additive noise before clustering. 
Specifically, we use the statistical data model Theorem~\ref{thm:noisycase} is based on. 
The results, depicted in Figure~\ref{fig:noise}, show that TSC can succeed even when the noise variance is large.

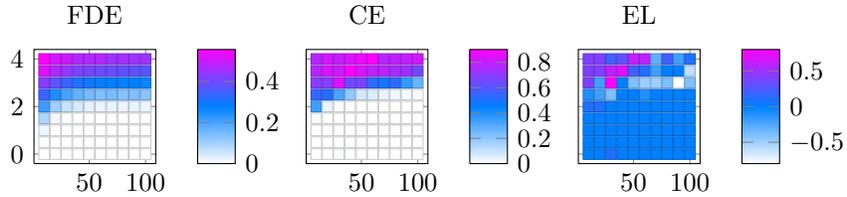
\begin{figure}
\centering
{
\begin{tikzpicture}[scale=1] 
\begin{groupplot}[group style={group size=3 by 1,horizontal sep=2cm,vertical sep=0.0cm,xlabels at=edge bottom, ylabels at=edge left,xticklabels at=edge bottom,yticklabels at=edge left},
width=3.2cm,height=3.1cm,/tikz/font=\small],point meta min = 0, point meta max=0.8]

  \nextgroupplot[title={FDE}, colorbar]
	 \addplot[mark=square*,only marks, scatter, scatter src=explicit,
	 mark size=2,colorbar=true]
	 file {./fig/data_rev/FDE_huge_noise.dat};
	   
	   \nextgroupplot[title={CE}, colorbar]
	 \addplot[mark=square*,only marks, scatter, scatter src=explicit,
	 mark size=2,colorbar=true]
	 file {./fig/data_rev/CE_huge_noise.dat};
	 
	 \nextgroupplot[title={EL}, colorbar]
	 \addplot[mark=square*,only marks, scatter, scatter src=explicit,
	 mark size=2,colorbar=true]
	 file {./fig/data_rev/LS_huge_noise.dat};
	 	
  \end{groupplot}
\end{tikzpicture}
}
\vspace{-0.3cm}
\caption{\label{fig:hugenoise}
Clustering error metrics for points taken from $L=5$ subspaces of $\reals^{400}$, each of which is $5$-dimensional, corrupted by additive Gaussian noise, as a function of the noise variance, $\sigma^2$, on the vertical and 
the number of points taken from each subspace, $n$,  on the horizontal axis.
}
\end{figure}

In Section~\ref{sec:noise}, we found that TSC can succeed even under massive noise (i.e., if $\sigma^2>1$), provided that $d/m$ is sufficiently small. To demonstrate this effect numerically, we generate $L=5$ subspaces in $\reals^{400}$, each of dimension $d=5$ (hence $d/m = 1/80$), and we choose the data points again according to the statistical model Theorem~\ref{thm:noisycase} is based on. 
We vary the number of points in each subspace, $n$, and the noise variance $\sigma^2$. 
 The corresponding results, depicted in Figure~\ref{fig:hugenoise}, confirm the analytical predictions of Theorem \ref{thm:noisycase}. 

\subsubsection{Detection of outliers}

In order to facilitate comparison with the outlier detection scheme proposed for SSC in \cite{soltanolkotabi_geometric_2011}, we perform our experiment with exactly the same parameters as used in \cite[Sec. 5.2]{soltanolkotabi_geometric_2011}. 
Specifically, we set $d=5$, vary $m \in \{50,100,200\}$, and generate $L=2 m/d$ subspaces  at random. 
We choose $n$ inliers per subspace and a total of $N_0 = Ln$ outliers according to the statistical model Theorem \ref{thm:outldete} is based on. The number of outliers is hence equal to the total number of inliers. We measure performance in terms of the misclassification error, defined as the number of misclassified points (i.e., outliers misclassified as inliers and inliers misclassified as outliers) divided by the total number of points in $\X$.   
We find a misclassification error of $\{0.017, 1.5 10^{-4}, 2.5 10^{-5}\}$ for $m=\{50,100,200\}$, respectively. The performance reported for SSC in \cite{soltanolkotabi_geometric_2011} is similar.

\subsection{\label{sec:clusMNIST}Clustering handwritten digits}

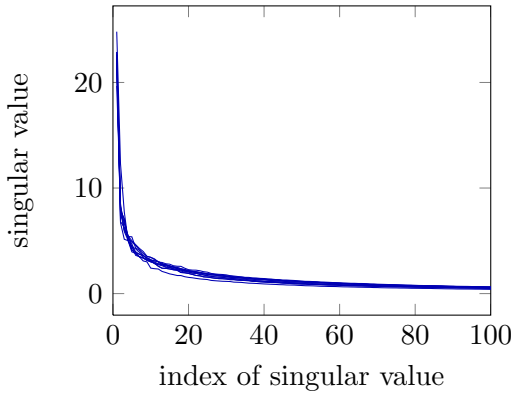
\begin{figure}
\centering
\begin{tikzpicture}[scale=1] 
    \begin{axis}[
        	xlabel=index of singular value,
	ylabel=singular value,
	xmin = 0, xmax=100,
	width=0.4\textwidth,
	]
	\addplot +[DarkBlue,mark=none,solid] table[x index=0,y index=1]{./fig/data/sval.dat}; 
	\addplot +[DarkBlue,mark=none,solid] table[x index=0,y index=1]{./fig/data/sval.dat};
	\addplot +[DarkBlue,mark=none,solid] table[x index=0,y index=2]{./fig/data/sval.dat};
	\addplot +[DarkBlue,mark=none,solid] table[x index=0,y index=3]{./fig/data/sval.dat};
	\addplot +[DarkBlue,mark=none,solid] table[x index=0,y index=4]{./fig/data/sval.dat};
	\addplot +[DarkBlue,mark=none,solid] table[x index=0,y index=5]{./fig/data/sval.dat};
	\addplot +[DarkBlue,mark=none,solid] table[x index=0,y index=6]{./fig/data/sval.dat};
	\addplot +[DarkBlue,mark=none,solid] table[x index=0,y index=7]{./fig/data/sval.dat};
	\addplot +[DarkBlue,mark=none,solid] table[x index=0,y index=8]{./fig/data/sval.dat};
	\addplot +[DarkBlue,mark=none,solid] table[x index=0,y index=9]{./fig/data/sval.dat};
	\end{axis}
\end{tikzpicture}
\caption{\label{fig:svals} Singular values of the matrices with columns corresponding to the vectorized images of a given digit from the MNIST data base.}
\end{figure}

\begin{figure}
\centering
\includegraphics{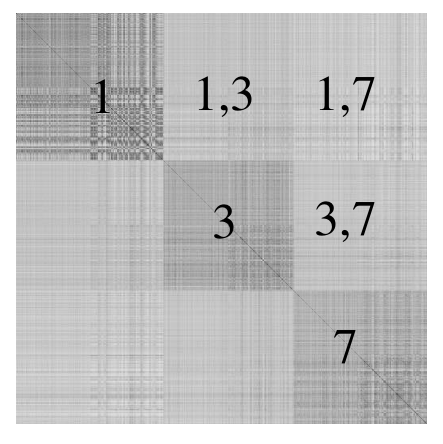}
\caption{\label{fig:adjacdig} 
Matrix with entries $\mA_{ij} = \exp(-\arccos(\left| \innerprod{\vx_j }{ \vx_i }\right|))$ for all pairs $\vx_j, \vx_i$ of vectorized images of the digits $\{1,3,7\}$ from the MNIST data base. }
\end{figure}

We next apply TSC to the problem of clustering handwritten digits. Specifically, we work with the MNIST test data set \cite{mnist_2013}  
that contains 10,000 centered $28\times 28$ pixel images of handwritten digits. 
The assumption underlying the idea of posing this problem as a subspace clustering problem is that the vectorized images of the different handwritten versions of a single digit lie approximately in a low-dimensional subspace \cite{hastie_metrics_1998}. To validate this assumption, we compute the singular values of the 
matrices $\mX_\l$ with columns corresponding to the vectorized images of the $\l$th digit, $\l = 0,1,...,9$, and sort them in descending order.  The results, plotted in Figure \ref{fig:svals}, show that the singular values of the matrices $\mX_\l$, indeed, decay to zero rapidly ($m=784$). 
As mentioned in Section \ref{sec:tsc}, TSC is built on the premise that the vectors close to $\vx_j$ in terms of the distance $\tilde \s (\vx_i,\!\vx_j) =   \arccos(\left|\innerprod{\vx_i}{\vx_j}\right|)$ also lie in the subspace $\vx_j$ lies in. 
As our analytical results in Sections \ref{sec:detss} and \ref{sec:noise} show, this premise is met (with high probability) for the statistical data model used throughout the paper. 
To see whether the premise is also met in practice, we  compute $\exp(-\tilde \s (\vx_i,\vx_j))$ for all pairs $\vx_j, \vx_i$ of vectorized images of the digits $\{1,3,7\}$ from the MNIST data set. In other words, we compute the adjacency matrix for $q = N$. 
The results, depicted in Figure \ref{fig:adjacdig}, show that, indeed, $\exp(-\tilde \s (\vx_i,\vx_j))$ for $\vx_i, \vx_j$ coming from the same digit is typically larger than for $\vx_i,\vx_j$ coming from  different digits.

\begin{figure}
\begin{center}
\pgfplotsset{
    /tikz/every mark/.append style={solid}
}
\pgfplotsset{
    /pgfplots/error/.append style={solid}
}

\begin{tikzpicture}

\begin{groupplot}[group style={group size=2 by 1,horizontal sep=2cm,vertical sep=0.2cm,xlabels at=edge bottom, ylabels at=edge left, xticklabels at=edge bottom,},
xlabel=number of points of each digit $n$,
width=0.48\textwidth,/tikz/font=\small]

    \nextgroupplot[ylabel={CE}]

        \addplot[mark=|,color=violet,dashed,error bars/.cd,
    y dir=both,y explicit,    error bar style={solid,violet},] coordinates {
(25,1.322667e-01) +- (0,8.565176e-02)
(50,8.843333e-02) +- (0,6.017832e-02)
(75,6.071111e-02) +- (0,2.872157e-02)
(100,5.273333e-02) +- (0,1.798104e-02)
(125,4.840000e-02) +- (0,1.599719e-02)
(150,4.317778e-02) +- (0,1.383008e-02)
(175,4.152381e-02) +- (0,1.095039e-02)
(200,3.928333e-02) +- (0,1.002873e-02)
(225,3.734815e-02) +- (0,8.052136e-03)
(250,3.549333e-02) +- (0,8.108756e-03)
(275,3.597576e-02) +- (0,7.237708e-03)
(300,3.546667e-02) +- (0,7.282927e-03)
(325,3.412308e-02) +- (0,7.184451e-03)
(350,3.276190e-02) +- (0,6.118462e-03)
(375,3.178667e-02) +- (0,6.827241e-03)
}; 
\addlegendentry{SSC}
                \addplot[blue,mark=star,color=black,dotted,error bars/.cd,
    y dir=both,y explicit,
    error bar style={solid,blue},
                ]    
                 coordinates {
(25,1.598667e-01) +- (0,1.126359e-01)
(50,1.179000e-01) +- (0,1.191981e-01)
(75,7.355556e-02) +- (0,8.332862e-02)
(100,4.430000e-02) +- (0,2.580378e-02)
(125,3.901333e-02) +- (0,1.639317e-02)
(150,3.517778e-02) +- (0,1.497241e-02)
(175,2.996190e-02) +- (0,9.007068e-03)
(200,2.948333e-02) +- (0,7.762232e-03)
(225,2.774815e-02) +- (0,7.603408e-03)
(250,2.672000e-02) +- (0,6.479307e-03)
(275,2.632727e-02) +- (0,6.219121e-03)
(300,2.531111e-02) +- (0,5.064936e-03)
(325,2.573333e-02) +- (0,5.058516e-03)
(350,2.508571e-02) +- (0,4.636581e-03)
(375,2.400889e-02) +- (0,4.426442e-03)
}; 
\addlegendentry{TSC}

                \addplot[color=black,solid,mark=x,error bars/.cd,
    y dir=both,y explicit,
    error bar style={solid,black},
                ]    
                 coordinates {
(25,1.636000e-01) +- (0,1.184995e-01)
(50,1.265333e-01) +- (0,1.217699e-01)
(75,7.173333e-02) +- (0,7.583633e-02)
(100,5.013333e-02) +- (0,4.990883e-02)
(125,4.085333e-02) +- (0,1.750449e-02)
(150,3.422222e-02) +- (0,1.330712e-02)
(175,3.040000e-02) +- (0,1.027324e-02)
(200,2.935000e-02) +- (0,8.674482e-03)
(225,2.708148e-02) +- (0,8.037765e-03)
(250,2.633333e-02) +- (0,6.994948e-03)
(275,2.601212e-02) +- (0,6.116473e-03)
(300,2.512222e-02) +- (0,5.350643e-03)
(325,2.568205e-02) +- (0,5.976767e-03)
(350,2.484762e-02) +- (0,4.914524e-03)
(375,2.389333e-02) +- (0,4.746259e-03)
}; 
\addlegendentry{LS-TSC}

    \nextgroupplot[ylabel=FDE]

\addplot +[mark=none,color=violet,dashed,error bars/.cd, y dir=both,y explicit,]
    table[x index=0,y index=1,y error index=2]{./fig/FDE_SSC.dat};

\addplot +[mark=none,color=blue,dotted,error bars/.cd, y dir=both,y explicit,]
    table[x index=0,y index=1,y error index=2]{./fig/FDE_TSCa.dat};

\addplot +[mark=none,color=black,solid,error bars/.cd, y dir=both,y explicit,]
    table[x index=0,y index=1,y error index=2]{./fig/FDE_TSCb.dat};

  \end{groupplot}  
\end{tikzpicture}  
\end{center}

\caption{\label{fig:compssctsc}  Empirical mean and standard deviation of the CE and FDE for handwritten digits from the MNIST data base.}
\end{figure}
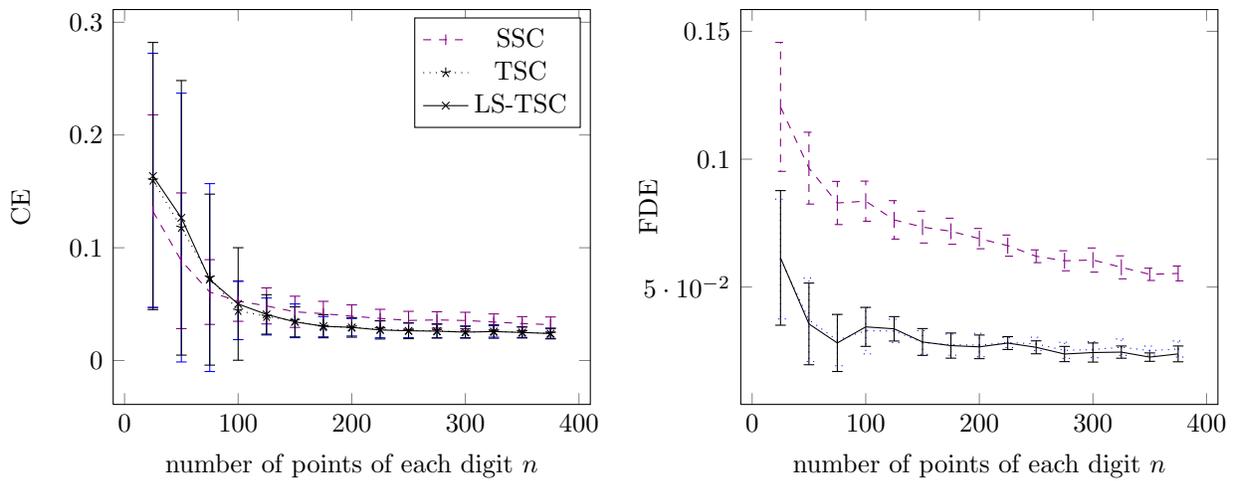

We compare the performance of TSC, LS-TSC, and SSC/RSSC. For SSC, we use the implementation from \cite{elhamifar_sparse_2013}. 
The empirical mean and variance of the CE are computed by averaging over 100 of the following problem instances. 
We choose the digits $\{2,4,8\}$ and for each digit we choose $n$ images uniformly at random from the set of all images of that digit. 
The results, summarized in Figure~\ref{fig:compssctsc}, show that SSC performs better than both TSC and LS-TSC when the data set contains few ($n\lesssim 80$) 
images of each digit, TSC and LS-TSC outperform  SSC when the data set contains many ($n\gtrsim 80$) images of each digit. 
Note that the FDE for TSC is significantly smaller than that for SSC, even in the regime $n\lesssim 80$ where SSC performs better. 
This is a result of $q$ being small, which yields  a sparse adjacency matrix for TSC, and therefore increases the chance of the nonzero entries, indeed, corresponding to points within the same subspace.


\subsection{Clustering faces}

We finally apply TSC to the problem of clustering images of faces taken under varying illumination conditions. 
The motivation for applying TSC to this problem stems from the insight that the vectorized images of a given face taken under varying illumination conditions lie approximately in a 9-dimensional linear subspace \cite{basri_lambertian_2003}. Each 9-dimensional subspace $\cS_\l$ would then contain the images corresponding to a given person. 

 We work with the extended Yale Face Database B \cite{georghiades_illumination_2001,lee_acquiring_2005}, which contains $192 \times 168$ pixel 
 images of 38 persons, each taken under 64 different illumination conditions. 
To be able to compare our results to those reported in \cite{elhamifar_sparse_2013} for SSC, SCC, 
Local Subspace Affinity (LSA) \cite{yan_general_2006}, Low-Rank Subspace Clustering (LRSC) \cite{favaro_closed_2011}, 
and LatLRR \cite{liu_latent_2011}, we apply TSC to exactly the same data sets as used in \cite[Sec.~7.2]{elhamifar_sparse_2013}. 
 %
 The averages of the CE and the FDE we obtain for $L =2,3,5,8,10$ subjects  are reported in Table \ref{tab:faces}, along with the values from \cite[Table 5]{elhamifar_sparse_2013}. 
 Comparing these results to \cite[Table 5]{elhamifar_sparse_2013} shows that TSC performs better than LSA and SCC, but worse 
 than LRR, LatLRR, LRSC, and SSC, with the latter exhibiting the best performance in the group LSC, SCC, LRR, LatLRR, LRSC, TSC, SSC.  

As pointed out in \cite[Section~3.3]{zhang_hybrid_2012} the subspaces corresponding to different persons are extremely close to each other, which renders the corresponding clustering problem hard. Discrimination 
between the clusters (and hence persons) can be improved through preprocessing of the data set as described in \cite[Section~3.3]{zhang_hybrid_2012}. 
Specifically, the preprocessed data set $\tilde \X$ is obtained by removing  the first two principal components of $\mX$, where $\mX$ is the matrix whose columns are the data points in $\X$, and taking the points in $\tilde \X$ as the columns of the resulting matrix. 
We applied TSC with preprocessing to the same data sets as used in \cite{elhamifar_sparse_2013}. 
 Comparing the corresponding results, summarized in the first four rows in Table \ref{tab:faces}, 
to the results reported in \cite[Table 5]{elhamifar_sparse_2013}, and reproduced in our Table \ref{tab:faces}, for completeness, we can see that TSC with preprocessing performs better than LSA, SCC, and LRR 
applied to the raw data, but worse than LatLRR for $L = 2,3,5$, LRSC for $L=2,3$, and SSC for all $L$ considered, in all cases applied to the raw data. 
We note that TSC with preprocessing remains computationally less demanding than the other algorithms without preprocessing.

\begin{table}
\begin{center}
{
\setlength{\tabcolsep}{4pt}
\begin{tabular}{l*{5}{c}r}
$L$             & 2 & 3 & 5 & 8 & 10  \\
\hline
CE,  TSC, orig. dat. & 12.42\% & 19.85\% & 29.17\% & 36.84\% & 39.84\%  \\
FDE, TSC, orig. dat. &
0.0248 &  0.0419   &  0.0648 &  0.0863 &  0.0971 \\ 
CE, TSC, whitening & 8.06\% & 9\% & 10.14\% & 12.58\% & 17.86\% \\
FDE, TSC, whitening & 
0.0154 &   0.0245 & 0.0384 &  0.0525 & 0.0591 \\
CE, LSA   & 32.8\% & 52.29\% & 58.02\% & 59.19\% & 60.42\% \\
CE, SCC  & 16.62\% & 38.16\% & 58.90\% & 66.11\% & 73.02\% \\
CE, LRR  & 9.52\% & 19.52\% & 34.16\% & 41.19\% & 38.85\% \\
CE, LatLRR & 2.54\% & 4.21\% & 6.9\% & 14.34\% & 22.92\% \\
CE, LRSC & 5.32\% & 8.47\% & 12.24\% & 23.72\% & 30.36\% \\
CE, SSC & 1.86\% & 3.1\% & 4.31\% & 5.85\% & 10.94\% 
\end{tabular}
}
\end{center}
\caption{
CE and FDE for clustering faces for TSC. The CEs for LSA, SCC, LRR, LatLRR, LRSC, and SSC are taken from \cite[Table 5]{elhamifar_sparse_2013}.   \label{tab:faces}
}
\end{table}

\section*{Acknowledgments}
We would like to thank Eirikur Agustsson for helpful discussions, in particular a result he obtained inspired our proof of Lemma~\ref{lem:connectivityknng}. 
Moreover, we would like to thank Mahdi Soltanolkotabi for helpful and inspiring discussions. 

\appendices

\section{Proof of Theorem~\ref{thm:TSCprob}}
\label{app:thm:TSCprob}

The $\X_\l$ in Theorem \ref{thm:TSCprob} are obtained by choosing $n_\l$ points uniformly from $\{\vx \in \cS_\l \colon \norm[2]{\vx} = 1 \}$.  As mentioned previously, this is equivalent to choosing the points according to $\vx_j^{(\l)} = \mU^{(\l)} \va^{(\l)}_j, j \in [n_\l]$, where the $\va^{(\l)}_j$ are i.i.d.~uniform on $\US{\d_\l}$, and $\mU^{(\l)} \in \reals^{m\times \d_\l}$ is an orthonormal basis for the subspace $\cS_\l$. 

The proof is effected by showing that  the connected components in the (random) graph $G$ with adjacency matrix $\mA$ (constructed by the TSC algorithm) correspond to the $\X_\l$ with high probability. 
As mentioned previously, normalized spectral clustering will identify these components perfectly \cite[Prop.~4]{luxburg_tutorial_2007} and hence yield correct segmentation of $\X$. 

We prove that the connected components in $G$ correspond to the $\X_\l$ by showing that $G$ has no false connections and the subgraphs $G(\X_\l)$ corresponding to the $\X_\l$ are connected, for all $\l$. To this end, we define the events $\mathrm{NFC} \defeq \{ G \text{ has no false connections}\}$ and $\mathrm{C}\defeq \{ G(\X_\l) \text{ is connected}, \allowbreak\text{ for all } \l \}$ and upper-bound the probability $\PR{\overline{\mathrm{C}  \text{ and } \mathrm{NFC} }}$. This will be accomplished by exploiting the fact that conditioned on $\mathrm{NFC}$, owing to 
 $\innerprod{\vx_i}{\vx_j} = \innerprod{\va_i}{\va_j}$, for $\vx_i, \vx_j \in \X_\l$ (by orthonormality of the $\mU^{(\l)}$), 
$G(\X_\l)$ is the $q$-nearest neighbor graph of $\X_\l$ with respect to the distance $\arccos(\left|\innerprod{\va_i}{\va_j}\right|)$. An analysis of the connectivity properties of $G(\X_\l)$ will then yield an upper bound on $\PR{ \overline{ \mathrm{C} } | \mathrm{NFC} }$ which together with an upper bound on $\PR{ \overline{\mathrm{NFC} } }$ delivers the final result according to 
\begin{align}
\PR{\overline{\mathrm{C}  \text{ and } \mathrm{NFC} }  } 
&=
\PR{\overline{\mathrm{C}} \text{ or } \overline{ \mathrm{NFC} } } \nonumber \\
&= \PR{ \overline{\mathrm{NFC} } } +  \PR{ \overline{ \mathrm{C} } \text{ and } \mathrm{NFC} }
\nonumber  \\
&\leq \PR{ \overline{\mathrm{NFC} } } +  \PR{ \overline{ \mathrm{C} } | \mathrm{NFC} }   \label{eq:cnfcub}. 
\end{align}
We proceed by establishing the upper bounds on the terms in the RHS of \eqref{eq:cnfcub}.

We will use Lemma \ref{lem:cluster_isolation} below, proven in Appendix \ref{app:prooflemconn}, to upper-bound $\PR{ \overline{\mathrm{NFC} } }$. 
The lemma is also a key ingredient of the proof of  Theorem~\ref{thm:fullyrandomnew} pertaining to incomplete data, and is hence stated in a form general enough to cover that case as well.

\begin{lemma}
Suppose that $\X_\l$ is obtained by choosing $n_\l$ points in $\cS_\l$ according to $\vx_j^{(\l)} = \mU^{(\l)} \va^{(\l)}_j, j \in [n_\l]$, where the $\va^{(\l)}_j$ are i.i.d.~uniform on $\US{\d_\l}$, $\mU^{(\l)} \in \reals^{m\times \d_\l}$ (not necessarily orthonormal), and let $\X = \X_1 \cup ...  \cup  \X_L$. Assume that in each $\vx_j \in \X$ up to $s$ arbitrary entries (possibly different for different $\vx_j$) are unobserved, i.e., set to $0$. 
Pick $\rho\in [0,1)$ and suppose that $n_\l\geq n_0$, for all $\l\in [L]$, where $n_0$ is a constant that depends on $d_{\max}$ and $\rho$ only. Suppose that $q \leq  n_{\min}^\rho$ and 
\begin{align}
\frac{
\max_{k,\l \colon k\neq \l, \D\colon |\D| \leq 2s} \big\| \herm{\mU^{(k)}_{\D}  } \mU^{(\l)}  \big\|_{2\to2}
}{
\min_{\l, \D\colon |\D| \leq 2s , \norm[2]{\va}=1 }  \big\| \herm{\mU^{(\l)}_{\D}  } \mU^{(\l)}  \va \big\|_2 }      < 1
\label{eq:condTSCaffp}
\end{align} 
where $\mU^{(\l)}_{\D} \in \reals^{m\times \d_\l}$  is the matrix obtained from $\mU^{(\l)}$ by setting the rows with indices in $\D$ to zero. 
Then, $G$ has no false connections with probability at least
$
1-\sum_{\l=1}^L n_\l e^{-c_1(n_\l -1)},
$ 
where $c_1>0$ is a numerical constant.
\label{lem:cluster_isolation}
\end{lemma}

It follows from Lemma \ref{lem:cluster_isolation} with $s=0$ that 
\begin{align}
\PR{ \overline{\mathrm{NFC} } } \leq \sum_{\l=1}^L n_\l e^{-c_1(n_\l -1)}.
\label{eq:probnofalscon}
\end{align}
To see this note that for $s=0$ \eqref{eq:condTSCaffp} reduces to \eqref{eq:condTSCaffpmthm}. 
Specifically, for $s=0$, $\mU^{(\l)}_{\D} = \mU^{(\l)}$ and since the $\mU^{(\l)}$ are orthonormal we have $\herm{\mU^{(\l)}_{\D}  } \mU^{(\l)} = \herm{\mU^{(\l)}  } \mU^{(\l)} = \mI_{\d_\l}$. The denominator in \eqref{eq:condTSCaffp}  therefore equals $1$, and the numerator reduces to $ \max_{k,\l \colon k\neq \l} \big\| \herm{\mU^{(k)} } \mU^{(\l)}  \big\|_{2\to2}= \max_{k,\l \colon k\neq \l} \affp(\cS_k,\cS_\l)$ which establishes the equivalence of  \eqref{eq:condTSCaffp} and \eqref{eq:condTSCaffpmthm}.

It remains to upper-bound $\PR{ \overline{ \mathrm{C} } | \mathrm{NFC} }$. By a union bound argument, we get
\begin{align}
\PR{ \overline{ \mathrm{C} } | \mathrm{NFC} } 
\leq \sum_{\l=1}^L  \PR{ G(\X_\l) \text{ is not connected }    | \mathrm{NFC}  }.  
\label{eq:prob3condadf}
\end{align}
As mentioned above, conditioned on $\mathrm{NFC}$, $G(\X_\l)$ is the $q$-nearest neighbor graph of $\X_\l$ with pseudo-distance metric $\arccos(\left|\innerprod{\va_i}{\va_j}\right|)$ (recall that, conditioned on $\mathrm{NFC}$, we have $\innerprod{\vx_i}{\vx_j} = \innerprod{\va_i}{\va_j}$ for $\vx_i, \vx_j \in \X_\l$). 
It is this insight that allows us to find upper bounds on the terms $\PR{ G(\X_\l) \text{ is not connected }    | \mathrm{NFC}  }$ as formalized in Lemma \ref{lem:connectivityknng} below, which is proven in Appendix \ref{app:connectivity}.

\begin{lemma}
Let $\va_1,...,\va_n \in \reals^{\d}$ be drawn i.i.d.~uniformly on $\US{\d}$, $d>1$, and let $\tilde G$ 
be the corresponding $\tilde k$-nearest neighbor graph with respect to the pseudo-distance metric $\tilde s(\va_i, \va_j) = \arccos( \left|\innerprod{\va_i}{\va_j}\right|)$. 
Then, with $\tilde k \geq \gamma  \,   6 (12 \pi)^{\d-1}   \log n$,  for every $\gamma>1$,  
we have
\[
\PR{\tilde G \text{ is connected
}  } \geq 1  -  \frac{2}{n^{\gamma-1}   \gamma \log n}.
\]
\label{lem:connectivityknng}
\end{lemma}
Using Lemma \ref{lem:connectivityknng} we can then conclude that
 $\PR{ G(\X_\l) \text{ is not connected }    | \mathrm{NFC}  } \leq 2{n_\l}^{-\gamma+1}$ (using $\gamma \log n_\l \geq \log n_\l \geq \log n_0 \geq 1$) provided that $q\geq \gamma \, 6 (12 \pi)^{\d_{\l}-1} \log n_\l$, which is satisfied by the assumption $q \in \allowbreak[6 (12 \pi)^{\d_{\max}-1} \allowbreak\gamma \log n_{\max},\allowbreak   n_{\min}^\rho]$. 
Inserting into \eqref{eq:prob3condadf} yields
\begin{align}
\PR{ \overline{ \mathrm{C} } | \mathrm{NFC} }  \leq \sum_{\l = 1}^L 2{n_\l}^{-\gamma+1}. 
\label{eq:prob3cond1}
\end{align}
Finally, combining the upper bounds \eqref{eq:probnofalscon} and \eqref{eq:prob3cond1} in \eqref{eq:cnfcub}, we get
\[
\PR{\overline{\mathrm{C}  \text{ and } \mathrm{NFC} }  } \leq \sum_{\l =1}^L \left(  n_\l e^{-c_1(n_\l - 1)}  + 2{n_\l}^{-\gamma+1} \right) 
\]
as desired.

\subsection{\label{app:prooflemconn} Proof of Lemma \ref{lem:cluster_isolation}}
We need to show that $G$ has no false connections, i.e., for each $\vx_i^{(\l)} \in \X_\l$, the associated set $\S_i$ corresponds to points in $\X_\l$ only, for all $\l$. This is accomplished by proving that for $\vx_i^{(\l)} \in \X_\l$, we have
\begin{align}
z_{(n_\l - \q)}^{(\l)} > \max_{k\neq \l, j} z_{j}^{(k)}.
\label{eq:tscsdpfox22}
\end{align}
Here, $z_{j}^{(k)} \defeq \big| \big< \vx_j^{(k)} ,  \vx_i^{(\l)} \big> \big|$ and $z_{(1)}^{(\l)} \leq z_{(2)}^{(\l)} \leq ...\leq z_{(n_\l-1)}^{(\l)}$ are the order statistics of $\{z_{j}^{(\l)}\}_{j \in [n_\l] \setminus \{i\}}$. 
Note that, for simplicity of exposition, the notation $z_j^{(k)}$ does not reflect dependence on $\vx_i^{(\l)}$. 
Next, we upper-bound the probability of \eqref{eq:tscsdpfox22} being violated. A union bound over all $N$ vectors $\vx_i^{(\l)}, \l\in [L], i\in [n_\l]$, then yields the final result. 
We first note that for $k\neq \l$, by the Cauchy-Schwarz inequality, 
\begin{align*}
z_{j}^{(k)} 
&= \left| \innerprod{\vx_j^{(k)}}{ \vx_i^{(\l)} } \right| = \left| \innerprod{ \mU^{(k)}_{\D}  \va_j^{(k)}}{ \mU^{(\l)}_{\tD} \va_i^{(\l)} } \right| \\
&= \left|\innerprod{   \va_j^{(k)}}{ \herm{\mU^{(k)}_{\D}} \mU^{(\l)}_{\tD} \va_i^{(\l)} } \right| 
\leq 
\norm[2]{ \va_j^{(k)}} \norm[2]{ \herm{\mU^{(k)}_{\D}} \mU^{(\l)}_{\tD} \va_i^{(\l)}} \\
&\leq  \norm[2\to 2]{\herm{\mU^{(k)}_{\D}} \mU^{(\l)}_{\tD} } \norm[2]{  \va_j^{(k)}} \norm[2]{ \va_i^{(\l)}}
\\
&\leq \max_{k,\l \colon k\neq \l, \D\colon |\D| \leq 2s } \norm[2\to 2]{ \herm{\mU^{(k)}_{\D}  } \mU^{(\l)}  } 
\end{align*}
where the sets $\D, \tD \subset [m]$ contain the indices of the unobserved entries (set to zero) in $\vx_j^{(k)}$ and $\vx_i^{(\l)}$, respectively, and $\mU^{(\l)}_{\D},\mU^{(\l)}_{\tD}  \in \reals^{m\times \d_\l}$  are the matrices obtained from $\mU^{(\l)} \in \reals^{m\times \d_\l}$ by setting the rows with indices in $\D$ and $\tD$, respectively, to zero. 
Since the distribution of $\va_j^{(\l)}$ is rotationally invariant, we get, for a fixed $\va_i^{(\l)}$ with unit norm, that
\begin{align}
z_j^{(\l)}  
&= \left|\innerprod{   \va_j^{(\l)}}{ \herm{\mU^{(\l)}_{\D}} \mU^{(\l)}_{\tD} \va_i^{(\l)} } \right|  \nonumber \\
& = \left|\innerprod{   \va_j^{(\l)}}{  \frac{\herm{\mU^{(\l)}_{\D}} \mU^{(\l)}_{\tD} \va_i^{(\l)} }{\norm[2]{\herm{\mU^{(\l)}_{\D}} \mU^{(\l)}_{\tD} \va_i^{(\l)} }  } } \right|  \norm[2]{\herm{\mU^{(\l)}_{\D}} \mU^{(\l)}_{\tD} \va_i^{(\l)}} \nonumber \\
&\sim \left|\innerprod{   \va_j^{(\l)}}{ \va_i^{(\l)} } \right| \norm[2]{\herm{\mU^{(\l)}_{\D}} \mU^{(\l)}_{\tD} \va_i^{(\l)}} \nonumber \\
&\geq \min_{\l, \D\colon |\D| \leq 2s , \norm[2]{\va}=1 }  \norm[2]{ \herm{\mU^{(\l)}_{\D}  } \mU^{(\l)}  \va } \underbrace{ \left|\innerprod{  \va_j^{(\l)}}{ \va_i^{(\l)}} \right| }_{ \tilde z_j^{(\l)} \defeq }. \nonumber
\end{align}
This allows us to conclude that, for all $z \in \reals$, 
\[
\PR{z_j^{(\l)}  \leq z  }  \leq \PR{\min_{\l, \D\colon |\D| \leq 2s , \norm[2]{\va}=1 }  \norm[2]{ \herm{\mU^{(\l)}_{\D}  } \mU^{(\l)}  \va }  \tilde z_j^{(\l)}  \leq z   }
\]
and hence the probability of \eqref{eq:tscsdpfox22} being violated can be upper-bounded according to
\begin{align}
\PR{z_{(n_\l-\q)}^{(\l)} \leq \max_{k\neq \l, j} z_{j}^{(k)} } 
\leq 
\PR{\tilde z_{(n_\l-\q)}^{(\l)} \leq 1 - \eta }
\label{eq:probavdfil89}
\end{align} 
which, owing to \eqref{eq:condTSCaffp}, holds for an $\eta>0$. 
Next, observe that
\begin{align}
\PR{\tilde z_{(n_\l-\q)}^{(\l)} \leq 1-\eta }  
&= 
\mathrm{P} \big[ \text{there exists a set } I \subset [n_\l]\setminus \{i\} \nonumber \\
&\text{ with } |I| = n_\l-\q  \text{ such that } \tilde z_j^{(\l)} \leq 1-\eta  \text{ for all } j \in I \big] \nonumber \\
&\leq \binom{n_\l -1}{n_\l -\q}  \max_{I \colon |I| = n_\l -\q } \PR{ \tilde z_j^{(\l)} \leq 1-\eta, \text{ for all } j \in I } \label{eq:smalthanmax5} \\
&\leq \left( e \frac{n_\l-1 }{\q-1} \right)^{\q-1}  p^{n_\l-\q}      \label{eq:useboundbincoeff5} 
\end{align}
with $p = \PR{ \tilde z_j^{(\l)} \leq 1 - \eta }$ (recall that the $\tilde z_j^{(\l)}$ are i.i.d.), 
where we used a union bound to get \eqref{eq:smalthanmax5} and 
$\binom{n}{n-k} = \binom{n}{k} \leq \left( \frac{e n}{k} \right)^{k}$ \cite{cormen_introduction_2001} for \eqref{eq:useboundbincoeff5}. Since  \eqref{eq:useboundbincoeff5} is increasing in $q$, and $q \leq \, n_{\min}^\rho \leq n_{\l}^\rho$, by assumption, setting $\varrho = \frac{n_\l-1}{n_{\l}^\rho-1}$, we obtain
\begin{align}
\PR{\tilde z_{(n_\l-\q)}^{(\l)} \leq 1-\eta } 
&\leq(e \varrho)^{\frac{n_\l-1}{\varrho}} p^{(n_\l - 1) \left(1-\frac{1}{\varrho}\right)} \nonumber \\
&= \left( (e \varrho)^{\frac{1}{\varrho}} p^{1-\frac{1}{\varrho}}   \right)^{n_\l-1} \nonumber \\ 
&\leq e^{-(n_\l -1) c_1} \nonumber 
\end{align}
where the last inequality holds for a constant $c_1>0$, provided that $(e \varrho)^{\frac{1}{\varrho}} p^{1-\frac{1}{\varrho}} < 1$, i.e., if $(e\varrho)^{-\frac{1}{\varrho-1}} > p = \PR{ \tilde z_j^{(\l)} \leq 1 - \eta }$. 
This inequality can be satisfied for every given $p<1$ by taking $\varrho$ sufficiently large. Since the pdf of $\tilde z_j^{(\l)} = \left|\innerprod{  \va_j^{(\l)}}{ \va_i^{(\l)}} \right|$ is given by $f(z)= \frac{2}{\sqrt{\pi}} \frac{\Gamma(d_\l/2)}{\Gamma((d_\l-1)/2)}  (1-z^2)^{\frac{d_\l-3}{2}} \id{|z|\leq 1}$ and $\eta>0$, we, indeed, have $p < 1$. 
As $\varrho = \frac{n_\l-1}{n_{\l}^\rho-1}$ is increasing in $n_\l$ and $n_\l \geq n_0$, by assumption, $\varrho$ can, indeed, be made sufficiently large provided that $n_0$ is large enough.

\subsection{\label{app:connectivity} Proof of Lemma~\ref{lem:connectivityknng}}

Our proof is inspired by ideas from   \cite{balister_connectivity_2005,brito_connectivity_1997,xue_number_2004} dealing with the connectivity of nearest neighbor graphs for points chosen randomly in the plane. 
Here, we study the connectivity of nearest neighbor graphs $\tilde G$ for points chosen randomly on the unit sphere $\US{\d}$. The main idea of our proof is as follows. We first partition the unit sphere into $M$ regions $R_1,...,R_M$ of equal area and small diameter. 
Then we show that, for every given point $\va_i$, all points in the regions neighboring the region that contains $\va_i$ are among the $\tilde k$ nearest neighbors of $\va_i$. Next, we show that all regions $R_m$ contain at least one point, which combined with the fact that $R_1,...,R_M$ is the partitioning of a contiguous area, implies that $\tilde G$ is connected, as desired. 

We start by introducing the spherical distance metric $\s$ for points $\vx, \vy\in \US{d}$ as 
\[
\s(\vx,\vy) \defeq \arccos(\innerprod{\vx}{\vy})
\]
and defining the spherical cap around $\vp \in \US{d}$ of spherical radius $\theta \in [0,\pi/2]$ as
\[
\C(\vp, \theta) \defeq \{ \vx \in \US{\d} \colon \s(\vx, \vp) \leq \theta \}. 
\]
The distance metrics $s$ and $\tilde s$ are related according to
\begin{align}
\tilde s (\vx,\vy) 
&= \arccos(\left|\innerprod{\vx}{\vy}\right|)  \nonumber \\
&= \min(\arccos (\innerprod{\vx}{\vy}), \arccos(-\innerprod{\vx}{\vy})  ) \nonumber \\
&= \min(s(\vx,\vy) , s(-\vx,\vy)).
\label{eq:relstildes}
\end{align}
In the following, whenever we refer to the points in a region $Q$, we actually mean the points in $\{\va_1,...,\va_n\}$ that lie in $Q$, i.e., $\{\va_1,...,\va_n\}\cap Q$. We denote by $\#(Q)$ the number of points in $Q$, and by $N(\C(\va_i, \theta))$ the number of points in $\C(\va_i, \theta)$, excluding $\va_i$, i.e., $N(\C(\va_i, \theta)) \defeq |\C(\va_i, \theta) \cap \{\va_1,...\va_{i-1},\va_{i+1},...,\va_n\} |$. Note that the points contained in $\C(\va_i, \theta) \!\setminus\! \{\va_i\}$ are the $N(\C(\va_i, \theta))$ nearest neighbors of $\va_i$  with respect to (w.r.t.) the distance $s$. 
Later in the proof we will need the following relation between the nearest neighbors of a point $\va_i$ w.r.t.~the distance $s$  and the nearest neighbors of $\va_i$ w.r.t.~the distance $\tilde s$:  
The points contained in $(\C(\va_i, \theta)\setminus \{\va_i\}) \cup (\C(-\va_i, \theta) \setminus \{-
\va_i\})$ are the $N(\C(\va_i, \theta)) + N(\C(-\va_i, \theta))$ nearest neighbors of $\va_i$ w.r.t.~the distance $\tilde s$. 
To see this, first note that by \eqref{eq:relstildes} every point $\va_j$ in $\C(\va_i, \theta) \cup \C(-\va_i, \theta)$ satisfies $\tilde s(\va_i,\va_j)\leq \theta$. Since $\theta \leq \pi/2$ the caps $\C(\va_i, \theta)$ and $\C(-\va_i, \theta)$ are non-overlapping so that the total number of points in $(\C(\va_i, \theta)\setminus \{\va_i\}) \cup (\C(-\va_i, \theta) \setminus \{-
\va_i\})$ is given by $N(\C(\va_i, \theta)) + N(\C(-\va_i, \theta))$. 

We proceed to partitioning the unit sphere $\US{\d}$ into $M$ non-overlapping regions of equal area and small diameter. Such a partitioning was described in \cite{leopardi_partition_2006,leopardi_diameter_2009} and has found applications, e.g., in theoretical computer science \cite{feige_optimality_2002}. 

\begin{lemma}[{extracted from the proof of Lemma~6.2 in \cite{leopardi_diameter_2009}}]
For each $d>1$, there exists a partitioning $\mathrm{FS}(\d,M) = \{R_1,...,R_M\}$ of the unit sphere $\US{\d}$ into $M$ non-overlapping regions $R_1,...,R_M$ of equal area, 
with the spherical diameter of each $R_m$ satisfying $\sup\{ s(\vx,\vy) \colon \vx, \vy \in R_m   \} \leq    \theta^\star$. Here,
\begin{align}
\theta^\star \defeq 8 \Theta( \LM{\USL^{d-1}} / M)
\label{eq:defthetastar}
\end{align}
where $\Theta(\cdot)$ is the inverse function of $\LM{\C(\vp, \theta)}$ w.r.t.~$\theta$ $($recall that $\LM{\cdot}$ denotes the Lebesgue measure, and note that $\LM{\C(\vp, \theta)}$ is independent of $\vp \in \USL^{d-1}$$)$. 
\label{lem:partunitsphere}
\end{lemma}

Let $\mathrm{FS}(\d,M) = \{R_1,...,R_M\}$ be a partition of the unit sphere according to Lemma \ref{lem:partunitsphere}. 
Connectivity of $\tilde G$ will now be established by showing that each point $\va_i \in R_m$ is connected to all points that lie in neighboring regions of $R_m$, and in addition,  all regions contain at least one point. 
To this end, define the events $A \defeq \{\#(R_m)>0$, for all $m \in [M]\}$ and $B_m \defeq \{\#(\C(\vc_m,3\theta^\star)) \leq  k  \}$ where $\C(\vc_m,3 \theta^\star)$ is the spherical cap around an arbitrary, but fixed point $\vc_m \in R_m$, with $\theta^\star$ given by \eqref{eq:defthetastar}, and $k\defeq \tilde k/2$. We assume for expositional simplicity that $\tilde k$ is even (the proof applies with minor changes to general $k$ by setting $k\defeq \lfloor \tilde k/2\rfloor$). The proof is then effected by showing that i) on $A \cap \left( \cap_{m=1}^M B_m \right)$, $\tilde G$ is connected 
and ii) upper-bounding the probability that  $A \cap \left( \cap_{m=1}^M B_m \right)$ does not hold. 

\begin{figure}
\begin{center}
\includegraphics{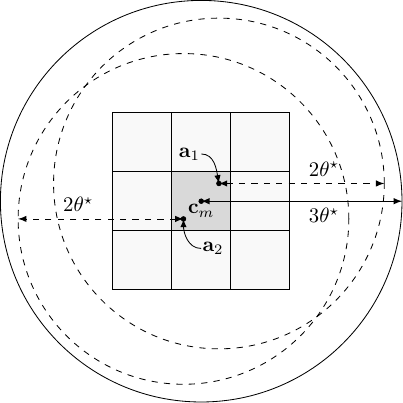}
\end{center}
\caption{\label{fig:illustspcap}$R_m$ (gray region) along with the spherical caps $\C( \va_1,2\theta^\star)$, $\C( \va_2,\theta^\star)$, 
and $\C( \vc_m,3\theta^\star)$. }
\end{figure}

By Lemma \ref{lem:partunitsphere}, the spherical cap $\C(\va_i,2\theta^\star)$ around a given $\va_i\in R_m$ contains all neighboring regions of $R_m$, and, since $\C(\va_i,2\theta^\star) \subset \C(\vc_m,3\theta^\star)$ (see Figure \ref{fig:illustspcap} for an illustration), 
on $B_m$ we have $N(\C(\va_i,2\theta^\star)) \leq  k$, for all $\va_i \in R_m$. 
All points in $\C(\va_i,2\theta^\star)\!\setminus\! \{\va_i\}$ are hence among the $k$ nearest neighbors of $\va_i$ w.r.t.~the distance $s$. 
W.l.o.g., suppose that $-\va_i \in R_{m'}$. 
By \eqref{eq:relstildes}, on $B_m \cap B_{m'}$ all points in  $(\C(\va_i, \theta) \!\setminus\! \{\va_i\}) \cup (\C(-\va_i, \theta) \!\setminus\! \{-
\va_i\})$ are therefore among the $\tilde k = 2 k$ nearest neighbors  of $\va_i$ w.r.t.~the distance $\tilde s$ (see the paragraph below \eqref{eq:relstildes}). 
On $A$, each $R_m$ contains at least one point; thus on $A \cap B_m \cap B_{m'}$, each neighboring region of $R_m$ and $R_{m'}$ contains at least one of the $\tilde k$ nearest neighbors of $\va_i$ w.r.t.~the distance $\tilde s$. 
Therefore, on $A \cap \left( \cap_{m, m'=1}^M B_m \cap B_{m'} \right) = A \cap \left( \cap_{m=1}^M B_m \right)$, each point $\va_i\in R_m$ is connected with all points in the neighboring regions of $R_m$ and each region contains at least one point. As this holds for all points $\va_1,...,\va_n$, $\tilde G$ is connected.

It remains to upper-bound the probability of $\overline{A \cap \left( \cap_{m=1}^M B_m \right)}$. We first note that
\begin{align*}
\PR{ \overline{A \cap \left( \cap_{m=1}^M B_m \right)} } 
&= 
\PR{\bar A \cup \left( \bigcup_{m=1}^{M}  \bar B_m \right)} \nonumber \\
&\leq \PR{\bar A} + \sum_{m=1}^M \PR{\bar B_m}
\end{align*}
and start by upper-bounding $\PR{\bar A}$. 
Set 
\begin{align}
M = \frac{n}{\gamma \log n}
\label{eq:defM}
\end{align}
 where $\gamma>1$ is the constant in the statement of Lemma~\ref{lem:connectivityknng}. Observe that
\begin{align}
\PR{\bar A} 
&=\PR{\cup_{m=1}^M  \{ \#(R_m)=0 \} } 
\leq \sum_{m=1}^M \PR{ \#(R_m)=0 } \nonumber \\
&= \sum_{m=1}^M \left(1 - \frac{1}{M} \right)^n \label{eq:useprnrmeqz}\\ 
&\leq M e^{ -n/ M} = \frac{n}{\gamma \log n} e^{-\gamma \log n} 
= \frac{n^{1-\gamma}}{\gamma \log n} \label{eq:leqpabar} 
\end{align}
where in \eqref{eq:useprnrmeqz} we used the fact that the $n$ points are chosen i.i.d.~and the probability of a given point ending up in $R_m$ is $1/M$.  

We next upper-bound $\PR{\bar B_m}$. To this end set $k = 3 n p$. We establish later that this choice of $k$ satisfies $\tilde k = 2 k \leq \gamma c_2 \log n$, for a constant $c_2$ depending on $d$ only. Since the $k'$-nearest neighbor graph of $\{\va_1,...,\va_n\}$ with $k'\geq \tilde k$ is connected if the $\tilde k$-nearest neighbor graph $\tilde G$ is connected, this will yield the desired result.

First note that $\#(\C(\vc_m,3\theta^\star))$ is binomially distributed with parameters $(n,p)$, where $p \defeq  \LM{\C(\vc_m,3\theta^\star)} / \LM{\US{\d}}$.  
By a tail bound on the binomial distribution \cite[Thm.~1]{janson_concentration_2002} we obtain, with $t= 2 np$, that 
\begin{align}
\PR{\bar B_m }  
&= \PR{\#(\C(\vc_m,3\theta^\star)) > np  + t  } \nonumber \\
&\leq e^{- \frac{t^2}{2( np + t/3)} }  = e^{-\frac{6}{5} np}  \leq e^{-np}.
\end{align}
Since $R_m \subset \C(\vc_m,3\theta^\star)$, we have
\[
p = \frac{\LM{\C(\vc_m,3\theta^\star) }}{ \LM{\US{\d}} }\geq \frac{\LM{R_m} }{ \LM{\US{\d}}} =  \frac{1}{M} = \frac{\gamma \log n}{ n}.
\]
By a union bound we thus get
\begin{align}
\PR{\bigcup_{m=1}^{M}  \bar B_m } 
 \leq 
 M e^{-np} \leq M e^{-n/M} = \frac{n^{1-\gamma}}{\gamma \log n}. 
\label{eq:anyngeqk}
\end{align}
Combining \eqref{eq:leqpabar} and \eqref{eq:anyngeqk} yields 
$\PR{\bar A \cup \left( \bigcup_{m=1}^{M}  \bar B_m \right)} \leq  \frac{2}{n^{\gamma-1}   \gamma \log n}$. 

It remains to show that there exists a constant $c_2$ (depending on $d$ only) such that $\tilde k = 2k  \leq   \gamma \, c_2 \log n$. 
This is accomplished by upper-bounding $\LM{\C(\vc_m,3\theta^\star)}$ and using this upper bound to establish that $k = 3 n p = 3 n \LM{\C(\vc_m,3\theta^\star)}/\LM{\US{\d}} \leq  \gamma \, \frac{c_2}{2} \log n$. 
To this end, we first upper-bound $\theta^\star$ in \eqref{eq:defthetastar} and then use this bound to upper-bound $\LM{\C(\vc_m,3\theta^\star)}$. By  \cite[Eq.~5.9]{leopardi_diameter_2009}, we have
\begin{align}
\theta^\star 
&= 8 \Theta( \LM{\USL^{d-1}} / M)   
\leq 8  \arcsin \! \left( \left( \frac{\LM{\USL^{\d-1}}}{\LM{\USL^{\d-2}}} \frac{d-1}{M} \right)^{\frac{1}{d-1}}\right)   \nonumber \\
&\leq 4 \pi \left( \frac{\LM{\USL^{\d-1}}}{\LM{\USL^{\d-2}}} \frac{d-1}{M} \right)^{\frac{1}{d-1}}
\label{eq:boundthetst}
\end{align}
where we used $\arcsin(x) \leq \frac{\pi}{2} x$, for $0\leq x \leq 1$. We next establish that the argument of $\arcsin$ in \eqref{eq:boundthetst} is, indeed, smaller than $1$. Using $\LM{\USL^{d-1}} = \frac{2 \pi^{d/2}}{\Gamma(d/2)}$ (e.g., \cite[p.~1]{leopardi_diameter_2009}) and $\frac{\Gamma(\frac{d-1}{2})  }{ \Gamma(\frac{d}{2})} \leq \sqrt{2} \frac{\sqrt{d}}{d-1}$ (e.g., ~\cite[Eq.~8.1]{foucart_mathematical_2013}), we obtain
\begin{align*}
&\frac{\LM{\USL^{\d-1}}}{\LM{\USL^{\d-2}}} \frac{d-1}{M} 
= \sqrt{\pi} \frac{\Gamma(\frac{d-1}{2})  }{ \Gamma(\frac{d}{2})} \frac{d-1}{M}  
\leq 
\sqrt{2\pi} \frac{\sqrt{d}}{M} 
= 
\sqrt{2\pi d } \frac{ \gamma \log n}{n} \leq 6(12\pi)^{d-1} \frac{ \gamma \log n}{n} \leq 1
\end{align*}
where we used $\sqrt{2\pi d } \leq 6(12\pi)^{d-1}$ for $d\geq1$, and the last inequality holds by the assumption $n \geq \tilde k \geq \gamma  \,   6 (12 \pi)^{\d-1}   \log n$. 

 Application of \cite[Eq.~5.2]{leopardi_diameter_2009} and subsequently of \eqref{eq:boundthetst} yields 
\begin{align*}
&\LM{\C(\vc_m,3\theta^\star)} 
\leq \frac{\LM{\USL^{\d-2}}}{\d-1} 
 (3\theta^\star)^{\d-1}  
\leq
\frac{\LM{\USL^{\d-2}}}{\d-1} (12 \pi)^{d-1} \frac{\LM{\USL^{d-1}}}{\LM{\USL^{d-2}}} \frac{d-1}{M}  
= 
(12\pi)^{d-1} \frac{\LM{\USL^{d-1}}}{M} .
\end{align*}
We thus have
\[
k 
= 
3 n p 
= 3 n \, \frac{\LM{\C(\vc_m,3\theta^\ast)} }{ \LM{\US{\d}} } 
\leq  
3\cdot (12 \pi)^{\d-1}
 \, \gamma
 \,\log n
\]
and hence $k \leq \gamma \frac{c_2}{2} \log n$ with $c_2 = 6 (12 \pi)^{\d-1}$, as desired.

\section{Proof of Theorem~\ref{thm:aff2} and Corollary \ref{cor:ofnoisycase}}
\label{app:thmaff2}

Analogously to Theorem \ref{thm:TSCprob}, the proof of Theorem~\ref{thm:aff2} is established by upper-bounding the probability $\PR{\overline{\mathrm{C}  \text{ and } \mathrm{NFC} }  } $ according to 
\begin{align}
\PR{\overline{\mathrm{C}  \text{ and } \mathrm{NFC} }  } 
\leq \PR{ \overline{\mathrm{NFC} } } +  \PR{ \overline{ \mathrm{C} } |\mathrm{NFC} }   
\label{eq:cnfcub2}
\end{align}
where $\mathrm{NFC} = \{ G \text{ has no false connections}\}$ and $\mathrm{C} = \{ G(\X_\l) \text{ is connected, for all } \l \}$, as in the proof of Theorem \ref{thm:TSCprob}. 
We start by upper-bounding $\PR{ \overline{\mathrm{NFC} } }$. 
Since $q \leq n_{\min}/6$ and \eqref{eq:condthmnoisycase} for $\sigma=0$ reduces to \eqref{eq:condTSCaff2} it follows from Theorem \ref{thm:noisycase} (the assumption $m \geq 6 \log N$, i.e.,  $\sqrt{6\log N}/\sqrt{m}= \beta / \sqrt{m} \leq 1$ relevant for Step 1 in the proof of Theorem \ref{thm:noisycase} is not needed owing to $\sigma=0$) that 
\begin{align}
\PR{ \overline{\mathrm{NFC} } } \leq \frac{10}{N} + \sum_{\l\in [L]} n_\l e^{-c(n_\l-1)}
\label{eq:probestlemads}
\end{align}
where $c>0$ is a numerical constant.

We next upper-bound $\PR{ \overline{ \mathrm{C} } |\mathrm{NFC} }$. 
In Appendix \ref{app:thm:TSCprob} we established that (cf.~\eqref{eq:prob3cond1} with $\gamma = 3$) 
\begin{align}
\PR{ \overline{ \mathrm{C} } | \mathrm{NFC} } 
\leq  \sum_{\l \in [L]} 2{n_\l}^{-2}
\label{eq:prob3cond1_rep}
\end{align}
provided that $q\geq  3 \cdot 6 (12 \pi)^{\d_{\l}-1}  \log n_\l$, for all $\l$, 
which is satisfied by the assumption $q \in [c_1 \log n_{\max},  \allowbreak n_{\min}/6]$ with $c_1 = 18 (12 \pi)^{\d_{\max}-1}$. 
Using \eqref{eq:probestlemads} and \eqref{eq:prob3cond1_rep} in \eqref{eq:cnfcub2} finally yields
\[
\PR{\overline{\mathrm{C}  \text{ and } \mathrm{NFC} }  } \leq 10/N+ \sum_{\l =1}^L \left(  n_\l e^{-c(n_\l - 1)}  + 2{n_\l}^{-2} \right)
\]
as desired. 

Corollary \ref{cor:ofnoisycase} follows directly from \eqref{eq:probestlemads}. 

\section{Proof of Theorem~\ref{thm:noisycase}}
\label{app:thm:noisycase}

As in the proof of Lemma \ref{lem:cluster_isolation}, we show that $G$ has no false connections by establishing that for each $\vx_i^{(\l)} \in \X_\l$ the associated set $\S_i$ corresponds to points in $\X_\l$ only. 
Again, this is accomplished by showing that 
\begin{align}
z_{(n_\l - \q)}^{(\l)} > \max_{k\neq \l, j} z_{j}^{(k)}
\label{eq:tscsdpfox2}
\end{align}
where $z_{j}^{(k)} = \big| \big< \vx_j^{(k)} ,  \vx_i^{(\l)} \big> \big|$. 
Next, we upper-bound the probability of \eqref{eq:tscsdpfox2} being violated. A union bound over all $N$ vectors $\vx_i^{(\l)}, i\in [n_\l], \l\in [L]$, will, as before, yield the final result. 
We start by setting 
\begin{align}
\tilde z_j^{(k)} \defeq  \left| \innerprod{  \va_j^{(k)}}{  \herm{\mU^{(k)}} \mU^{(\l)} \va_i^{(\l)}} \right|
\label{eq:defztildejk}
\end{align}
and noting that
\begin{align}
z_j^{(k)} = 
 \left| \innerprod{  \va_j^{(k)}}{  \herm{\mU^{(k)}} \mU^{(\l)} \va_i^{(\l)}} +  e_j^{(k)}  \right|
\label{eq:zjkdefinappb}
\end{align}
with 
\begin{align}
e_j^{(k)} \defeq \innerprod{ \ve_j^{(k)}}{ \ve_i^{(\l)} }  +\innerprod{  \ve_j^{(k)}}{   \mU^{(\l)} \va_i^{(\l)}}  +\innerprod{   \mU^{(k)} \va_j^{(k)} }{ \ve_i^{(\l)} }.
\label{eq:defejk}
\end{align}
Now recall that $z_{(1)}^{(\l)} \leq z_{(2)}^{(\l)} \leq ...\leq z_{(n_\l-1)}^{(\l)}$ are the order statistics of $\{z_{j}^{(\l)}\}_{j \in [n_\l] \setminus \{i\}}$. It follows that
\[
\tilde z_{(n_\l-\q)}^{(\l)} - \max_{j\neq i} |e_j^{(\l)}|  \leq z_{(n_\l-\q)}^{(\l)}
\]
and hence the probability of \eqref{eq:tscsdpfox2} being violated can be upper-bounded according to
\begin{align}
\PR{z_{(n_\l-\q)}^{(\l)} \leq \max_{k\neq \l, j} z_{j}^{(k)} }  
&\leq 
\PR{\tilde z_{(n_\l-\q)}^{(\l)} \!-\! \max_{j\neq i} \big|e_j^{(\l)}\big|   \leq \max_{k\neq \l, j} \tilde z_{j}^{(k)}  +  \max_{k\neq \l, j}  \big|e_{j}^{(k)} \big|   } \nonumber\\
&\leq 
\PR{ \tilde z_{(n_\l-\q)}^{(\l)} \leq \frac{\nu}{\sqrt{\d_\l}}   } \nonumber \\
&+\! \PR{ \alpha + 2\epsilon  \leq \max_{j\neq i} \big|e_j^{(\l)} \big|   \!+\! \max_{k\neq \l, j} \tilde z_{j}^{(k)}  \!+\!  \max_{k\neq \l, j}  \big|e_{j}^{(k)}\big|   } \label{eq:assumeasdage} \\
&\leq 
\PR{\tilde z_{(n_\l-\q)}^{(\l)} \leq \frac{\nu}{\sqrt{\d_\l}}   }  +\PR{ \max_{k\neq \l, j} \tilde z_{j}^{(k)}  \geq \alpha  }  \nonumber \\
&+ \underbrace{\PR{\max_{j\neq i} \big|e_{j}^{(\l)} \big| \geq  \epsilon   }   +  \PR{ \max_{k\neq \l, j} \big|e_{j}^{(k)}\big| \geq  \epsilon  }}_{\leq \sum_{(j,k)\neq (i,\l)} \PR{ \big|  e_j^{(k)} \big|   \,\geq \, \epsilon }  } \label{eq:prtoboundt}
\end{align}
where $\alpha, \epsilon$, and $\nu$ are chosen later. In \eqref{eq:assumeasdage} and \eqref{eq:prtoboundt} we used that for random variables $X$ and $Y$, possibly dependent, and constants $\phi$ and $\varphi$ satisfying $\phi \geq \varphi$, we have
 \begin{align}
\PR{X \leq Y} 
&\leq \PR{ \{X \leq \phi \} \cup \{\varphi \leq Y\} } \nonumber \\
&\leq \PR{X\leq \phi} + \PR{\varphi \leq Y}.
\label{eq:splitprob}
\end{align}
Specifically, in \eqref{eq:assumeasdage} we used \eqref{eq:splitprob} with $\phi=\frac{\nu}{\sqrt{\d_\l}}$ and $\varphi = \alpha + 2 \epsilon$, which leads to the assumption $\alpha + 2 \epsilon \leq \frac{\nu}{\sqrt{\d_\l}}$, 
resolved below. 
We next upper-bound the individual terms in \eqref{eq:prtoboundt} to get the following results proven at the end of this appendix:
\paragraph*{Step 1:} Setting $\epsilon = \frac{2 \sigma(1+\sigma)}{\sqrt{m}} \beta$, we have for all $\beta$ with $\frac{1}{\sqrt{2\pi}} \leq \beta \leq \sqrt{m}$ that 
\begin{align}
\PR{ \left|  e_j^{(k)} \right|   \geq  \epsilon } \leq 7 e^{-\frac{\beta^2}{2}}. 
\label{eq:boundonnoise}
\end{align}
\paragraph*{Step 2:} Setting 
\begin{align}
\alpha = \frac{\beta(1+\beta)}{\sqrt{d_\l}}  \max_{k\neq \l}  \frac{1}{\sqrt{d_k}}  \norm[F]{ \herm{\mU^{(k)}} \mU^{(\l)} }
\label{eq:defalpha}
\end{align}
we have for all $\beta \geq 0$ that
\begin{align}
&\PR{\max_{k\neq \l, j} \tilde z_{j}^{(k)}  \geq \alpha } \leq \sum_{k \in [L] \setminus \{\l\}} (1+2n_k) e^{- \frac{\beta^2}{2}} \leq 3N e^{- \frac{\beta^2}{2}} . \label{eq:maxboundadfa}
\end{align}
\paragraph*{Step 3:}
For $\nu = 2/3$ and $n_\l \geq 6  q $, 
there is a constant $c = c(\nu) > 1/20$ such that
\begin{align} 
\PR{  \tilde z_{(n_\l-\q)}^{(\l)} \leq \frac{\nu}{\sqrt{d_\l}} } \leq e^{-c (n_\l-1)} \label{eq:znidlexa}.
\end{align}
Before presenting the detailed arguments leading to \eqref{eq:boundonnoise}, \eqref{eq:maxboundadfa}, and \eqref{eq:znidlexa}, we show how the proof is completed. 
Setting $\beta = \sqrt{6 \log N }$ and using  \eqref{eq:boundonnoise}, \eqref{eq:maxboundadfa} (note that $\beta \leq \sqrt{m}$ is satisfied since, by assumption, $m\geq 6 \log N$), and \eqref{eq:znidlexa} in \eqref{eq:prtoboundt} yields   
\begin{align}
\PR{z_{(n_\l-\q)}^{(\l)} \leq \max_{k\neq \l, j} z_{j}^{(k)} } 
&\leq e^{-c(n_\l-1)} +  3N e^{- \frac{\beta^2}{2}}    + 7 N e^{-\frac{\beta^2}{2}}    \nonumber \\
&= \frac{10}{N^2} + e^{-c(n_\l-1)} .
\label{eq:prtscfails}
\end{align}
Taking the union bound over all vectors $\vx_i^{(\l)}, i \in [n_\l], \l \in [L]$, yields the desired lower bound on $G$ having no false connections. 

Recall that for \eqref{eq:assumeasdage}  we imposed the condition  $\alpha+ 2 \epsilon \leq \frac{\nu}{\sqrt{\d_\l}}$. 
With our choices for $\epsilon, \alpha$, and $\nu$ in Steps 1, 2, and 3, respectively, this condition 
becomes  
\begin{align}
\beta(1+\beta)  \max_{k\neq \l}  \frac{1}{\sqrt{d_k}}  \norm[F]{ \herm{\mU^{(k)}} \mU^{(\l)} } + 4 \sigma(1+\sigma) \frac{\sqrt{d_\l}}{\sqrt{m}} \beta \leq \frac{2}{3}.
\label{eq:condassumeadagesimp}
\end{align}
Next, note that
$
(1+\beta) \leq 
4 \sqrt{\log N}
$
as a consequence of $N\geq 6$ ($N = \sum_{\l=1}^L n_\l$, and $n_\l\geq 6 q \geq 6$, for all $\l$), 
by assumption. 
Therefore, \eqref{eq:condassumeadagesimp}  is implied by 
\[
\max_{k\neq \l}  \frac{1}{\sqrt{d_k}}  \norm[F]{ \herm{\mU^{(k)}} \mU^{(\l)} } +  \frac{\sigma(1+\sigma)}{\sqrt{\log N}} \frac{\sqrt{d_\l}}{\sqrt{m}}  \leq \frac{2}{3 \cdot 4 \sqrt{6}  \log N}
\]
which, in turn, is implied by \eqref{eq:condthmnoisycase}. 
This concludes the proof.

It remains to prove the bounds \eqref{eq:boundonnoise}, \eqref{eq:maxboundadfa}, and \eqref{eq:znidlexa}. 

\paragraph*{Step 1, proof of \eqref{eq:boundonnoise}:}
By an argument of the form \eqref{eq:splitprob}, we get
\begin{align}
\PR{\left|e_j^{(k)} \right|  \geq \epsilon} 
&\leq
\PR{ \left| \innerprod{ \ve_j^{(k)}}{ \ve_i^{(\l)}} \right|  \geq \frac{2 \sigma^2}{\sqrt{m}}   \beta}  \nonumber \\
&+
\PR{ \left| \innerprod{  \ve_j^{(k)}}{   \mU^{(\l)} \va_i^{(\l)}} \right| \geq \frac{\sigma}{\sqrt{m}} }   \nonumber \\
&+
\PR{ \left| \innerprod{ \mU^{(k)} \va_j^{(k)} }{ \ve_i^{(\l)} } \right|  \geq \frac{\sigma}{\sqrt{m}} }. 
\label{eq:unboundoneps}
\end{align}
We next upper-bound the probabilities in \eqref{eq:unboundoneps}. 
Conditional on $\va_j^{(k)}$, with $\norm[2]{\mU^{(k)} \va_j^{(k)}} = 1$, we have $\big< \mU^{(k)} \va_j^{(k)} , \ve_i^{(\l)} \big> \sim \mathcal N(0,\sigma^2/m)$. Using Lemma \ref{lem:qfunction} in Appendix \ref{app:lemmata}, for $\beta \geq \frac{1}{\sqrt{2\pi}}$, we hence get 
\begin{align}
\PR{  \left| \innerprod{   \mU^{(k)} \va_j^{(k)} }{ \ve_i^{(\l)} } \right|  \geq \frac{\sigma}{\sqrt{m}}  \beta  } \leq 2e^{-\frac{\beta^2}{2}}. 
\label{eq:noiseinss}
\end{align}
Next, we upper-bound the first term on the RHS of \eqref{eq:unboundoneps}. 
Conditional on $\ve_i^{(\l)}$, we have $\big< \ve_j^{(k)}  , \ve_i^{(\l)} \big> \sim \mathcal N \!\big(0,\frac{\sigma^2}{m} \big\|\ve_i^{(\l)} \big\|^2_2 \big)$.   Lemma \ref{lem:qfunction} yields, for  $\beta \geq \frac{1}{\sqrt{2\pi}}$, that 
\begin{align}
\PR{ \left| \innerprod{ \ve_j^{(k)}  }{ \ve_i^{(\l)} } \right|  \geq \beta \frac{\sigma}{\sqrt{m}} \norm[2]{\ve_i^{(\l)}}  } \leq 2e^{-\frac{\beta^2}{2}}.
\label{eq:adfj1}
\end{align}
Since $\beta =\sqrt{6\log N} \leq \sqrt{m}$, 
by assumption, we get
\begin{align}
\PR{  \norm[2]{ \ve_i^{(\l)} }   \geq 2 \sigma  } \leq 
\PR{  \norm[2]{ \ve_i^{(\l)} }   \geq  \left(1+\frac{\beta}{\sqrt{m}}  \right) \sigma  } \leq e^{-\frac{\beta^2}{2}}
\label{eq:adfj2}
\end{align}
where the second inequality follows from \eqref{eq:gaussnormconc}. 
Next, note that for random variables $X,Y$, possibly dependent, and a constant $\phi$, we have
\begin{align}
\PR{X \geq \phi}  
&= \PR{ \{ X \geq Y \geq \phi\} \cup \{X \geq \phi \geq Y\} \cup \{Y \geq X \geq \phi\} }  \nonumber \\
&\leq \PR{ \{ X \geq Y \} \cup \{Y \geq \phi \} }  \nonumber \\
&\leq \PR{X \geq Y} + \PR{Y \geq \phi}. 
\label{eq:combviaproduc}
\end{align}
Combining \eqref{eq:adfj1} and \eqref{eq:adfj2} via \eqref{eq:combviaproduc} yields
\begin{align}
\mathrm{P} 
\left[  \phantom{\innerprod{ \ve_j^{(k)}}{ \ve_i^{(\l)} } }  \right.
\hspace*{-1.7cm} 
\underbrace{\left| \innerprod{ \ve_j^{(k)}}{ \ve_i^{(\l)} } \right|}_{X}  
 \geq  
 \underbrace{ \frac{2 \sigma^2}{\sqrt{m}} \beta }_\phi 
\left.  \phantom{ \ve_i^{(\l)}}  
\hspace*{-0.6cm}
\right] 
&\leq 
\mathrm{P}\left[ \left| \innerprod{ \ve_j^{(k)}  }{ \ve_i^{(\l)} } \right|  \geq 
\right.
\underbrace{
\beta \frac{\sigma}{\sqrt{m}} \norm[2]{\ve_i^{(\l)}}  
}_{Y} 
\left.  \phantom{ \ve_i^{(\l)}}  
\hspace*{-0.6cm}
\right] 
+ \PR{  \norm[2]{ \ve_i^{(\l)} }   \geq  2 \sigma  } 
\leq 3 e^{-\frac{\beta^2}{2}}.
\label{eq:noisenoisinnerpr}
\end{align}
Finally, using \eqref{eq:noiseinss} and \eqref{eq:noisenoisinnerpr} in  \eqref{eq:unboundoneps} gives the desired result \eqref{eq:boundonnoise}.

\paragraph*{Step 2, proof of \eqref{eq:maxboundadfa}:}
We first upper-bound the probability of $\max_{j} \tilde z_{j}^{(k)}$, for a given $k$, 
to exceed a constant, which then yields, via a union bound over $k$, an upper bound on the probability of $\max_{k\neq \l, j} \tilde z_{j}^{(k)}$ exceeding a constant. For convenience, we set $\mB \defeq \herm{\mU^{(k)}} \mU^{(\l)}$ so that $\tilde z_j^{(k)} = \big| \big< \va_j^{(k)},   \mB \va_i^{(\l)}\big> \big|$. 
We start by noting \cite[Proof of Lem.~7.5]{soltanolkotabi_geometric_2011} that
\begin{align}
\PR{\norm[2]{\mB \va_i^{(\l)} }  \geq \frac{\norm[F]{\mB}}{\sqrt{d_\l}} + \kappa  } \leq e^{-d_\l \frac{\kappa^2}{2 \norm[2\to 2]{\mB}^2}}.
\label{eq:lipschunif}
\end{align}
Setting $\kappa = \beta \norm[F]{\mB}/\sqrt{d_\l}$ in \eqref{eq:lipschunif} yields
\begin{align}
\PR{\norm[2]{\mB \va_i^{(\l)}}  \geq \frac{1+\beta}{\sqrt{d_\l}}  \norm[F]{\mB}   } \leq e^{ -\frac{\beta^2}{2} \frac{\norm[F]{\mB}^2}{ \norm[2\to 2]{\mB}^2}}  \leq e^{- \frac{\beta^2}{2}}. 
\label{eq:normunoslm}
\end{align}
By Proposition \ref{thm:hoeffsphere} in Appendix \ref{app:lemmata}, we have
\begin{align}
\PR{ \left|\innerprod{\va_j^{(k)}}{ \mB \va_i^{(\l)} }\right|  > \frac{\beta}{\sqrt{d_k}} \norm[2]{ \mB \va_i^{(\l)}} } 
\leq 2 e^{-\frac{ \beta^2}{2}}. 
\label{eq:hoefftypeforsphere}
\end{align}
Now, using \eqref{eq:combviaproduc} with $X = \max_{j} \tilde z_{j}^{(k)}$, $\phi =\frac{\beta}{\sqrt{d_k}} \frac{1+\beta}{\sqrt{d_\l}} \norm[F]{\mB}$, and $Y=\frac{\beta}{\sqrt{d_k}} \norm[2]{ \mB \va_i^{(\l)}}$, we get
\begin{align}
\PR{\max_{j} \tilde z_{j}^{(k)}  \geq \frac{\beta}{\sqrt{d_k}} \frac{1+\beta}{\sqrt{d_\l}} \norm[F]{\mB} }  
&\leq   \PR{ \max_j \left|\innerprod{\va_j^{(k)}}{ \mB \va_i^{(\l)} }\right|  \geq \frac{\beta}{\sqrt{d_k}} \norm[2]{ \mB \va_i^{(\l)}} } \nonumber \\
&+
\PR{\norm[2]{\mB \va_i^{(\l)}}  \geq \frac{1+\beta}{\sqrt{d_\l}}  \norm[F]{\mB}   }
 \nonumber \\ 
&\leq  
 \sum_{j \in [n_k]} \PR{ \left|\innerprod{\va_j^{(k)}}{ \mB \va_i^{(\l)} }\right|  \geq \frac{\beta}{\sqrt{d_k}} \norm[2]{ \mB \va_i^{(\l)}} }   \nonumber \\
 &+ \PR{\norm[2]{\mB \va_i^{(\l)}}  \geq \frac{1+\beta}{\sqrt{d_\l}}  \norm[F]{\mB}   }
  \label{eq:solub} \\
 &\leq  
(1+2n_k) e^{- \frac{\beta^2}{2}} \label{eq:maxboundadfa2}
\end{align}
where a union bound is used to obtain \eqref{eq:solub}, and \eqref{eq:maxboundadfa2} follows from \eqref{eq:normunoslm} and \eqref{eq:hoefftypeforsphere}. Taking the union bound over $k \in [L] \! \setminus \! \{\l\}$ concludes the proof of \eqref{eq:maxboundadfa}.

\paragraph*{Step 3, proof of \eqref{eq:znidlexa}:}
We first note that the pdf of $\tilde z_j^{(\l)} = \big< \va^{(\l)}_j  , \va^{(\l)}_i \big>$ is given by $f(z)= \frac{1}{\sqrt{\pi}} \frac{\Gamma(d_\l/2)}{\Gamma((d_\l-1)/2)}  (1-z^2)^{\frac{d_\l-3}{2}} \id{|z|\leq 1}$. 
Hence, we get 
\begin{align}
\PR{\tilde z_j^{(\l)} \leq \frac{\nu}{\sqrt{d_\l}} }  
&\leq \frac{2}{\sqrt{\pi}} \frac{\, \Gamma(d_\l/2)}{\Gamma((d_\l-1)/2)}   \int_0^{\frac{\nu}{\sqrt{d_\l}}} (1-z^2)^{\frac{d_\l-3}{2}} \id{z\leq 1} dz \nonumber \\
&\leq \frac{2}{\sqrt{\pi}} \frac{\, \Gamma(d_\l/2)}{\Gamma((d_\l-1)/2)} \frac{\nu}{\sqrt{d_\l}}  \leq  \underbrace{\sqrt{\frac{2}{\pi}} \;\nu}_{p_\nu \defeq}
\label{eq:ubonprzgsd}
\end{align}
where the last inequality follows from \cite[Eq.~8.1]{foucart_mathematical_2013}. 
Next, observe that, 
\begin{align}
\PR{\tilde z_{(n_\l-\q)}^{(\l)} \leq \frac{\nu}{\sqrt{\d_\l}} } 
&= 
\mathrm{P} \big[ \text{there exists a set } I \subset [n_\l]  \!\setminus\! \{i\} \text{ with }\nonumber \\ 
&\hspace{1cm} |I| = n_\l-\q \text{ such that } \tilde z_j^{(\l)} \leq \frac{\nu}{\sqrt{d_\l}}  \text{ for all } j \in I \big] \nonumber \\
&\leq \binom{n_\l-1}{n_\l-\q}  \max_{I \colon |I| = n_\l-\q } \PR{ \tilde z_j^{(\l)} \leq \frac{\nu}{\sqrt{d_\l}}, \text{ for all } j \in I } \label{eq:smalthanmax} \\
&\leq \left( e \frac{n_\l-1 }{\q-1} \right)^{\q-1}  \left( \PR{ \tilde z_j^{(\l)} \leq \frac{\nu}{\sqrt{d_\l}}  } \right)^{n_\l-\q}      \label{eq:useboundbincoeff} \\
&\leq \left( e \frac{ n_\l-1}{\q-1} \right)^{\q-1}  p_\nu^{n_\l-\q}     
=(e \varrho)^{\frac{n_\l-1}{\varrho}} p_\nu^{(n_\l - 1) \left(1-\frac{1}{\varrho}\right)} \label{eq:useeq:ubonprzgsd} \\
&=   
\exp\left( \phantom{\frac{1}{2}} \hspace{-0.4cm}  \right. -(n_\l-1)  \underbrace{\left( \log\left( \frac{1}{p_\nu}\right) \left(1-\frac{1}{\varrho} \right)  - \frac{1}{\varrho} \log(e \varrho) \right)}_{c(\varrho, \nu) \defeq } \left. \phantom{\frac{1}{2}} \hspace{-0.3cm} \right)
\end{align}
where we used a union bound to get \eqref{eq:smalthanmax},  
$\binom{n}{n-k} = \binom{n}{k} \leq \left( \frac{e n}{k} \right)^{k}$ \cite{cormen_introduction_2001} and the fact that the $\tilde z_j^{(\l)}$ are i.i.d.~for \eqref{eq:useboundbincoeff}, and  \eqref{eq:ubonprzgsd} yields \eqref{eq:useeq:ubonprzgsd}; we also set $\varrho \defeq \frac{n_\l-1}{\q-1}$ for notational convenience. 
Here, $c(\varrho,\nu)$ satisfies $c(\varrho,\nu) > 1/20$ for $\nu = 2/3$ and $\varrho \geq 6$, as desired. Note that $\varrho = \frac{n_\l-1}{\q-1} \geq  \frac{n_\l}{\q} \geq 6$, where both inequalities follow from $n_\l \geq 6q$, for all $\l$, which holds by assumption.

\section{Proof of Theorem~\ref{thm:fullyrandomnew}}
\label{app:thm:fullyrandomnew}

Theorem~\ref{thm:fullyrandomnew} follows from Lemma~\ref{lem:cluster_isolation} by establishing that the clustering condition \eqref{eq:condTSCaffp} is satisfied for i.i.d.~Gaussian random matrices $\mU^{(\l)} \in \reals^{m\times d}$ with high probability. This is accomplished via the following lemma, which shows that certain submatrices of Gaussian matrices $\mU^{(\l)} \in \reals^{m\times d}$ are approximately pairwise orthogonal, as long as $m$ is sufficiently large relative to $d$. 

\begin{lemma}
Let the entries of the $\mU^{(\l)} \in \reals^{m\times d}$, $\l\in[L]$, be i.i.d.~$\mathcal N(0, 1/m)$, and let $\mU^{(\l)}_{\D} \in \reals^{m\times d}$ be the matrix obtained from $\mU^{(\l)} \in \reals^{m\times d}$ by setting the rows with indices in $\D \subseteq [m]$ to zero. 
Then, we have for $\delta \in (0,1)$ with probability at least $1- 4e^{-c' m}$, where $c'$ is a numerical constant, that
\begin{align}
\min_{\l, \, \D\colon |\D| \leq 2s, \, \norm[2]{\va}=1 }  \norm[2]{ \herm{\mU^{(\l)}_{\D}  } \mU^{(\l)}  \va }    \geq  (1-\delta) \frac{m-2s}{m}   
\label{eq:genthmc1again}
\end{align}
and 
\begin{align}
\max_{k,\ell\colon k\neq \l, \, \D\colon |\D| \leq 2s } \norm[2\to 2]{ \herm{\mU^{(k)}_{\D}  } \mU^{(\l)}  }  \leq \delta
\label{eq:genthmc2again}
\end{align}
provided that
\begin{align}
m \geq 
 \frac{c_2}{\delta^2} \left( 3d + \log L + s \log\left(\frac{me}{2s}\right)   \right) + c_3 s  
\label{eq:condconcsm_inthm}
\end{align}
where $c_2,c_3>0$ are numerical constants. 
\label{thm:fullyrandom}
\end{lemma}

Before proving Lemma~\ref{thm:fullyrandom}, we show how the proof of Theorem~\ref{thm:fullyrandomnew} can be completed. 
Set $\delta = \frac{c_1}{2} \frac{c_3-2}{c_3}$, 
where $c_3$ is the constant in \eqref{eq:finalcondfullyrand} and $c_1$ is a constant satisfying $c_1<1$. With this choice of $\delta$, 
 \eqref{eq:finalcondfullyrand} (with $c_4 = c_2/\delta^2 =  c_2 \left( 2c_3/(c_1(c_3-2))\right)^2$) implies \eqref{eq:condconcsm_inthm} and hence, by Lemma~\ref{thm:fullyrandom}, with probability $\geq 1- 4e^{- c' m}$, we have
\begin{align}
&\frac{
\max_{k,\l \colon k\neq \l, \D\colon |\D| \leq 2s } \big\| \herm{\mU^{(k)}_{\D}  } \mU^{(\l)}  \big\|_{2\to2}
}{
\min_{l, \D\colon |\D| \leq 2s , \norm[2]{\va}=1 }  \big\| \herm{\mU^{(\l)}_{\D}  } \mU^{(\l)}  \va \big\|_2
} \leq \frac{\delta}{(1-\delta) \frac{m-2s}{m}}
\leq 2 \delta \frac{m}{m-2s}   
\leq 2 \delta \frac{c_3}{c_3-2} = c_1 < 1
\end{align}
 where we used $\delta \leq \frac{1}{2}$ ($c_1 < 1$, by assumption, implies $\delta = \frac{c_1}{2} \frac{c_3-2}{c_3} \leq \frac{1}{2}$), and $\frac{m}{m-2s} \leq  \frac{c_3}{c_3-2}$ as a consequence of $m\geq c_3s$   
 (from \eqref{eq:finalcondfullyrand}) and $c_3>2$ ($c_3$ can be chosen freely as long as $c_3>0$). We therefore established that \eqref{eq:condTSCaffp} holds with probability $\geq 1- 4e^{- c' m}$, and application of Lemma~\ref{lem:cluster_isolation} concludes the proof.

\begin{proof}[Proof of Lemma~\ref{thm:fullyrandom}]
The proof relies on the following result from \cite{foucart_mathematical_2013}, which builds on a covering argument and the concentration inequality the Johnson-Lindenstrauss Lemma \cite{johnson_extensions_1984} is based on. 

\begin{lemma}[{\cite[Eq.~9.12 with $\rho=\frac{2}{e^3-1}$ ]{foucart_mathematical_2013}}]
Let $\mU$ be a $p\times s$ random matrix satisfying, for some $\tilde c >0$, for every $\vx \in \reals^s$, and for every $t\in (0,1)$, 
\begin{align}
\PR{\left|  \norm[2]{\mU \vx}^2 - \norm[2]{\vx}^2 \right| \geq t \norm[2]{\vx}^2 } \leq 2 e^{- \tilde c t^2 \p}.
\label{eq:johnslindeq}
\end{align}
Then, we have 
\begin{align}
\label{eq:boundopnuumi}
\PR{ \norm[2\to 2]{\herm{\mU} \mU - \mI_s } \geq \delta }  \leq 2 e^{- 0.6 \tilde c \delta^2 \p  + 3 s}.
\end{align}
\label{lem:boundopnuumi}
\end{lemma}

We note that \eqref{eq:johnslindeq} is satisfied, inter alia, for random matrices with i.i.d.~$\mathcal N(0, 1/p)$ entries.  

We show below that  \eqref{eq:genthmc2again} and \eqref{eq:genthmc1again} hold individually with probability $\geq 1-2 e^{-c' m}$. By a union bound, \eqref{eq:genthmc2again} and \eqref{eq:genthmc1again} thus hold simultaneously with probability $\geq 1-4 e^{-c' m}$, as desired.  
We start with \eqref{eq:genthmc2again}. 
First, note that since the rows of $\mU^{(k)}_{\D}$ indexed by $\D$ have all entries equal to zero by definition, we have $\herm{\mU^{(k)}_{\D} } \mU^{(\l)} = \herm{\mV}_i \mV_j$, where $\mV_i \in \reals^{p \times d}$ and $\mV_j \in \reals^{p \times d}$, with $p = m - |\D|$, denote the restrictions of $\mU^{(k)}$ and $\mU^{(\l)}$, respectively, to the rows indexed by $[m] \!\setminus \! \D$. Set $\tilde \mV_i = \sqrt{m/p} \mV_i$,  
 let 
$\mU = [\tilde \mV_i \; \tilde \mV_j ] \in \reals^{p \times 2d}$, 
and note that the entries of $\mU$ are i.i.d.~$\mathcal N(0,1/p)$. 
Using $m\geq p$, we have
\begin{align*}
\norm[2\to 2]{ \herm{ \mV}_i  \mV_j }
&\leq
\frac{m}{p}\norm[2\to 2]{ \herm{ \mV}_i  \mV_j } \\
&=
\norm[2\to 2]{ \herm{\tilde \mV}_i \tilde \mV_j } 
\leq 
\norm[2\to 2]{\herm{\mU} \mU -\mI_{2\d}}
\end{align*}
where the last inequality follows from the fact that  $\herm{\tilde \mV}_i \tilde \mV_j$ is a principal submatrix of $\herm{\mU} \mU -\mI_{2\d}$ \cite[Cor.~ 8.1.20]{horn_matrix_1986}. 
Therefore, we get
\begin{align}
\PR{\norm[2\to 2]{ \herm{\mV}_i \mV_j }  \geq \delta } 
&\leq
\PR{  \norm[2\to 2]{ \herm{\mU} \mU -\mI_{2\d}} \geq \delta } \nonumber \\
 &\leq 2  e^{-c_0 \delta^2 p + 6 d 
 } \label{eq:useboundopnuumi} \\
&\leq 2  e^{-c_0 \delta^2 (m- 2 s)  + 6 d}
\label{eq:usepleqmmts}
\end{align}
where $c_0 = 0.6 \tilde c$ ($\tilde c$ is the constant in Lemma \ref{lem:boundopnuumi}), \eqref{eq:useboundopnuumi} follows from Lemma \ref{lem:boundopnuumi}, 
and \eqref{eq:usepleqmmts} is a consequence of $p\geq m-2s$. 
Taking the union bound over all pairs $(i,j)$, i.e., over all pairs $(k,\l)$ with $k,\l \in [L]$ and for each of those pairs $(k,\l)$ over all $\D \subseteq [m]$ with $|\D| = 2 s$,  i.e., over $\binom{m}{2s} \leq \left( \frac{me}{2s} \right)^{2s}$ sets, we obtain
\begin{align}
\PR{\max_{i\neq j}\norm[2\to 2]{\herm{\mV_i}\mV_j} \geq \delta } 
&\leq L^2   \left(\frac{me}{2s}\right)^{2s}  2  e^{-c_0 \delta^2 (m- 2 s)  + 6 d}  \label{eq:tobesameasbel} \\ 
&=   2 e^{-c_0 \delta^2 (m- 2 s)  + 6 d    + 2\log L  + 2s \log\left(\frac{me}{2s} \right) } \nonumber \\
&=   2 e^{-c_0 \delta^2 \left[  m- 2 s  - \frac{2}{c_0 \delta^2} \left( 3 d    + \log L  + s \log\left(\frac{me}{2s}\right) \right)  \right] } \nonumber \\
&\leq   2 e^{-c_0 \delta^2 \left[  m- 2 s  - \frac{2}{c_0 c_2}  (m-c_3 s )  \right] } \label{eq:squarebracket} \\
&= 2 e^{-c_0 \delta^2  \left[   \left(1-  \frac{2}{c_0 c_2}\right)  m  + 2 s \left( \frac{c_3}{c_0 c_2}   - 1 \right)  \right] }   \nonumber \\
&\leq 2 e^{-c' m}   \label{eq:finalboundmaxvv}
\end{align}
where we used \eqref{eq:condconcsm_inthm} for \eqref{eq:squarebracket}, and \eqref{eq:finalboundmaxvv} holds with $c' = c_0 \delta^2 \left(1-  \frac{2}{c_0 c_2}\right)$ provided that $c_3 \geq c_0 c_2$, which, in turn, is guaranteed by choosing $c_3$ sufficiently large (recall that $c_3$ can be chosen freely). Note that $c'= c_0 \delta^2 \left(1-  \frac{2}{c_0 c_2}\right) > 0$ provided that $c_0 c_2 > 2$, which holds if $c_2$ is chosen sufficiently large. 
This concludes the proof of \eqref{eq:genthmc2again} holding with probability $\geq 1-2 e^{-c' m}$. 

It remains to show that \eqref{eq:genthmc1again} holds with probability $\geq 1-2 e^{-c' m}$. 
Applying Lemma \ref{lem:boundopnuumi} to $\tilde \mV_i$ (recall that the entries of $\mV_i$ are i.i.d.~$\mathcal N(0,1/p)$), we get
\[
\PR{ \norm[2\to 2]{ \herm{\tilde \mV_i} \tilde\mV_i -\mI_\d} \geq \delta } \leq 2  e^{-c_0 \delta^2 p  + 3 d} .
\]
Next, taking the union bound over all $L$ subspaces and over all $\D \subseteq [m]$ with $|\D| \leq 2 s$, yields 
\begin{align}
\PR{\max_{i} \norm[2\to 2]{ \herm{\tilde \mV_i} \tilde\mV_i -\mI_\d} \geq \delta } 
&\leq L   \left(\frac{me}{2s}\right)^{2s} 2  e^{-c_0 \delta^2 p  + 3 d}    \label{eq:sameasabad} \\ 
&\leq  2 e^{-c' m}
\label{eq:bmaxaerilml}
\end{align}
where we used the fact that the RHS of \eqref{eq:sameasabad} is smaller than the RHS of \eqref{eq:tobesameasbel} (recall that $p\geq m-2s$) and therefore \eqref{eq:bmaxaerilml} follows from 
\eqref{eq:finalboundmaxvv}. 
Next, note that for every $\va\in \reals^d$, we have
\begin{align*}
\norm[2]{\va}^2 - \norm[2]{\tilde \mV_i \va}^2 
&= \innerprod{(\mI_\d -\herm{\tilde \mV_i} \tilde\mV_i)\va }{\va} \\
&\leq \norm[2]{ (\mI_\d -\herm{\tilde \mV_i} \tilde\mV_i) \va } \norm[2]{\va}
\leq \norm[2\to 2]{\herm{\tilde \mV_i} \tilde\mV_i - \mI_\d } \norm[2]{\va}^2.
\end{align*}
It follows that
\[
1- \norm[2\to 2]{\herm{\tilde \mV_i} \tilde\mV_i - \mI_\d  } 
\leq \min_{\norm[2]{\va} = 1} \norm[2]{\tilde\mV_i \va}^2 = \min_{\norm[2]{\va} = 1} \norm[2]{\herm{\tilde \mV_i} \tilde\mV_i \va}
\]
and therefore (recall that $\tilde \mV_i = \sqrt{m/p} \mV_i$)
\begin{align}
\min_{i,\, \norm[2]{\va}=1} \norm[2]{ \herm{\mV}_i \mV_i  \va} 
&= \frac{p}{m} \min_{i, \norm[2]{\va}=1 } \norm[2]{ \herm{\tilde \mV_i} \tilde \mV_i  \va} \nonumber \\
&\geq  \frac{p}{m} \left( 1- \max_{i} \norm[2\to 2]{ \herm{\tilde \mV_i} \tilde\mV_i -\mI_\d} \right).
\label{eq:nowuseadf}
\end{align}
From \eqref{eq:nowuseadf}, and $p\geq m-2s$, we get
\begin{align*}
\PR{ \min_{i, \norm[2]{\va}=1} \norm[2]{ \herm{\mV}_i \mV_i  \va} \leq (1\!- \! \delta) \frac{m\!-\! 2 s}{m}   } 
&\leq \PR{ \min_{i, \norm[2]{\va}=1} \norm[2]{ \herm{\mV}_i \mV_i  \va} \leq (1\!-\! \delta) \frac{p}{m}   } \\
&\leq \PR{\max_{i} \norm[2\to 2]{ \herm{\tilde \mV_i} \tilde\mV_i -\mI_\d} \geq \delta } \leq 2 e^{-c' m}
\end{align*}
where the last inequality follows by application of \eqref{eq:bmaxaerilml}. 
\end{proof}

\section{Proof of Theorem~\ref{thm:outldete}}
\label{app:thm:outldete}

The proof consists of two parts, corresponding to the two statements in Theorem~\ref{thm:outldete}. First, we bound the probability of the outlier detection scheme failing to detect one or more outliers, and then we bound the probability of one or more of the inliers being misclassified as an outlier.

We start by bounding the probability of the outlier detection scheme failing to detect a given outlier. 
A union bound over all $N_0$ outliers will then yield a bound on the probability of the outlier detection scheme failing to detect one or more outliers. 
Let $\vx_j$ be an outlier. The probability of \eqref{eq:outldetrule} with $c = \sqrt{6}$ being violated for $\vx_j$, and therefore $\vx_j$ 
being misclassified as an inlier, can be upper-bounded as
\begin{align}
\PR{ \max_{i \in [N]\setminus \{j\}} \left|\innerprod{\vx_i}{\vx_j}\right|  > \frac{\sqrt{6 \log N }}{\sqrt{m}} } 
&\leq \sum_{i \in [N] \setminus \{j\}} \!
\PR{ \left|\innerprod{\vx_i}{\vx_j}\right|  > \frac{\sqrt{6 \log N }}{\sqrt{m}} \norm[2]{\vx_i}  } \nonumber \\
&\leq 
2 N e^{-3 \log N} = \frac{2}{N^2}
\label{eq:boundoadia1}
\end{align}
where we used a union bound and $\norm[2]{\vx_i} = 1$ in the first inequality and Proposition \ref{thm:hoeffsphere} in Appendix \ref{app:lemmata} in the second. 
Taking the union bound over all $N_0$ outliers we have thus established that the probability of our scheme failing to detect one or more outliers is at most $2N_0/N^2$. 

Next, we bound the probability of the outlier detection scheme misclassifying a given inlier $\vx_j \in \X_\l$
 as an outlier. 
 A union bound over all $n_\l$ inliers in $\X_\l$ will then complete the proof. 
For an inlier $\vx_j \in \X_\l$, we have
\begin{align*}
\max_{i \in [N] \setminus \{j\}} \left| \innerprod{\vx_i}{\vx_j}\right| 
\geq \max_{i \in [n_\l] \setminus \{j\}} \left| \innerprod{\vx_i^{(\l)}}{\vx_j^{(\l)}}\right| 
= \max_{i \in [n_\l] \setminus \{j\}} \left| \innerprod{\va_i^{(\l)}}{\va_j^{(\l)}}\right|.
\end{align*}
Using \eqref{eq:condoutldet}, i.e., $\sqrt{6\log N}/\sqrt{m} \leq 1/\sqrt{\d_{\max}} \leq 1/\sqrt{d_\l}$, the probability of \eqref{eq:outldetrule} holding 
can then be upper-bounded as
\begin{align}
\PR{\max_{i \in [N] \setminus \{j\}} \left| \innerprod{\vx_i}{\vx_j}\right|    \leq  \frac{\sqrt{6\log N}}{\sqrt{m}} } 
&\leq 
\PR{\max_{i \in [n_\l] \setminus \{j\}} \left| \innerprod{\va_i^{(\l)}}{\va_j^{(\l)}}\right|      \leq  \frac{1}{\sqrt{\d_\l}}  } \nonumber \\
&= \prod_{i \in [n_\l] \setminus \{j\}} \PR{\left| \innerprod{\va_i^{(\l)}}{\va_j^{(\l)}} \right| \leq \frac{1}{\sqrt{\d_\l}} } \nonumber \\
&\leq   \prod_{i \in [n_\l] \setminus \{j\}} \sqrt{\frac{2}{\pi}}  = e^{-\frac{1}{2}\log\left(\frac{\pi}{2}\right) (n_\l-1) } 
\label{eq:outlbinlprb}
\end{align}
where \eqref{eq:outlbinlprb} follows from \eqref{eq:ubonprzgsd} with $\nu=1$. 
Taking the union bound over all inliers in $\X_\l$ yields the desired upper bound $n_\l e^{-\frac{1}{2}\log\left(\frac{\pi}{2}\right) (n_\l-1)}$ on the outlier detection scheme misclassifying one or more of the inliers in $\X_\l$ as an outlier. 

\section{Proof of Theorem~\ref{thm:outldetnoisyc}}
\label{app:outldetnoisy}

The basic structure of the proof is the same as that of the proof of  Theorem~\ref{thm:outldete}. The individual steps are, however, a bit more technical, owing to the additive noise term. 

We start by bounding the probability of the outlier detection scheme failing to detect a given outlier. 
 A union bound over all $N_0$ outliers will, as before, yield the desired result. 
Let $\vx_j$ be an outlier and set $\beta = \sqrt{6 \log N }$. The probability that \eqref{eq:outldetrule} with $c=2.3 \sqrt{6}$ is violated for $\vx_j$ can be upper-bounded as
\begin{align}
\PR{ \max_{i \in [N] \setminus \{j\}}  \left|\innerprod{\vx_i}{\vx_j}\right|  > \frac{2.3 \beta}{\sqrt{m}} }  
&\leq \sum_{i \in [N] \setminus \{j\}} 
\PR{ \left|\innerprod{\vx_i}{\vx_j}\right|  > \frac{2.3 \beta}{\sqrt{m}}  } \nonumber \\
&\leq \! \! \sum_{i \in [N] \setminus \{j\}} \left(\PR{ \left| \innerprod{ \vx_i }{ \vx_j } \right|  \geq \frac{\beta}{\sqrt{m}} \norm[2]{\vx_i}  } +  \PR{  \norm[2]{ \vx_i }   \geq  2.3 } \right)
\label{eq:boundoutlinpro}
\end{align}
where we applied \eqref{eq:combviaproduc} with $X =  \left|\innerprod{\vx_i}{\vx_j}\right|$, $Y=\frac{\beta}{\sqrt{m}} \norm[2]{\vx_i}$,  and $\phi = \frac{2.3 \beta}{\sqrt{m}}$ to get \eqref{eq:boundoutlinpro}. 
We next bound the first term in the sum in \eqref{eq:boundoutlinpro}. 
Since $\vx_j \sim \mathcal N(\vect{0},(1/m) \mI_m)$, 
we have that, conditioned on $\vx_i$, $\innerprod{ \vx_j }{ \vx_i } \sim  \mathcal N(0, \norm[2]{\vx_i}^2 /m)$. Hence,  with $\beta = \sqrt{6\log N}$, it follows from \eqref{eq:qfunctionb} that 
\begin{align}
\PR{ \left| \innerprod{ \vx_i}{ \vx_j } \right|  \geq \frac{\beta}{\sqrt{m}} \norm[2]{\vx_i}  } \leq 2e^{-\frac{\beta^2}{2}} = \frac{2}{N^3}.
\label{eq:innprodwonc}
\end{align}
We next bound the second term in the sum in \eqref{eq:boundoutlinpro} and treat the cases where $\vx_i$ is an inlier and where it is an outlier separately. 
First, suppose that $\vx_i$ is an inlier. 
Since $\frac{1+2\sigma}{\sqrt{1+\sigma^2}} \leq 2.3$ for $\sigma\geq 0$, we have 
\begin{align}
\PR{  \norm[2]{ \vx_i }   \geq  2.3 }  
&\leq \PR{ \frac{1}{\sqrt{1+\sigma^2}} \norm[2]{ \mU^{(\l)} \va^{(\l)}_i  + \ve^{(\l)}_i }   \geq  \frac{1+2\sigma}{\sqrt{1+\sigma^2} }}
  \nonumber \\
&\leq \PR{ \norm[2]{ \mU^{(\l)} \va^{(\l)}_i } +  \norm[2]{\ve^{(\l)}_i }   \geq  1+2 \sigma   }  \label{eq:apptriangleineq} \\
&= \PR{  \norm[2]{\ve^{(\l)}_i }   \geq  2 \sigma   }  \nonumber \\
&\leq \frac{1}{N^3}  \label{eq:inliernormlb}
\end{align}
where \eqref{eq:apptriangleineq} follows from the triangle inequality, and for \eqref{eq:inliernormlb} we applied \eqref{eq:adfj2} with $\beta = \sqrt{6\log N}$ and used that $\beta \leq \sqrt{m}$, by assumption. 

Next, suppose that $\vx_i$ is an outlier. Applying \eqref{eq:adfj2} with $\sigma=1$ (again using that $\beta \leq \sqrt{m}$, by assumption), we have
\begin{align}
\PR{  \norm[2]{ \vx_i }   \geq  2.3 } \leq \frac{1}{N^3}.
\label{eq:outliernormlb}
\end{align}
Finally, combining \eqref{eq:innprodwonc}, \eqref{eq:inliernormlb} (for $\vx_i$ an inlier), and \eqref{eq:outliernormlb} (for $\vx_i$ an outlier) in  \eqref{eq:boundoutlinpro} yields 
\begin{align}
\PR{ \max_{i \in [N] \setminus \{j\}}  \left|\innerprod{\vx_i}{\vx_j}\right|  > \frac{2.3 \sqrt{6 \log N }}{\sqrt{m}} } 
\leq \sum_{i \in [N] \setminus \{j\}} \left(\frac{2}{N^3} + \frac{1}{N^3} \right) 
\leq 
\frac{3}{N^2}.
\nonumber
\end{align}
Taking the union bound over all $N_0$ outliers yields the desired result. 

Next, we bound the probability of our outlier detection scheme misclassifying a given inlier
 as an outlier. 
Consider the inlier $\vx_j\in \X_\l$. Then, we have
\begin{align*}
\max_{i \in [N] \setminus \{j\}} \left| \innerprod{\vx_i}{\vx_j}\right| 
&\geq \max_{i \in [n_\l] \setminus \{j\}} \left| \innerprod{\vx_i^{(\l)}}{\vx_j^{(\l)}}\right| \\
&\geq \max_{i \in [n_\l] \setminus \{j\}} \frac{1}{1+\sigma^2} \left|\innerprod{\va_i^{(\l)}}{\va_j^{(\l)}} + e_i^{(\l)}\right| \\ 
&\geq \max_{i \in [n_\l] \setminus \{j\}} \frac{1}{1+\sigma^2} (\tilde z_i^{(\l)} - |e_i^{(\l)}|)
\end{align*}
where we used the reverse triangle inequality, and $\tilde z_i^{(\l)}$ and $e_i^{(\l)}$ were defined in \eqref{eq:defztildejk} and \eqref{eq:defejk}, respectively. 
Thus, for $\epsilon \geq 0$, under the assumption  
\begin{align}
\frac{1}{1+\sigma^2} \left(\frac{1}{\sqrt{d_\l}}-\epsilon \right) \geq \frac{2.3 \sqrt{6 \log N }}{\sqrt{m}} 
\label{eq:condrudimentfod}
\end{align}
resolved below, we have 
\begin{align}
\PR{ \max_{i \in [N] \setminus \{j\}} \left|\innerprod{\vx_i}{\vx_j}\right|  \leq\frac{2.3 \sqrt{6 \log N }}{\sqrt{m}}  }  
&\leq 
\PR{\max_{i \in [n_\l] \setminus \{j\}} \frac{1}{1+\sigma^2} (\tilde z_i^{(\l)} - |e_i^{(\l)}|)  \leq \frac{1}{1+\sigma^2}\left( \frac{1}{\sqrt{d_\l}} - \epsilon \right) }  \nonumber \\
&\leq \PR{ \max_{i \in [n_\l] \setminus \{j\}} \tilde z_i^{(\l)}  -   \max_{i \in [n_\l] \setminus \{j\}} |e_i^{(\l)}|    \leq \frac{1}{\sqrt{\d_\l}} - \epsilon   } \nonumber \\
&\leq \PR{ \max_{i \in [n_\l] \setminus \{j\}} \tilde z_i^{(\l)}  \leq \frac{1}{\sqrt{\d_\l}}   }  +  \PR{  \epsilon  \leq  \max_{i \in [n_\l] \setminus \{j\}} |e_i^{(\l)}| }  \label{eq:useageq:splitprob} 
\end{align}
where  \eqref{eq:useageq:splitprob} follows from \eqref{eq:splitprob} with $X = \max_{i \in [n_\l] \setminus \{j\}} \tilde z_i^{(\l)} - \frac{1}{\sqrt{\d_\l}}$, $Y=\max_{i \in [n_\l] \setminus \{j\}} |e_i^{(\l)}|-\epsilon$, and $\phi = \varphi = 0$. 
Next, note that  \eqref{eq:ubonprzgsd} with $\nu=1$ yields 
\begin{align}
\PR{ \max_{i \in [n_\l] \setminus \{j\}} \tilde z_i^{(\l)}  \leq \frac{1}{\sqrt{\d_\l}}   } &= \prod_{i \in [n_\l] \setminus \{j\}} \PR{  \tilde z_i^{(\l)}  \leq \frac{1}{\sqrt{\d_\l}}   } 
\leq \prod_{i \in [n_\l] \setminus \{j\}} \sqrt{\frac{2}{\pi}} =  e^{-\frac{1}{2}\log\left(\frac{\pi}{2}\right) (n_\l-1) }.
\label{eq:asdflooik}
\end{align}
Application of \eqref{eq:asdflooik} and \eqref{eq:boundonnoise} with $\epsilon = \frac{2 \sigma(1+\sigma)}{\sqrt{m}} \sqrt{6 \log N}$ (using that $\beta \leq \sqrt{m}$, as verified below) to \eqref{eq:useageq:splitprob} yields 
\begin{align}
\PR{ \max_{i \in [N] \setminus \{j\}} \left|\innerprod{\vx_i}{\vx_j}\right|  \leq\frac{2.3 \sqrt{6 \log N }}{\sqrt{m}}  }  \leq e^{-\frac{1}{2}\log\left(\frac{\pi}{2}\right) (n_\l-1) }  +  n_\l \frac{7}{N^3}.  \label{eq:lastineqinas}
\end{align}

We next show that choosing $c_1$ sufficiently small, specifically $c_1\leq 1/6$, guarantees that $\beta \leq \sqrt{m}$. To this end simply note that  \eqref{eq:condnoisyoutldet} implies 
\[
\frac{1}{m} \leq \frac{d_{\max}}{m} \leq \frac{c_1}{(1+\sigma^2)^2 \log N} \leq \frac{c_1}{\log N} 
\] 
and take $c_1 \leq 1/6$. 

Taking a union bound over all inliers in $\X_\l$ shows that the probability of the outlier detection scheme misclassifying one or more of the inliers in $\X_\l$ as an outlier is at most 
\[
n_\l\left(e^{-\frac{1}{2}\log\left(\frac{\pi}{2}\right) (n_\l-1) }  +  n_\l \frac{7}{N^3}\right).
\]

Finally, we resolve \eqref{eq:condrudimentfod} by showing that it is implied, for all $\l\in [L]$, by \eqref{eq:condnoisyoutldet}. Rewriting \eqref{eq:condrudimentfod}
yields 
\[
\frac{1}{1+\sigma^2} \frac{1}{\sqrt{6 \log N}} \geq \frac{\sqrt{\d_\l}}{\sqrt{m}} \left( 2.3 + \frac{2 \sigma(1+\sigma)}{1+\sigma^2} \right).
\]
Since $\frac{\sigma(1+\sigma)}{1+\sigma^2}\leq 1.3$ for $\sigma\geq 0$, \eqref{eq:condrudimentfod} is implied by
\[
\frac{1}{1+\sigma^2} \frac{1}{\sqrt{6 \log N}} \geq \frac{\sqrt{d_{\max}}}{\sqrt{m}}  4.9 
\]
which equals \eqref{eq:condnoisyoutldet} with $c_1=\frac{1}{ (4.9)^2 \cdot 6}$.

\section{Supplementary results}
\label{app:lemmata} 

For convenience, in the following, we summarize tail bounds from the literature that are frequently used throughout this paper. We start with a well-known tail bound on Gaussian random variables. 

\begin{lemma}[{\cite[Prop.~19.4.2]{lapidoth_foundation_2009}}]
Let $x\sim\mathcal N(0,1)$. For $\beta \geq \frac{1}{\sqrt{2\pi}}$, we have
\begin{equation}
\PR{x \geq \beta} \leq e^{- \frac{\beta^2}{2}}.
\label{eq:qfunctionb}
\end{equation}
\label{lem:qfunction}
\end{lemma}

\begin{theorem}[{\cite[Eq. 1.6]{ledoux_probability_1991}}] 
Let $f$ be Lipschitz on $\reals^m$ with Lipschitz constant $L\in \reals$,  i.e., $|f(\vb_1) - f(\vb_2)| \leq L \norm[2]{\vb_1-\vb_2}$, for all $\vb_1,\vb_2 \in \reals^m$, and let $\vx \sim \mathcal N(\mathbf 0,\mI_m)$. Then, for $\beta>0$, we have
\[
\PR{f(\vx) \geq \EX{f(\vx)} + \beta } \leq e^{- \frac{\beta^2}{ 2 L^2} }.
\]
\label{thm:talagrands_conceq}
\end{theorem}

Let $\vx \sim \mathcal N(\mathbf 0,\mI_m)$. 
Applying the concentration inequality in Theorem \ref{thm:talagrands_conceq} to $f(\vx) = \norm[2]{ \vx}$ which has Lipschitz constant $L=1$, and using Jensen's inequality to get $\left( \EX{\norm[2]{\vx}}  \right)^2 \leq   \mathbb E \big[ \norm[2]{\vx}^2 \big] = m$, we obtain
\begin{align}
\PR{\norm[2]{\vx} \geq \sqrt{m} + \beta } \leq e^{  - \frac{\beta^2}{ 2 } }.
\label{eq:gaussnormconc}
\end{align}

\begin{proposition}[{E.g.,~\cite[Ex.~5.25]{vershynin_introduction_2012}}]
Let $\va$ be uniformly distributed on $\US{m}$ and fix $\vb \in \reals^m$. Then, for $\beta\geq 0$, we have
\[
\PR{ \left|\innerprod{\va}{ \vb }\right|  > \frac{\beta}{\sqrt{m}} \norm[2]{ \vb} } 
\leq 2 e^{-\frac{ \beta^2}{2}}. 
\]
\label{thm:hoeffsphere}
\end{proposition}

\end{document}